%% file: main.tex
\ifdefined\XeTeXversion\else\pdfoutput=1\fi
\documentclass{article}
\PassOptionsToPackage{table}{xcolor}
\usepackage[T1]{fontenc}
\usepackage{iclr2027_conference,times}
\usepackage{microtype}

\usepackage{amsmath,amssymb,amsthm,mathtools}
\usepackage{booktabs}
\usepackage{multirow}
\usepackage{array}
\usepackage{graphicx}
\usepackage{wrapfig}
\usepackage{float}
\usepackage{placeins}
\usepackage{needspace}
\usepackage{algorithm}
\usepackage{algorithmic}
\usepackage{xcolor}
\makeatletter
\renewcommand\float@makebox[1]{%
  \vbox{\hsize=#1 \@parboxrestore
    \begingroup\normalcolor%
    \@fs@pre\@fs@iftopcapt
      \ifvoid\@floatcapt\else\unvbox\@floatcapt\par\@fs@mid\fi
      \unvbox\@currbox
    \else\unvbox\@currbox
      \ifvoid\@floatcapt\else\par\@fs@mid\unvbox\@floatcapt\fi
    \fi\par\@fs@post\vskip\z@
    \endgroup}}
\makeatother
\input{style/tables}
\usepackage{pgfplots}\pgfplotsset{compat=1.18}
\usepackage{hyperref}
\hypersetup{
  colorlinks=true,
  citecolor=blue!50!black,
  linkcolor=black,
  urlcolor=blue!50!black
}
\usepackage{url}
\usepackage{cleveref}
\crefname{figure}{Figure}{Figures}\Crefname{figure}{Figure}{Figures}
\crefname{table}{Table}{Tables}\Crefname{table}{Table}{Tables}
\crefname{section}{Section}{Sections}\Crefname{section}{Section}{Sections}
\crefname{appendix}{Appendix}{Appendices}\Crefname{appendix}{Appendix}{Appendices}
\crefname{equation}{Eq.}{Eqs.}\Crefname{equation}{Eq.}{Eqs.}
\crefname{proposition}{Proposition}{Propositions}\Crefname{proposition}{Proposition}{Propositions}
\crefname{finding}{Observation}{Observations}\Crefname{finding}{Observation}{Observations}


\newcommand{\vth}{v_\theta}
\newcommand{\vfix}{v_{\mathrm{ref}}^{\mathrm{fix}}}
\newcommand{\vmov}{v_{\mathrm{ref}}^{(k)}}
\newcommand{\sg}[1]{\mathrm{sg}\!\left[#1\right]}
\newcommand{\EE}{\mathbb{E}}
\newcommand{\RR}{\mathbb{R}}
\newcommand{\clip}{\mathrm{clip}}
\DeclareMathOperator*{\argmin}{arg\,min}

\newcolumntype{C}[1]{>{\centering\arraybackslash}p{#1}}

\newtheorem{proposition}{Proposition}

\definecolor{hlSigned}{HTML}{0B63CE}
\definecolor{hlReg}{HTML}{C2255C}
\newcommand{\hs}[1]{\textcolor{hlSigned}{#1}}
\newcommand{\hr}[1]{\textcolor{hlReg}{#1}}
\newcommand{\E}{\mathbb{E}}

\title{
SIFR: Stable Signed Flow Reinforcement}

\newcommand{\authsep}{\hspace{1.1em}}
\author{%
\makebox[\dimexpr\textwidth-2\tabcolsep\relax][c]{\begin{tabular}[t]{@{}c@{}}
Runlong Liao$^{1,2,*}$\authsep Xingqian Xu$^{1}$\authsep Lizhang Chen$^{2}$\authsep Baiyu Su$^{2}$\authsep Yufan Huang$^{1}$ \\[5pt]
Zhonghao Wang$^{1}$\authsep Akash Gokul$^{1}$\authsep Ming-Yu Liu$^{1}$\authsep Qiang Liu$^{2}$\authsep Yogesh Balaji$^{1}$ \\[6pt]
\normalfont $^{1}$NVIDIA\qquad $^{2}$The University of Texas at Austin \\[3pt]
\normalfont \url{https://research.nvidia.com/labs/cosmos-lab/sifr/}
\end{tabular}}%
}

\iclrfinalcopy

\begin{document}

\maketitle
\vspace{-20pt}%
\lhead{}
\begingroup
\renewcommand{\thefootnote}{\fnsymbol{footnote}}
\footnotetext[1]{Work done during an internship at NVIDIA Cosmos Lab.}
\endgroup

\begin{abstract}
Reward-based fine-tuning of flow models requires effectively promoting high-reward and suppressing low-reward samples, while maintaining training stability over many rounds of iterative updates.
Iterative reward fine-tuning provides a simple starting point, but the positive reward weights give no explicit unfitting signal, and repeated reweighting concentrates the distribution and causes collapse.
We propose Stable Signed Flow Reinforcement (SIFR), which replaces the reward weights by signed advantages to supply the missing unfitting signal, and constrains the velocity drift from a reference model to stabilize the updates, with a Lagrange multiplier that adapts the regularization strength to a prescribed drift budget.
We further show that existing flow RL methods fall within this framework: they implicitly minimize the signed loss, differing in their regularization choices.
We validate SIFR on SD3.5 Medium (2.5B) and FLUX.2 Klein (9B) and scale it to Cosmos 3 Super Text2Image (64B): across all 12 backbone--reward pairs with preference and rule-based rewards, it matches or exceeds existing methods, reaches their peak rewards much faster, and keeps improving over long training where they plateau or collapse.
\end{abstract}

\raggedbottom
\begingroup
\setlength{\intextsep}{2pt} \setlength{\abovecaptionskip}{6pt}
\begin{figure}[H]
\centering
\makebox[\textwidth][c]{\resizebox{\textwidth}{!}{\input{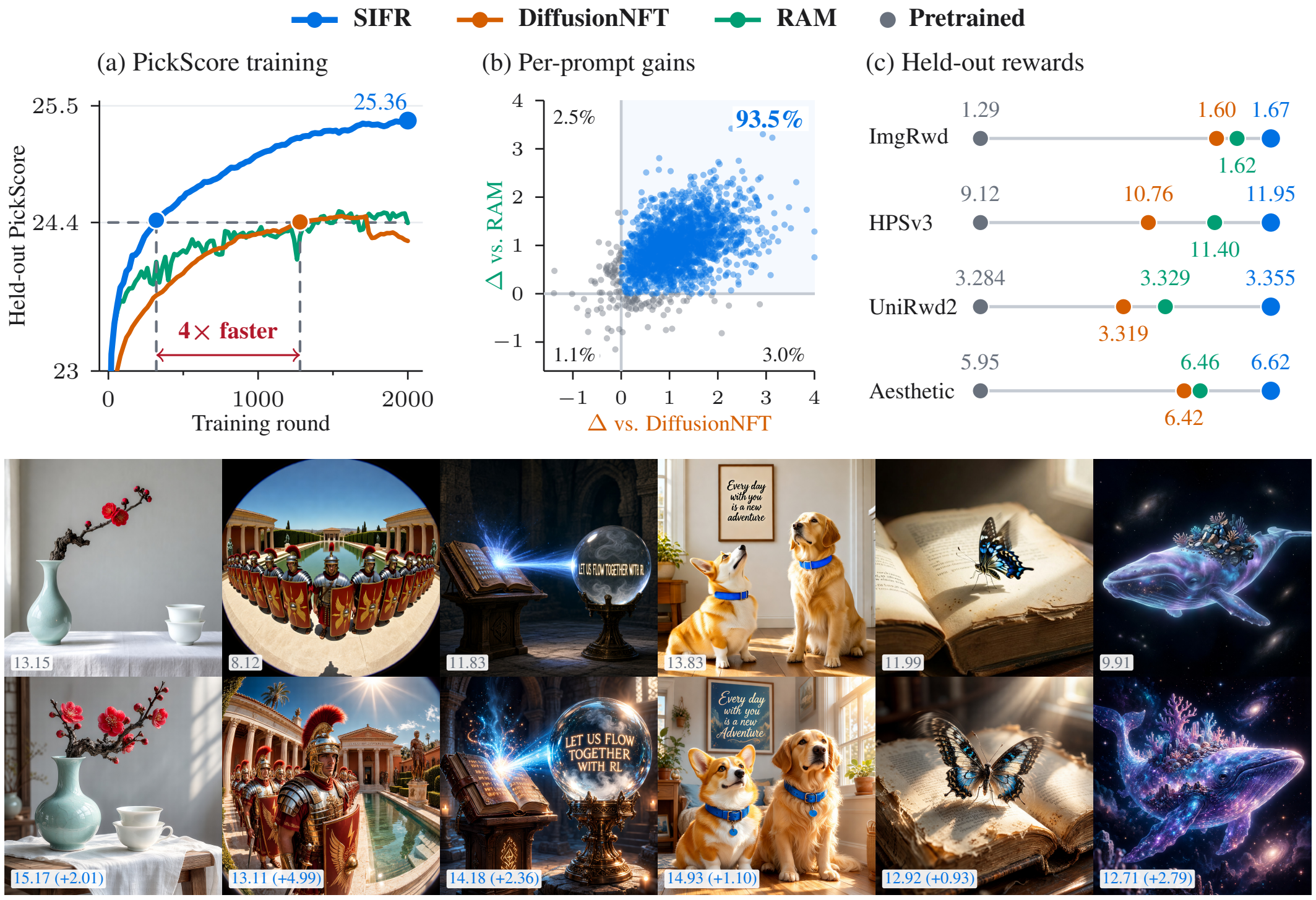}}}
\par\vspace{-0.3em}%
\caption{\textbf{SIFR on Cosmos 3 Super Text2Image.} (a) Held-out PickScore during single-reward training; SIFR reaches the baselines' plateau $4\times$ sooner and keeps improving. (b) Per-prompt gains at round $2000$; $93.5\%$ of prompts beat both baselines. (c) Held-out rewards after $500$ rounds of multi-reward training. Images: pretrained above SIFR, labeled with held-out HPSv3 and its gain.}
\label{fig:teaser}
\end{figure}
\endgroup

\clearpage
\flushbottom

\section{Overview}
\label{sec:overview}

Reward-based post-training has become a standard component of language-model training~\citep{christiano2017deep,stiennon2020summarize,ouyang2022rlhf,bai2022helpful,touvron2023llama2,shao2024grpo,deepseek2025r1} and is increasingly being adopted for image and video generation with flow- and diffusion-based models~\citep{fan2023dpok,black2024ddpo,clark2024draft,liu2025flowgrpo,xue2025dancegrpo,liu2025videofeedback,fan2025orw,xue2025awm,zheng2026diffusionnft,bergmeister2026ram,kveton2026advantageflow,fu2026flowawr}.
Despite substantial recent progress, post-training methods for these generative models have yet to converge on a standard paradigm as they have for LLMs.
A central challenge is how to effectively promote high-reward samples and suppress low-reward ones while maintaining training stability over many rounds of iterative updates.

A basic building block of reinforcement learning (RL) is iterative reward fine-tuning~\citep{peters2007rwr}.
For a rectified flow (RF) model~\citep{liu2023flow,lipman2023flow,albergo2023stochastic} with velocity field $v_\theta$ and a positive reward $r$, each round $k$ refits the model to its own rollouts, weighted by reward:
\[
\theta^{(k+1)}
=
\argmin_\theta\;
\EE_{X_0\sim\pi_0,\;X_1\sim\pi_{\theta^{(k)}}}\!\big[
\hs{r(X_1)}
\|v_\theta(X_t,t)-(X_1-X_0)\|^2
\big],
\]
where $X_1$ is a rollout from the current model $\theta^{(k)}$, $X_0\sim\pi_0$ is initial noise, $X_t=(1-t)X_0+tX_1$ with time $t\in[0,1]$, and we abbreviate the expectation over these draws as $\EE_{\theta^{(k)}}$ below.
However, iterative reward fine-tuning of flow models fails in practice for two reasons: 1) repeated reward weighting concentrates the model distribution and leads to collapse; and 2) positive weights encourage fitting high-reward samples but provide no explicit signal to unfit low-reward samples.

A natural remedy is to replace the reward weighting term $r(X_1)$ with the advantage $A(X_1)$, so that below-average samples receive negative weights and are explicitly suppressed.
However, the resulting signed loss, an MSE loss with negative weights, can be unbounded below and lead to training instability.

Interestingly, although their original losses are positive, existing flow RL methods have the same updates as the signed loss combined with an implicit regularization toward a reference model, and differ mainly in the form of this regularization (\Cref{sec:weighted}).
This regularization is therefore the key to stable training: if it is too weak, training becomes unstable; if it is too strong, the model stays close to its reference, learns slowly, and reaches a lower reward ceiling.

In this work, we propose \textbf{Stable Signed Flow Reinforcement} (\textbf{SIFR}), which casts flow RL as a constrained optimization problem, replacing this implicit regularization with an explicit constraint: it minimizes the signed loss subject to a bound on the velocity drift from a reference model.
At each round $k$, SIFR solves
\[
\min_\theta\;
\EE_{\theta^{(k)}}\!\big[
\hs{A(X_1)}
\|v_\theta(X_t,t)-(X_1-X_0)\|^2
\big]
\quad\mathrm{s.t.}\quad
\hr{
\EE_{\theta^{(k)}}\!\big[
\|v_\theta(X_t,t)-v_{\mathrm{ref}}^{(k)}(X_t,t)\|^2
\big]\le \delta
}.
\]
where $v_{\mathrm{ref}}^{(k)}$ is the reference velocity field in round $k$, and $\delta\ge0$ is the drift budget.
In practice, we solve this problem by alternating updates of the model parameters and the Lagrange multiplier, which keeps the regularization only as strong as the budget requires, raising the reward ceiling while keeping optimization stable.
We summarize our contributions below.

\begingroup\setlength{\leftmargini}{1.2em}\setlength{\topsep}{0pt}\setlength{\partopsep}{0pt}%
\begin{itemize}
\item We introduce SIFR, which formulates online flow RL as a \textbf{constrained optimization problem}. We characterize its saddle point and derive a practical algorithm that alternates model updates with a Lagrange multiplier update (\Cref{sec:method}). SIFR imposes no restriction on the sampler and needs no likelihoods: its objective is the rectified flow loss with signed weights and a quadratic drift penalty, which makes it straightforward to implement on top of pretraining code.
\item We show that existing flow RL methods, such as DiffusionNFT~\citep{zheng2026diffusionnft}, RAM~\citep{bergmeister2026ram}, and AWM~\citep{xue2025awm}, are \textbf{gradient-equivalent to the signed loss} although their original losses are positive, and that they differ mainly in their regularization toward a reference model, explaining their stability and limits (\Cref{sec:weighted}).
\item We validate SIFR on SD3.5 Medium (2.5B) and FLUX.2 Klein (9B) and scale it to Cosmos 3 Super Text2Image (64B) with four rewards (\Cref{sec:exp}). Across all 12 backbone--reward pairs, SIFR matches or exceeds DiffusionNFT and RAM; on preference rewards it reaches their peak scores \textbf{2--4$\times$ faster} and keeps improving where they plateau, and its gains transfer to held-out rewards and prompt sets.
\end{itemize}
\endgroup

\Needspace{8\baselineskip}
\section{Flow RL as Regularized Signed Regression}
\label{sec:weighted}

\paragraph{Rectified Flow}
Let $\pi_0$ be a source distribution on $\RR^d$, such as $\mathcal N(0,I)$, and let $\pi_1$ be the target data distribution of the same dimension.
Given noise-data pairs $(X_0,X_1)$ with $X_0\sim\pi_0$ and $X_1\sim\pi_1$, rectified flow (RF) learns a velocity field $v_\theta(x,t)$ by minimizing the mean squared error
\begin{equation}
\min_\theta\;\EE_{X_0\sim\pi_0,\;X_1\sim\pi_1}\big[\|v_\theta(X_t,t)-(X_1-X_0)\|^2\big],
\qquad
X_t = (1-t)X_0 + tX_1,
\label{eq:rf}
\end{equation}
where $X_t$ interpolates between the endpoints, and $\E[\cdot]$ averages over $(X_0,X_1)$ and an independent time $t\in[0,1]$.
The minimizer over unrestricted velocity fields is the conditional expectation
\[
v^\star(x,t)=\EE[X_1-X_0\mid X_t=x].
\]
One of the key properties of RF is marginal preservation: under standard regularity conditions, the process $Z_t$ generated by the ODE $\dot Z_t=v^\star(Z_t,t)$, initialized with $Z_0\sim\pi_0$, has the same distribution as $X_t$ at every $t\in[0,1]$.
In particular, $Z_1\sim\pi_1$.
See \citet{liu2023flow,liu2022rectified} for further details.

\subsection{Iterative Reward Fine-Tuning and Its Collapse}
\label{sec:collapse}
\paragraph{Reward Tilting of Rectified Flow}
Assume we are given a reward function $r(x)\geq 0$ without loss of generality.
The reward-tilted distribution is $\pi_1^r(x)=r(x)\pi_1(x)/Z_r$, where $Z_r=\EE[r(X_1)]$.
The RF velocity that generates $\pi_1^r$ is the minimizer of the reward-weighted RF loss:
\begin{equation}
\min_\theta\;\EE\big[r(X_1)\|v_\theta(X_t,t)-(X_1-X_0)\|^2\big]
\implies
v^r(x,t)=\frac{\EE[r(X_1)(X_1-X_0)\mid X_t=x]}{\EE[r(X_1)\mid X_t=x]}.
\label{eq:weighted-loss}
\end{equation}
Intuitively, reweighting the regression loss is equivalent to reweighting the data samples, so that high-reward samples are generated more frequently.

\paragraph{Iterative Reward Fine-Tuning}
The reward tilting above reweights samples from a fixed data distribution $\pi_1$.
A more aggressive approach is to alternate generation and fine-tuning, using samples from the updated model at each round, as in online rejection sampling fine-tuning for language models~\citep{touvron2023llama2,shao2024grpo}.
For RF, the corresponding update with rewards $r(x)\ge 0$ can be expressed as
\begin{equation}
\theta^{(k+1)}
=\argmin_\theta\;\EE_{\theta^{(k)}}\!\big[r(X_1)\|v_\theta(X_t,t)-(X_1-X_0)\|^2\big].
\label{eq:iterative-weighted}
\end{equation}
At round $k$, the current velocity field $v_{\theta^{(k)}}$ generates the rollouts $X_1$, the model is updated with optimizer steps on their reward-weighted loss, and the updated field generates the next round's rollouts, so the training data are continually refreshed.
Unfortunately, iterative reward fine-tuning as in \Cref{eq:iterative-weighted} fails in practice, for two reasons: 1) repeated tilting progressively concentrates probability mass near reward maximizers and leads to distributional collapse (\cref{app:repeated-tilting}); and 2) positive weights encourage the model to fit high-reward samples but provide no explicit signal to unfit low-reward samples, limiting the effectiveness of the updates.

\subsection{Signed Regression and Its Instability}
\label{sec:signed-regression}
\paragraph{Signed Tilting}
A natural next step is to replace $r(X_1)$ with a signed advantage $A(X_1)$, giving
\begin{equation}
\theta^{(k+1)} = \argmin_\theta\;\EE_{\theta^{(k)}}\!\big[A(X_1)\|v_\theta(X_t,t)-(X_1-X_0)\|^2\big],
\qquad
A(x)=\frac{r(x)-\mu_r}{\sigma_r+\epsilon},
\label{eq:naive-signed-mse}
\end{equation}
where $\mu_r$ and $\sigma_r$ are the reward mean and standard deviation~\citep{shao2024grpo}.
Optimizing this signed loss \emph{minimizes} the loss for above-average samples $r(X_1)>\mu_r$ and \emph{maximizes} it for below-average samples, which yields an unfitting signal.

However, this unfitting signal introduces a new source of instability: an MSE loss with negative weights can be unbounded below.
To understand its stability, it is useful to expand the quadratic terms of the loss as
\[
\EE_{\theta^{(k)}}\big[m(X_t,t)\|v_\theta(X_t,t)\|^2-2n(X_t,t)^\top v_\theta(X_t,t)\big],
\]
where the quadratic and linear coefficients (whose dependence on the round $k$ we suppress) are
\[
m(x,t)=\EE_{\theta^{(k)}}[A(X_1)\mid X_t=x],
\qquad
n(x,t)=\EE_{\theta^{(k)}}[A(X_1)(X_1-X_0)\mid X_t=x].
\]
Hence, the optimization in \eqref{eq:naive-signed-mse} is stable when $m(x,t)>0$, in which case the solution is
\begin{equation}
v^A(x,t)=\frac{n(x,t)}{m(x,t)}
=\frac{\EE_{\theta^{(k)}}[A(X_1)(X_1-X_0)\mid X_t=x]}{\EE_{\theta^{(k)}}[A(X_1)\mid X_t=x]}.
\end{equation}
When $m(x,t)<0$, the ratio $v^A(x,t)$ is a conditional maximum, and the conditional loss is unbounded below (\cref{prop:signed-geometry}).
If both sign regions of $m(\cdot,t)$ have positive training measure, then $v^A$ can only be a saddle of the signed loss.
Moreover, the velocity field $v^A$ becomes singular near boundary $\{x:m(x,t)=0\}$.

Despite these singularities, the velocity field $v^A$ can still define valid sampling dynamics through Signed Rectified Flow (Signed RF)~\citep{liao2026signed}.
Assume $Z_A=\EE[A(X_1)]>0$.
With the regularity assumptions of Signed RF, one can show (\cref{app:signed-weight}) that ODE trajectories driven by $v^A$ and initialized from an independent source $\pi_0$ remain in the positive region $\{x:m(x,t)>0\}$ for $t\in[0,1)$, and the generated density is a truncated density of $\pi_1(x) A(x)$ within the positive region:
\[
\rho_1(x)=A(x)\pi_1(x)
\mathbf{1}\{x\in\Omega_{\mathrm{reach}}\}/Z_A,
\qquad
\Omega_{\mathrm{reach}}\subseteq\{x:A(x)>0\},
\]
where the reachable region $\Omega_{\mathrm{reach}}$ is determined by the flow dynamics and satisfies the normalization condition $\E[A(X_1)\mathbf{1}\{X_1\in\Omega_{\mathrm{reach}}\}]=Z_A$.
For a fixed advantage $A$, Signed RF thus generates the density on a reachable subset of the positive region and excludes the negative region.

However, learning this field $v^A$ remains difficult: minimizing the signed loss does not recover its saddle, and singularities near the zero boundary of $v^A$ make direct neural approximation and numerical integration sensitive to errors.
As we show next, existing flow RL methods stay stable only because they regularize this signed loss implicitly.

\paragraph{Regularized Signed Regression}
We show that although existing flow RL methods such as DiffusionNFT~\citep{zheng2026diffusionnft} and RAM~\citep{bergmeister2026ram} are formulated with positive MSE losses, their stop-gradient operations and parameterizations make their update gradients equivalent to those of the signed loss with regularization toward a reference model, and the methods differ mainly in that regularization.
We summarize the gradient-equivalent objectives in the unified form:
\begin{equation}
  \begin{aligned}
    \mathcal L^{(k)}(\theta)
    =\EE_{\theta^{(k)}}\!\Big[
    &\hs{w(X_1)}\|v_\theta(X_t,t)-(X_1-X_0)\|^2
    \hr{
      + c_{\mathrm{ref}}^{(k)}D(v_\theta,v_{\mathrm{ref}}^{(k)})
      + c_{\mathrm{ref}}^{\mathrm{fix}} D(v_\theta,v_{\mathrm{ref}}^{\mathrm{fix}})
    }
    \Big],
  \end{aligned}
  \label{eq:comparison-form}
\end{equation}
where $D(v,v_{\mathrm{ref}})=\|v(X_t,t)-v_{\mathrm{ref}}(X_t,t)\|^2$ measures the drift from a reference model, which is either fixed ($v_{\mathrm{ref}}^{\mathrm{fix}}$, e.g., a pretrained teacher) or updated across rounds ($v_{\mathrm{ref}}^{(k)}$, given by the current snapshot $\theta^{(k)}$ or an exponential moving average $\theta_{\mathrm{ema}}^{(k)}$ of past iterates); both are detached during model updates.
\Cref{tab:connections} below lists for each method the weight $w(X_1)$ and the coefficients $c_{\mathrm{ref}}^{(k)}$ and $c_{\mathrm{ref}}^{\mathrm{fix}}$ that reproduce its update gradient; derivations are in \cref{app:implicit-proof}.

\begingroup
\setlength{\abovecaptionskip}{6pt} \setlength{\belowcaptionskip}{3pt} \newsavebox{\connectionsbox} \newcommand{\connectionsep}{\arrayrulecolor{black!22}\specialrule{0.3pt}{1.5pt}{1.5pt}\arrayrulecolor{black}} \newcommand{\connectionmethod}[2]{#1\hspace{0.25em}{\footnotesize\citep{#2}}}
  \begin{table}[H]%
    \centering
\footnotesize \setlength{\tabcolsep}{5pt} \renewcommand{\arraystretch}{1.0}
    \sbox{\connectionsbox}{%
      \begin{tabular}{cc!{{\color{black!22}\vrule width 0.3pt}}ccc}
\toprule
        \textbf{Method} & $v_{\mathrm{ref}}^{(k)}$ & $w(X_1)$
          & $c_{\mathrm{ref}}^{(k)}$ & $c_{\mathrm{ref}}^{\mathrm{fix}}$ \\
\midrule \connectionmethod{ORW-CFM-W2}{fan2025orw}
          & \textemdash & $e^{\tau r(X_1)}$ & $0$ & $\alpha$ \\
\connectionsep \raisebox{-0.7pt}{\connectionmethod{AWM}{xue2025awm}}
          & EMA & $(1-t)A(X_1)$ & $\beta(1-t)$ & $0$ \\
\connectionsep \multirow[c]{2}{*}[-1.2pt]{\connectionmethod{DiffusionNFT}{zheng2026diffusionnft}}
          & No EMA & \multirow[c]{2}{*}[-1.2pt]{$\beta A(X_1)$} & $0$ & \multirow[c]{2}{*}[-1.2pt]{$0$} \\
          & EMA & & $\beta(\beta-A(X_1))$ & \\
\connectionsep \multirow[c]{2}{*}[-1.2pt]{\connectionmethod{RAM}{bergmeister2026ram}}
          & No EMA & \multirow[c]{2}{*}[-1.2pt]{$\eta A(X_1)$} & $0$ & \multirow[c]{2}{*}[-1.2pt]{$1$} \\
          & EMA & & $-\eta A(X_1)$ & \\
\connectionsep \raisebox{-0.7pt}{\connectionmethod{AdvantageFlow}{kveton2026advantageflow}}
          & EMA & $(1-t)^2A(X_1)$ & $(1-t)^2\gamma(X_1)$ & $(1-t)^2\lambda$ \\
\connectionsep \raisebox{-0.7pt}{\connectionmethod{FlowAWR}{fu2026flowawr}}
          & EMA & $A(X_1)$ & $1-A(X_1)$ & $0$ \\
\bottomrule
      \end{tabular}}
\usebox{\connectionsbox}
    \caption{\footnotesize\textbf{Existing flow RL methods minimize a regularized signed loss.} Each row gives the weight $w(X_1)$ and the coefficients of \cref{eq:comparison-form} that reproduce the method's update. The $v_{\mathrm{ref}}^{(k)}$ column is the dynamic reference, an EMA of the parameters or the detached current snapshot (No EMA). ORW-CFM-W2 uses positive exponential weights, the others signed advantages. Reported values: $\gamma(X_1)\in\{1.1,\,1-A(X_1)\}$ for AdvantageFlow, $\eta\in\{100,1000\}$ for RAM, $\beta\in\{0.1,1\}$ for DiffusionNFT, and a task-dependent $\tau$ for ORW-CFM-W2.}
    \label{tab:connections}
  \end{table}
\endgroup

\section{Stable Signed Flow Reinforcement}
\label{sec:method}

\Cref{sec:weighted} shows that existing flow RL methods implicitly minimize the signed loss and differ mainly in their coefficients, in particular the choice of regularization toward the reference model (\cref{tab:connections}), which leads to complicated designs.
We now introduce SIFR, which formulates the problem as a constrained optimization problem, replacing the regularization penalty with a constraint.

\subsection{Constrained Formulation}

\paragraph{Stable Signed Flow Reinforcement}
At each round, SIFR minimizes the signed loss of \cref{eq:naive-signed-mse} on rollouts from the reference model, subject to a budget $\delta$ on the drift from the reference.
Formally, given $\theta^{(k)}$, we obtain $\theta^{(k+1)}$ by solving
\begin{equation}
\begin{aligned}
\min_\theta
\EE_{\theta_{\mathrm{ref}}^{(k)}}\!\big[
\hs{A(X_1)}
\|\vth(X_t,t)-(X_1-X_0)\|^2
\big] \;~~~
\mathrm{s.t.}\;
\hr{
\EE_{\theta_{\mathrm{ref}}^{(k)}}\!\big[
D(\vth,v_{\mathrm{ref}}^{(k)})
\big]\le \delta},
\end{aligned}
\label{eq:constrained}
\end{equation}
where $D(\vth,v_{\mathrm{ref}}^{(k)})=\|\vth(X_t,t)-v_{\mathrm{ref}}^{(k)}(X_t,t)\|^2$ is the squared velocity drift as in \cref{eq:comparison-form}, $v_{\mathrm{ref}}^{(k)}=v_{\theta_{\mathrm{ref}}^{(k)}}$ is the reference model in round $k$, which also generates the rollouts, and $\delta$ is the drift budget.
Multiple constraints, one per reference such as $v_{\mathrm{ref}}^{\mathrm{fix}}$ and $v_{\mathrm{ref}}^{(k)}$, can be included; we use one here for simplicity.

The Lagrange dual of \cref{eq:constrained} yields a minimax problem $\max_{\lambda\ge0}\min_\theta \mathcal L^{(k)}(\theta, \lambda)$ with
\begin{equation}
\mathcal L^{(k)}(\theta, \lambda) =
\EE_{\theta_{\mathrm{ref}}^{(k)}}\!\big[\hs{A(X_1)}\|\vth(X_t,t)-(X_1-X_0)\|^2
+\hr{\lambda\big(D(\vth,v_{\mathrm{ref}}^{(k)})-\delta\big)}
\big]
\label{eq:signed-two-term}
\end{equation}
One can search for a saddle point of the Lagrangian by alternating updates of $\theta$ and $\lambda$, such as
$$
\theta^{(k+1)} \gets \theta^{(k)}
-\epsilon \nabla_{\theta}  \mathcal L^{(k)} (\theta^{(k)},
\lambda^{(k)}), ~~~~
\lambda^{(k+1)} \gets  \max(\lambda^{(k)} + \epsilon \nabla_\lambda \mathcal L^{(k)}(\theta^{(k+1)}, \lambda^{(k)}),~~ 0).
$$
where the $\theta$ update minimizes a regularized loss with coefficient $\lambda$, while the $\lambda$ update seeks to enforce the constraint by increasing $\lambda$ when the drift exceeds $\delta$ and decreasing it when it falls below.

\paragraph{Saddle Point}
For the fixed round-$k$ rollouts, a finite saddle point $(v^\star,\lambda^\star)$ of $\mathcal L^{(k)}$ satisfies
\[
v^\star(x,t)
=\frac{n(x,t)+\lambda^\star v_{\mathrm{ref}}^{(k)}(x,t)}
       {m(x,t)+\lambda^\star},
\qquad
m(x,t) + \lambda^\star >0,~~~~~\text{almost everywhere},
\]
together with the standard feasibility and complementary slackness conditions (\cref{app:smoothing-proof}).
Here $v^\star$ coincides with the Signed RF velocity with weight $A+\lambda^\star$, with $\lambda^\star$ adaptively adjusted to satisfy both the stability condition $m(x,t)+\lambda^\star\ge0$ and the drift constraint.
The stability condition is necessary because the multiplier must eliminate negative conditional curvature almost everywhere at a finite saddle point.
Intuitively, otherwise, minimizing the loss with respect to $v$ would drive its magnitude to infinity and violate the drift constraint, pushing $\lambda$ upward to restore $m(x,t)+\lambda\ge0$.

\subsection{Practical Algorithm}
\label{sec:controller}

\textbf{Reference Choice}\quad We recommend a single step constraint with $\theta_{\mathrm{ref}}^{(k)}=\theta_{\mathrm{ema}}^{(k)}$, an exponential moving average (EMA) of the parameters over past rounds with decay $\beta=0.5$.
In our experiments, this setting works well across scales, whereas RAM warms its decay up to $0.9$ and DiffusionNFT uses task-specific schedules that reach $0.5$ or $0.999$.
Optionally, a frozen reference $v_{\mathrm{ref}}^{\mathrm{fix}}$ can be added as a second constraint with its own budget and multiplier, which covers the full design space of \cref{tab:connections}, but it is not part of our default.

\textbf{Multiplier Update}\quad At the start of RL training, we initialize the budget to $\kappa$ times the RF loss of the pretrained model on its own rollouts, $\delta=\kappa\,\EE_{\theta^{(0)}}\!\|v_{\theta^{(0)}}(X_t,t)-(X_1-X_0)\|^2$.
The budget remains fixed throughout training.
After each round of model updates, we measure the drift $D^{(k)}=\EE_{\theta_{\mathrm{ref}}^{(k)}}\!\big[D(v_{\theta^{(k+1)}},v_{\mathrm{ref}}^{(k)})\big]$ and update its moving average $D_{\mathrm{ema}}^{(k)}$.
The multiplier is then updated by
\begin{equation}
\log\lambda^{(k+1)}
=\clip\Big(\!
\log\lambda^{(k)}+\tau\,\clip\Big(\frac{D_{\mathrm{ema}}^{(k)}}{\delta}-1,\,-1,\,1\Big),
\;\log\lambda_{\min},\;\log\lambda_{\max}
\Big).
\label{eq:ascent}
\end{equation}
Here $\tau$ sets the step size, and $\lambda_{\min}$ and $\lambda_{\max}$ bound $\lambda$.
Algorithm~\ref{alg:method} summarizes the procedure.

\begingroup
\begin{algorithm}[H]
\caption{SIFR with an EMA step constraint}
\label{alg:method}
\renewcommand{\algorithmicrequire}{\textbf{Require:}}
\begin{algorithmic}
\setlength{\itemsep}{1pt}
\REQUIRE Reward model $r$; trainable velocity field $v_{\theta^{(0)}}$.
\STATE Initialize $\theta_{\mathrm{ref}}^{(0)}\leftarrow\theta^{(0)}$, budget $\delta$, and $\lambda^{(0)}>0$.
\vspace{0.1em}
\FOR{$k=0,1,2,\ldots$}
\STATE Roll out $X_1\leftarrow\operatorname{ODESolve}(v_{\mathrm{ref}}^{(k)})$; evaluate $r(X_1)$ and compute $A(X_1)$.
\vspace{0.2em}
\STATE Update $v_{\theta^{(k+1)}}\leftarrow v_{\theta^{(k)}}$ with $S$ optimizer steps using\\
\vspace{0.1em}
\hspace*{1em}$\nabla_\theta~\EE_{\theta_{\mathrm{ref}}^{(k)}}\!\Big[\hs{A(X_1)}\|\vth(X_t,t)-(X_1-X_0)\|^2+\hr{\lambda^{(k)}D(\vth,v_{\mathrm{ref}}^{(k)})}\Big]$.
\vspace{0.1em}
\STATE Measure drift $D^{(k)}\leftarrow\EE_{\theta_{\mathrm{ref}}^{(k)}}\!\big[D(v_{\theta^{(k+1)}},v_{\mathrm{ref}}^{(k)})\big]$; update $D_{\mathrm{ema}}^{(k)}$.
\vspace{0.1em}
\STATE Update multiplier $\lambda^{(k+1)}\leftarrow\lambda^{(k)}$ using \cref{eq:ascent}.
\vspace{0.1em}
\STATE Update reference $\theta_{\mathrm{ref}}^{(k+1)}\leftarrow\beta\,\theta_{\mathrm{ref}}^{(k)}+(1-\beta)\theta^{(k+1)}$.
\vspace{0.1em}
\ENDFOR
\end{algorithmic}
\end{algorithm}
\endgroup

\input{sections/experiments}

\section{Conclusion}
\label{sec:conclusion}
We presented SIFR, which replaces the reward weights of the rectified flow loss by signed advantages and constrains the velocity drift from a reference model, with a Lagrange multiplier that adapts the regularization strength to a prescribed drift budget.
The signed advantages supply the unfitting signal that positive reward weights lack, and the adaptive constraint keeps training on the signed loss stable.
Existing flow RL methods turn out to minimize the same signed loss with different regularization choices.
Validated on SD3.5 Medium and FLUX.2 Klein and scaled to Cosmos 3 Super Text2Image, SIFR matches or exceeds DiffusionNFT and RAM across all 12 backbone--reward pairs, reaches their peak preference rewards about 2--4$\times$ sooner, keeps improving where they plateau, and carries its gains to held-out rewards and prompt sets.

\bibliography{references}
\bibliographystyle{iclr2027_conference}

\appendix

\clearpage
\begingroup
\raggedbottom \setlength{\intextsep}{6pt}
\input{sections/appendix_qualitative}
\clearpage
\endgroup

\section{Related work}
\label{app:related}

\Cref{sec:weighted} connects existing flow RL methods through a regularized signed loss.
Here we discuss how SIFR relates to flow modeling, reward-based post-training, and constrained optimization.

\paragraph{Rectified Flow and Signed RF}
Rectified flow~\citep{liu2023flow,liu2022rectified,liu2024flowbook,hu2025improving,liao2026momentum,su2026rectified} and flow matching~\citep{lipman2023flow,albergo2023stochastic} regress the interpolation velocity by least squares.
The conditional-mean velocity transports the source distribution to the data distribution.
Signed RF~\citep{liao2026signed} extends this construction to signed target measures, and \cref{sec:weighted} connects the signed loss to its sampling dynamics.
SIFR regularizes the signed velocity through a drift constraint.
\Cref{app:smoothing-proof} derives the resulting velocity and shows when the multiplier acts as an additive shift of the advantage weights.

\paragraph{RL for Diffusion and Flow Models}
Policy-gradient methods treat the sampler as a Markov decision process and need the transition likelihoods of a stochastic sampler~\citep{black2024ddpo,fan2023dpok,liu2025flowgrpo,xue2025dancegrpo}.
Reward backpropagation instead differentiates the reward through the sampling chain~\citep{clark2024draft,prabhudesai2023alignprop}.
SIFR uses velocity regression, which requires neither transition likelihoods nor reward gradients.
Related approaches include DiffusionNFT~\citep{zheng2026diffusionnft}, ORW-CFM-W2~\citep{fan2025orw}, AWM~\citep{xue2025awm}, Reward Score Matching~\citep{lee2026rsm}, RAM~\citep{bergmeister2026ram}, Adjoint Matching~\citep{domingoenrich2024adjoint}, AdvantageFlow~\citep{kveton2026advantageflow}, and FlowAWR~\citep{fu2026flowawr}.
For the methods in \cref{tab:connections}, \cref{app:implicit-proof} derives gradient-equivalent objectives of the form in \cref{eq:comparison-form}; the signed weights of DiffusionNFT and RAM appear only in their update gradients, not in their original losses.
The positive starting point of this family is reward-weighted regression~\citep{peters2007rwr,peng2019awr}, whose exponential weights solve KL-regularized reward maximization in closed form~\citep{todorov2007linearly,kappen2005path}, the objective also used in RLHF~\citep{ouyang2022rlhf}.

\paragraph{Negative Information in Generation and Unlearning}
Classifier-free guidance~\citep{ho2022cfg}, autoguidance~\citep{karras2024autoguidance}, and momentum guidance~\citep{liao2026momentum} extrapolate the sampling velocity away from a weaker prediction, using negative information at inference time; SIFR instead applies the negative weight during training to below-average rollouts.
Concept erasure~\citep{gandikota2023erasing} fine-tunes a model toward a negative-guidance target, and dynamic negative guidance~\citep{koulischer2025dynamic} makes the negative scale state-dependent.
In language models, gradient ascent on the negative log-likelihood of examples to be unlearned can collapse the model~\citep{jang2023knowledge,maini2024tofu}.
Negative preference optimization reduces this ascent as the examples become less likely relative to a reference model~\citep{zhang2024npo}.
SIFR controls signed velocity updates by constraining their drift from a reference model.

\paragraph{Constrained Policy Optimization}
Trust-region methods constrain or penalize the KL divergence between consecutive policies~\citep{schulman2015trpo,schulman2017ppo}, including under explicit constraints~\citep{achiam2017cpo}, and Lagrangian methods for constrained RL update a multiplier from the measured constraint violation~\citep{tessler2019rcpo,stooke2020pid}.
SIFR shares this constrained spirit but addresses a different problem: a flow model has no explicit likelihood, so a KL constraint between consecutive policies cannot be imposed directly.
SIFR uses the mean squared drift of the velocity field from a reference model as a proxy, and the multiplier is the coefficient of this drift penalty in \cref{eq:signed-two-term}.

\section{Derivations}
\label{app:proofs}

We derive the collapse of repeated tilting (\cref{app:repeated-tilting}), the conditional geometry of the signed loss and its Signed RF interpretation (\cref{app:geometry-proof,app:signed-weight,app:moment-regression}), the gradient-equivalent objectives of the methods in \cref{tab:connections} (\cref{app:implicit-proof}), and the stationary velocity under the drift constraint (\cref{app:smoothing-proof}).
For the velocity-field derivations, we hold the sampling distribution fixed.

\subsection{Repeated reward tilting}
\label{app:repeated-tilting}

Repeated reward tilting amplifies relative densities exponentially.
Let $\pi_1^{(k)}$ denote the generated distribution after $k$ rounds, starting from $\pi_1^{(0)}$.
For a fixed reward function $r(x)\ge0$, suppose each round realizes the exact RF tilt in \cref{eq:weighted-loss}.
The generated distribution then satisfies
\[
\pi_1^{(k+1)}(x)
=\frac{r(x)\pi_1^{(k)}(x)}{\int r(y)\pi_1^{(k)}(y)\,\mathrm{d}y}.
\]
Repeated substitution gives
\[
\pi_1^{(k)}(x)
=\frac{r(x)^k\pi_1^{(0)}(x)}{\int r(y)^k\pi_1^{(0)}(y)\,\mathrm{d}y}.
\]
For points $x,y$ with positive rewards and positive initial densities, the common normalizer cancels:
\[
\frac{\pi_1^{(k)}(x)}{\pi_1^{(k)}(y)}
=\left(\frac{r(x)}{r(y)}\right)^k
\frac{\pi_1^{(0)}(x)}{\pi_1^{(0)}(y)}.
\]
Thus, if $r(x)>r(y)>0$, the relative density at $x$ grows exponentially with the number of rounds.

\Cref{fig:collapse-strip} illustrates the collapse of positive-weight training and of unconstrained signed training on a fixed prompt and noise seed.

\begin{figure}[!t]
\centering
\input{figures/collapse_strip}
\caption{Training trajectories on SD3.5 Medium PickScore. Top: positive weights mapped to $[0,1]$, without a constraint; middle: signed weighting without a constraint; bottom: SIFR. Images share a prompt and noise seed; labels give PickScore. Below: training-batch reward, each run's training loss (symmetric log scale), and gradient norm.}
\label{fig:collapse-strip}
\end{figure}

\Needspace{10\baselineskip}
\subsection{Conditional geometry of signed regression}
\label{app:geometry-proof}

The sign of the conditional mean weight determines whether the stationary velocity minimizes or maximizes the conditional loss.
For a general signed weight $w$, write $\mathcal L_w(v)=\EE\big[w(X_1)\|v(X_t,t)-(X_1-X_0)\|^2\big]$ and define
\begin{equation}
m(x,t)=\EE[w(X_1)\mid X_t=x],\qquad
n(x,t)=\EE[w(X_1)(X_1-X_0)\mid X_t=x].
\label{eq:conditional-weight}
\end{equation}
The choice $w=A$ recovers the coefficients $m(\cdot,t)$ and $n(\cdot,t)$ used in \cref{sec:weighted}.
At a fixed input $(x,t)$, the signed loss is the quadratic
\begin{equation}
\begin{aligned}
&\EE\big[w(X_1)\|v(x,t)-(X_1-X_0)\|^2\mid X_t=x\big]\\
&\qquad=m(x,t)\|v(x,t)\|^2-2v(x,t)^\top n(x,t)+\mathrm{const}.
\end{aligned}
\label{eq:conditional-quadratic}
\end{equation}

\begin{proposition}[Stationary velocity of signed regression]
\label{prop:signed-geometry}
Where $m(x,t)\ne0$, the unique conditional stationary point is
\begin{equation}
v^w(x,t)=\frac{n(x,t)}{m(x,t)}
=\frac{\EE[w(X_1)(X_1-X_0)\mid X_t=x]}{\EE[w(X_1)\mid X_t=x]}.
\label{eq:signed-weighted-loss}
\end{equation}
It is a strict conditional minimum if $m(x,t)>0$ and a strict conditional maximum if $m(x,t)<0$.
In the latter case the conditional loss is unbounded below.
If $m(X_t,t)\ne0$ almost surely and takes both signs with positive probability, then $v^w$ is a saddle point of $\mathcal L_w$.
\end{proposition}

\begin{proof}[Derivation of \Cref{prop:signed-geometry}]
For $m(x,t)\ne0$, completing the square in \cref{eq:conditional-quadratic} gives
\[
m(x,t)\|v(x,t)-v^w(x,t)\|^2+\mathrm{const}.
\]
The sign of $m(x,t)$ therefore determines whether $v^w(x,t)$ is a minimum or a maximum, and a negative coefficient makes the conditional loss unbounded below.
At $v=v^w$, expanding the full loss gives
\[
\mathcal L_w(v^w+h)-\mathcal L_w(v^w)
=\EE\big[m(X_t,t)\|h(X_t,t)\|^2\big],
\]
since stationarity eliminates the linear term in $h$.
Perturbations supported where $m>0$ increase the loss, while those supported where $m<0$ decrease it, giving a saddle when both signs occur.
Scaling a perturbation in the negative region makes the loss unbounded below.
\end{proof}

\subsection{Signed reward tilting}
\label{app:signed-weight}

The stationary velocity $v^w(x,t)$ has a sampling interpretation through Signed RF~\citep{liao2026signed}.
For a signed weight with $Z_w=\EE[w(X_1)]>0$, write $w_+(x)=\max\{w(x),0\}$ and $w_-(x)=\max\{-w(x),0\}$.
Let $a=\EE[w_+(X_1)]$ and $b=\EE[w_-(X_1)]$, so $Z_w=a-b$.
When $b>0$, the normalized positive and negative parts $\pi_1^\pm(x)=w_\pm(x)\pi_1(x)/\EE[w_\pm(X_1)]$ give
\[
\frac{w\pi_1}{Z_w}
=\frac{a}{a-b}\pi_1^+-\frac{b}{a-b}\pi_1^-
=(1+\alpha)\pi_1^+-\alpha\pi_1^-,
\qquad \alpha=\frac{b}{a-b}.
\]
Let $\pi_t^\pm$ and $v^\pm(x,t)$ denote the RF marginals and velocities for these two distributions, with the same independent source $\pi_0$.
Because $X_0$ and $t$ are independent of $X_1$, $\pi_t^\pm(x)=\pi_t(x)\,\EE[w_\pm(X_1)\mid X_t=x]/\EE[w_\pm(X_1)]$ and $\pi_t^\pm(x)v^\pm(x,t)=\pi_t(x)\,\EE[w_\pm(X_1)(X_1-X_0)\mid X_t=x]/\EE[w_\pm(X_1)]$.
Taking the signed combination, the reweighted marginal is $\pi_t^{\mathrm{sign}}(x)=(1+\alpha)\pi_t^+(x)-\alpha\pi_t^-(x)=\pi_t(x)m(x,t)/Z_w$ and the flux is $(1+\alpha)\pi_t^+(x)v^+(x,t)-\alpha\pi_t^-(x)v^-(x,t)=\pi_t(x)n(x,t)/Z_w$.
Their ratio is the stationary velocity in \cref{eq:signed-weighted-loss}:
\begin{equation}
v^w(x,t)=
\frac{(1+\alpha)\pi_t^+(x)v^+(x,t)-\alpha\pi_t^-(x)v^-(x,t)}{(1+\alpha)\pi_t^+(x)-\alpha\pi_t^-(x)}
=\frac{\EE[w(X_1)(X_1-X_0)\mid X_t=x]}{\EE[w(X_1)\mid X_t=x]}.
\label{eq:signed-to-weighted}
\end{equation}
Endpoint weighting leaves the source marginal $\pi_0$ unchanged, and $\pi_t^{\mathrm{sign}}$ satisfies the signed continuity equation under $v^w(x,t)$.
Under the assumptions of the Signed RF sampling theorem, the ODE driven by $v^w$ generates samples from the reachable subset of the positive region $\{x:w(x)>0\}$ and avoids the negative region, as in \cref{sec:signed-regression}.
For the centered advantage $A$ of \cref{eq:naive-signed-mse}, $Z_A=0$, so this interpretation applies to the shifted weight $A+\lambda$ with $\lambda>0$ (\cref{prop:shifted-weight}) rather than to $A$ itself.

\subsection{Learning the numerator and denominator separately}
\label{app:moment-regression}

The conditional moments in \cref{eq:conditional-weight} can be learned separately without minimizing the signed loss.
Let $n_\xi(x,t)$ predict the vector $n(x,t)$ and let $m_\psi(x,t)$ predict the scalar $m(x,t)$.
They can be learned by two ordinary MSE regressions on the same interpolation pairs:
\begin{equation}
\begin{aligned}
\min_\xi\;&\EE\big[\|n_\xi(X_t,t)-w(X_1)(X_1-X_0)\|^2\big],\\
\min_\psi\;&\EE\big[(m_\psi(X_t,t)-w(X_1))^2\big].
\end{aligned}
\label{eq:moment-regression-losses}
\end{equation}
The sign of $w$ is carried by the regression targets; every loss term remains nonnegative.

Minimizing over unrestricted prediction functions recovers the conditional moments in \cref{eq:conditional-weight}:
\begin{equation}
n^\star(x,t)=n(x,t),
\qquad
m^\star(x,t)=m(x,t).
\label{eq:moment-regression-optima}
\end{equation}
Their ratio $n(x,t)/m(x,t)$ is $v^w(x,t)$ wherever $m(x,t)\ne0$, and sampling uses the estimate $n_\xi(x,t)/m_\psi(x,t)$.

\input{sections/appendix_method_comparison}

\subsection{Constraint smoothing and the active multiplier}
\label{app:smoothing-proof}

The drift penalty adds $\lambda$ to the conditional quadratic coefficient of the signed loss.
Fix round $k$, the reference $v_{\mathrm{ref}}^{(k)}(x,t)$, and the law $\pi_1^{(k)}$ of its rollouts $X_1=\operatorname{ODESolve}(v_{\mathrm{ref}}^{(k)})$; all expectations in this subsection are $\EE_{\theta_{\mathrm{ref}}^{(k)}}$ as in \cref{eq:constrained}.
Set $w=A$ in \cref{eq:conditional-weight} under this law, so that $m(x,t)=\EE_{\theta_{\mathrm{ref}}^{(k)}}[A(X_1)\mid X_t=x]$ and $n(x,t)=\EE_{\theta_{\mathrm{ref}}^{(k)}}[A(X_1)(X_1-X_0)\mid X_t=x]$.
We write $\mathcal L^{(k)}(v,\lambda)$ for the Lagrangian in \cref{eq:signed-two-term} optimized over velocity fields, with the sampling law held fixed, and $\mathcal L_A(v)=\EE_{\theta_{\mathrm{ref}}^{(k)}}\big[A(X_1)\|v(X_t,t)-(X_1-X_0)\|^2\big]$ for the signed loss, so that $\mathcal L^{(k)}(v,\lambda)=\mathcal L_A(v)+\lambda\big(\EE_{\theta_{\mathrm{ref}}^{(k)}}[D(v,v_{\mathrm{ref}}^{(k)})]-\delta\big)$.
For fixed $\lambda\ge0$, its conditional stationary velocity is
\begin{equation}
v^\lambda(x,t)
=\frac{n(x,t)+\lambda v_{\mathrm{ref}}^{(k)}(x,t)}{m(x,t)+\lambda},
\label{eq:smoothed-field}
\end{equation}
where $m(x,t)+\lambda\ne0$.

\paragraph{Saddle-Point Conditions}
At a saddle point $(v^\star,\lambda^\star)$, conditional optimality requires
\[
\big(m(x,t)+\lambda^\star\big)v^\star(x,t)
=n(x,t)+\lambda^\star v_{\mathrm{ref}}^{(k)}(x,t),
\qquad m(x,t)+\lambda^\star\ge0.
\]
Writing $D^\star=\EE_{\theta_{\mathrm{ref}}^{(k)}}\!\big[D(v^\star,v_{\mathrm{ref}}^{(k)})\big]$, feasibility and complementary slackness require
\[
D^\star\le \delta,\qquad \lambda^\star\ge0,\qquad
\lambda^\star(D^\star-\delta)=0.
\]

\begin{proposition}[Regularized signed regression]
\label{prop:smoothed}
Where $m(x,t)+\lambda>0$, the conditional Lagrangian is strictly convex and its minimizer is \cref{eq:smoothed-field}.
If $m(x,t)+\lambda>0$ holds almost everywhere under the rollout law and $v^\lambda$ satisfies the drift constraint, with equality whenever $\lambda>0$, then $v^\lambda$ solves the constrained problem \cref{eq:constrained}.
\end{proposition}

\begin{proof}[Derivation of \Cref{prop:smoothed}]
Conditioning the Lagrangian on $(X_t,t)$ and collecting its terms in $v(x,t)$ gives
\[
(m(x,t)+\lambda)\|v(x,t)\|^2
-2v(x,t)^\top\big(n(x,t)+\lambda v_{\mathrm{ref}}^{(k)}(x,t)\big)+\mathrm{const}.
\]
Completing the square gives
\[
(m(x,t)+\lambda)\|v(x,t)-v^\lambda(x,t)\|^2+\mathrm{const},
\]
which is minimized at $v^\lambda(x,t)$ when $m(x,t)+\lambda>0$.
If the resulting field satisfies the constraint and complementary slackness, then for any feasible $v$,
\[
\mathcal L_A(v)\ge\mathcal L^{(k)}(v,\lambda)
\ge\mathcal L^{(k)}(v^\lambda,\lambda)=\mathcal L_A(v^\lambda),
\]
where the first inequality uses $\lambda\ge0$ and feasibility of $v$, the second that $v^\lambda$ minimizes the Lagrangian, and the equality uses complementary slackness.
\end{proof}

\paragraph{Relation to the Signed Velocity}
Where $m(x,t)\ne0$, \cref{eq:smoothed-field} can be written as
\begin{equation}
v^\lambda(x,t)-v_{\mathrm{ref}}^{(k)}(x,t)
=\frac{m(x,t)}{m(x,t)+\lambda}
\big(v^A(x,t)-v_{\mathrm{ref}}^{(k)}(x,t)\big).
\label{eq:smoothing-factor}
\end{equation}
When $m(x,t)>0$, the factor lies in $[0,1]$ and shrinks the signed velocity toward the reference.
When $m(x,t)<0$ but $m(x,t)+\lambda>0$, the factor is negative, so $v^\lambda$ departs from the reference in the direction opposite to $v^A$, while the conditional quadratic coefficient $m(x,t)+\lambda$ is positive.

\paragraph{Choosing a Multiplier That Meets the Budget}
For $\lambda$ with $m(x,t)+\lambda>0$ almost everywhere, the squared drift of \cref{eq:smoothed-field} is
\begin{equation}
D_\lambda
=\EE_{\theta_{\mathrm{ref}}^{(k)}}\!\left[
\frac{\|n(X_t,t)-m(X_t,t)v_{\mathrm{ref}}^{(k)}(X_t,t)\|^2}{(m(X_t,t)+\lambda)^2}
\right].
\label{eq:regular-drift-root}
\end{equation}
Increasing $\lambda$ can only reduce this drift.
The multiplier update in \cref{eq:ascent} searches for the root $D_\lambda=\delta$ from the measured drift.

\begin{proposition}[Additive weight shift under an RF reference]
\label{prop:shifted-weight}
Let $v^{\mathrm{RF}}(x,t)=\EE_{\theta_{\mathrm{ref}}^{(k)}}[X_1-X_0\mid X_t=x]$ be the ordinary RF velocity under the same rollout law.
For a fixed scalar $\lambda\ge0$ and $v_{\mathrm{ref}}^{(k)}(x,t)=v^{\mathrm{RF}}(x,t)$, the Lagrangian equals, up to a constant independent of $v$,
\begin{equation}
\EE_{\theta_{\mathrm{ref}}^{(k)}}\!\big[(A(X_1)+\lambda)\|v(X_t,t)-(X_1-X_0)\|^2\big].
\label{eq:shifted-loss}
\end{equation}
Where $m(x,t)+\lambda\ne0$, the stationary velocity of \cref{eq:shifted-loss} is $v^{A+\lambda}(x,t)=\big(n(x,t)+\lambda v^{\mathrm{RF}}(x,t)\big)/\big(m(x,t)+\lambda\big)$, the conditional ratio of \cref{eq:signed-weighted-loss} with weight $A+\lambda$.
If $\EE_{\theta_{\mathrm{ref}}^{(k)}}[A(X_1)]+\lambda>0$, the corresponding endpoint target, an RF tilt when $A+\lambda\ge0$ and a Signed RF target otherwise, is
\begin{equation}
\pi_1^{(k),\lambda}(x)=\frac{A(x)+\lambda}{\EE_{\theta_{\mathrm{ref}}^{(k)}}[A(X_1)]+\lambda}\,\pi_1^{(k)}(x).
\label{eq:shifted-endpoint}
\end{equation}
For a general reference $v_{\mathrm{ref}}^{(k)}$, \cref{eq:smoothed-field} differs from this shifted-weight velocity by
\begin{equation}
v^\lambda(x,t)-v^{A+\lambda}(x,t)
=\frac{\lambda\big(v_{\mathrm{ref}}^{(k)}(x,t)-v^{\mathrm{RF}}(x,t)\big)}{m(x,t)+\lambda}.
\label{eq:anchor-bias}
\end{equation}
\end{proposition}

\begin{proof}[Derivation of \Cref{prop:shifted-weight}]
By the definition of $v^{\mathrm{RF}}$, the residual $(X_1-X_0)-v^{\mathrm{RF}}(X_t,t)$ has zero conditional mean given $X_t$.
The cross term therefore vanishes in the expansion
\[
\begin{aligned}
\EE_{\theta_{\mathrm{ref}}^{(k)}}\!\|v(X_t,t)-(X_1-X_0)\|^2
&=\EE_{\theta_{\mathrm{ref}}^{(k)}}\!\|v(X_t,t)-v^{\mathrm{RF}}(X_t,t)\|^2\\
&\quad+\EE_{\theta_{\mathrm{ref}}^{(k)}}\!\|v^{\mathrm{RF}}(X_t,t)-(X_1-X_0)\|^2.
\end{aligned}
\]
When $v_{\mathrm{ref}}^{(k)}(x,t)=v^{\mathrm{RF}}(x,t)$, substituting this identity into the drift penalty yields \cref{eq:shifted-loss}, up to a constant independent of $v$.
Normalizing its endpoint weight $A+\lambda$ gives \cref{eq:shifted-endpoint}.
The corresponding stationary velocity has numerator $n(x,t)+\lambda v^{\mathrm{RF}}(x,t)$ and denominator $m(x,t)+\lambda$, so subtracting it from \cref{eq:smoothed-field} gives the reference correction in \cref{eq:anchor-bias}.
\end{proof}

\paragraph{Endpoint Weights}
At $t=1$, $m(x,1)=A(x)$, so the stability condition $m(x,t)+\lambda\ge0$ at the endpoint reads $A(x)+\lambda\ge0$.
Under the RF reference, \cref{eq:shifted-endpoint} is then an ordinary RF tilt with positive weights.
Since $m(x,1)=A(x)$, the stability condition at a finite saddle point requires $\lambda^\star\ge-\operatorname{ess\,inf}A$, so $A+\lambda^\star\ge0$ almost everywhere: under the RF reference the saddle point is an ordinary positive tilt, and the negative weights act during the approach to it; reaching it also requires $\lambda_{\max}$ in \cref{eq:ascent} to exceed $-\operatorname{ess\,inf}A$.

\input{sections/appendix_results}

\end{document}

%% file: sections/experiments.tex
\section{Experiments}
\label{sec:exp}

We validate SIFR on SD3.5 Medium (2.5B)~\citep{esser2024sd3,stabilityai2024sd35medium} and FLUX.2 Klein (9B)~\citep{bfl2026flux2klein}, and scale it to Cosmos 3 Super Text2Image (64B)~\citep{nvidia2026cosmos3}, comparing against DiffusionNFT~\citep{zheng2026diffusionnft} and RAM~\citep{bergmeister2026ram} throughout.
The experiments examine whether SIFR optimizes the training reward faster and to a higher ceiling (\cref{sec:exp-main}) and whether the gains survive held-out rewards and prompt sets (\cref{sec:exp-generalization}), followed by ablation studies (\cref{sec:exp-collapse}).

\subsection{Experimental setup}
\label{sec:exp-setup}

\paragraph{Rewards and Prompt Sets}
We optimize four reward functions in separate runs.
For preference alignment, we consider two settings: \emph{single-reward training} with PickScore~\citep{kirstain2023pickscore}, and \emph{multi-reward training} with the sum of PickScore, HPSv2.1~\citep{wu2023hpsv2}, and ClipScore~\citep{radford2021clip}.
Both settings use the Pick-a-Pic training prompts~\citep{kirstain2023pickscore}.
For the OCR task and compositional generation with GenEval~\citep{ghosh2023geneval}, we follow the reward and prompt pipelines of Flow-GRPO~\citep{liu2025flowgrpo}.
To assess performance beyond the training rewards, we report ImageReward~\citep{xu2023imagereward}, HPSv3~\citep{ma2025hpsv3}, UnifiedReward-2.0 (UniRwd2)~\citep{wang2025unifiedreward,codegoat2026unifiedreward2}, and Aesthetic Score~\citep{schuhmann2022aesthetic} as held-out rewards.
We further evaluate on held-out prompts from each task, the $600$ UniGenBench prompts~\citep{wang2025unigenbench} augmented with $570$ physical prompts from Cosmos 3~\citep{nvidia2026cosmos3}, and $200$ DrawBench prompts~\citep{saharia2022imagen} with $5$ images per prompt.

\paragraph{Training Configuration}
We fine-tune all models with LoRA~\citep{hu2022lora} (rank $32$, $\alpha=64$), using AdamW~\citep{loshchilov2019adamw} at a learning rate of $3\times10^{-4}$.
At each RL round, we sample $48$ prompt groups with a group size of $24$.
Rollout images are generated at $512\times512$ for SD3.5 Medium and FLUX.2 Klein, and $1024\times1024$ for Cosmos 3 Super Text2Image.
For experiments on SD3.5 Medium and FLUX.2 Klein, SIFR and RAM use an Euler sampler with $20$ steps, while DiffusionNFT uses its original DPM-Solver++ sampler with its per-task step count. For experiments on Cosmos 3 Super Text2Image, all three methods use an Euler sampler with $35$ steps.

\input{style/findings}

\subsection{Optimizing the Training Rewards}
\label{sec:exp-main}

\experimentfinding{SIFR more effectively optimizes the training rewards, achieving higher rewards across model scales and continuing to improve with longer training.}

\paragraph{Preference Rewards}
We first optimize the preference rewards, in the single-reward and multi-reward settings, on all three backbones.
\Cref{fig:preference-training} tracks the optimized reward on held-out prompts: in all six panels, SIFR climbs faster than DiffusionNFT and RAM and ends higher within $500$ rounds.
The gap widens with longer training.
On Cosmos 3 Super Text2Image, extending single-reward training to $2000$ rounds leaves both baselines on a plateau near $24.4$, which SIFR passes in roughly one quarter of their rounds and then keeps climbing (\cref{fig:teaser}a); the same continued improvement holds under multi-reward training.
\Cref{tab:main,tab:single-reward} report the $500$-round scores: when PickScore, HPSv2.1, and ClipScore are optimized jointly, SIFR obtains the best value on every backbone.
\vspace{-0.6em}

\input{figures/preference_results}
\paragraph{Rule-Based Rewards}
\Cref{tab:rule-based} reports the scores on text rendering (OCR) and compositional generation (GenEval) within $500$ rounds and, since rule-based rewards saturate quickly, the first round at which each method reaches $0.97$.
All three methods approach saturation, and SIFR holds the highest score in five of the six backbone--task comparisons.
By rounds to $0.97$, SIFR is never slower than either baseline, and on SD3.5 Medium OCR and both Cosmos 3 Super Text2Image tasks it needs about half the rounds of the faster baseline.

\paragraph{Long-Caption Training}
Cosmos 3 Super Text2Image natively supports the long, detailed prompts produced by prompt upsampling~\citep{nvidia2026cosmos3}.
Such prompts pin down the image more tightly, so rollouts of the same prompt are less diverse and the reward signal is weaker.
Training on $57{,}918$ long captions, SIFR still leads both baselines on held-out long captions (\cref{fig:longprompt}).

\input{figures/rule_based_results}
\subsection{Transferability across Held-Out Rewards and Prompts}
\label{sec:exp-generalization}

\experimentfinding{Across SD3.5 Medium, FLUX.2 Klein, and Cosmos 3 Super Text2Image, SIFR's gains are shared by most samples and extend to held-out rewards and new prompt sets.}

\paragraph{Per-Prompt Reward Gains}
For each held-out prompt and noise seed, we compare the images generated by the three methods.
SIFR scores above both DiffusionNFT and RAM in $79.9\%$--$92.4\%$ of the cases, on every backbone and under both training rewards (\cref{fig:reward-gain-distributions,fig:reward-gain-single}).

\begin{figure}[!b]
\centering
\input{figures/reward_gain_multi.tex}
\caption{\textbf{Per-prompt reward gains after multi-reward training.} Left: each point is one held-out Pick-a-Pic prompt and noise seed, placed by SIFR's gain over DiffusionNFT (horizontal) and over RAM (vertical); blue points improve over both, and the percentage is their share. Right: reward distribution of the samples on Cosmos 3 Super T2I; ticks below the axis mark the means. \Cref{fig:reward-gain-single} shows single-reward training.}
\label{fig:reward-gain-distributions}
\end{figure}

\paragraph{Held-Out Rewards and Prompt Sets}
After multi-reward training, SIFR has the highest ImageReward and HPSv3 on all three backbones (\cref{tab:main}), and on Cosmos 3 Super Text2Image it leads on all four held-out evaluators.
On new prompt sets, UniGenBench augmented with physical prompts, DrawBench, and $2048$ held-out long captions, the same checkpoints give SIFR the highest score on all seven reported rewards on Cosmos 3 Super Text2Image (\cref{tab:cosmos-generalization}), and the highest PickScore and Aesthetic Score on UniGenBench and DrawBench for every backbone and both training rewards (\cref{tab:crossset-unigenbench,tab:crossset-drawbench}).

\input{tables/prompt_sets_cosmos}
\subsection{Ablation studies}
\label{sec:exp-collapse}

\experimentfinding{Negative advantages are necessary even under the drift constraint, the budget-driven multiplier ends above every fixed coefficient, and the gains hold with and without CFG.}

\paragraph{Negative Advantages}
\Cref{fig:ablation}a replaces $A(X_1)$ with $\max\{A(X_1),0\}$ on SD3.5 Medium and keeps everything else fixed.
Both positive-weight variants collapse, with or without the drift penalty, while the signed loss trains stably: regularization alone cannot replace the unfitting signal.

\begin{figure}[tb]
\centering
\resizebox{\textwidth}{!}
{\input{figures/ablation_guidance.tex}}
\caption{\textbf{Ablations and evaluation guidance.} (a) SD3.5 Medium PickScore training with signed or positive weights under the drift budget. (b) FLUX.2 Klein multi-reward training with the updated multiplier and five fixed coefficients. (c,d) Multi-reward and OCR scores of the Cosmos 3 Super Text2Image checkpoints evaluated with (filled) and without (open) CFG.}
\label{fig:ablation}
\label{fig:guidance}
\end{figure}

\Needspace{12\baselineskip}
\begingroup
\setlength{\columnsep}{12pt}
\begin{wrapfigure}{r}{168pt}
\vspace{-12pt}%
\centering
\input{figures/feedback_control.tex}
\caption{Feedback control on FLUX.2 Klein.}
\label{fig:feedback-inline}
\end{wrapfigure}
\paragraph{Multiplier Update}
\Cref{fig:ablation}b compares the multiplier updated by \cref{eq:ascent} against five fixed coefficients between $0.01$ and $0.5$ on FLUX.2 Klein multi-reward training.
At round $500$, the updated multiplier ends above every fixed choice; \cref{fig:feedback-inline} shows how it moves under feedback control during one run.

\paragraph{Evaluation Guidance}
\Cref{fig:guidance}c,d evaluate every Cosmos 3 Super Text2Image checkpoint with and without classifier-free guidance (CFG) on the same prompts and noise seeds.
SIFR leads on both rewards in both settings, and even without CFG it exceeds the guided baselines.

\par
\ifnum\value{WF@wrappedlines}>1
  \vspace{\dimexpr\value{WF@wrappedlines}\baselineskip-\baselineskip\relax}
\fi
\WFclear
\endgroup

%% file: figures/preference_results.tex
\begin{figure}[H]
\centering
\setlength{\parskip}{0pt}
\setlength{\intextsep}{0pt}
\setlength{\abovecaptionskip}{4pt}
\setlength{\belowcaptionskip}{0pt}
\input{tables/multi_reward}
\par\vspace{10pt}
\resizebox{\textwidth}{!}{\input{figures/preference_training.tex}}
\par\vspace{-8pt}
\caption{\textbf{Preference-reward training across model scales.} Test on held-out prompts.
}
\label{fig:preference-training}
\par\vspace{-12pt}
\end{figure}

%% file: tables/multi_reward.tex
\begin{table}[H]
\centering
\begingroup\small
\setlength{\tabcolsep}{2.1pt}
\renewcommand{\arraystretch}{1.06}
\newsavebox{\preferencebox}
\sbox{\preferencebox}{%
\begin{tabular}{@{}l@{\hspace{7pt}}C{38pt}!{\hspace{4pt}}*{2}{C{38pt}}!{\hspace{4pt}}*{4}{C{38pt}}@{}}
\toprule[0.7pt]
\multicolumn{1}{@{}c@{\hspace{7pt}}}{\multirow{2}{*}[-2.5pt]{\footnotesize Method}} & \multicolumn{3}{c}{\makebox[0pt]{\footnotesize Training rewards}} & \multicolumn{4}{c}{\footnotesize Held-out rewards} \\
\cmidrule(r){2-4}\cmidrule(l){5-8}
& \footnotesize PickScore & \footnotesize HPSv2.1 & \footnotesize ClipScore & \footnotesize ImgRwd & \footnotesize HPSv3 & \footnotesize UniRwd2 & \footnotesize Aesthetic \\
\specialrule{0.35pt}{2.25pt}{2.75pt}
\textit{SD3.5 Medium}\,{\scriptsize\textcolor{black!55}{(2.5B)}} & 21.78 & 0.295 & 0.289 & 1.02 & 9.18 & 3.194 & 5.96 \\
\hspace{0.6em}+\,DiffusionNFT & 23.06 & 0.348 & 0.318 & 1.56 & 10.52 & 3.249 & \underline{6.12} \\
\hspace{0.6em}+\,RAM & \underline{23.37} & \underline{0.361} & \underline{0.328} & \underline{1.61} & \underline{11.92} & \textbf{3.294} & 6.08 \\
\hspace{0.6em}+\,SIFR & \textbf{23.76} & \textbf{0.374} & \textbf{0.339} & \textbf{1.67} & \textbf{12.12} & \underline{3.290} & \textbf{6.15} \\
\arrayrulecolor{black!22}\specialrule{0.25pt}{2.25pt}{2.75pt}\arrayrulecolor{black}
\textit{FLUX.2 Klein}\,{\scriptsize\textcolor{black!55}{(9B)}} & 21.82 & 0.299 & 0.290 & 1.18 & 9.47 & 3.249 & 5.86 \\
\hspace{0.6em}+\,DiffusionNFT & \underline{23.69} & \underline{0.364} & \underline{0.328} & \underline{1.63} & \underline{12.20} & \underline{3.290} & \underline{6.21} \\
\hspace{0.6em}+\,RAM\smash{\textsuperscript{\ensuremath{\dagger}}} & 23.26 & 0.351 & 0.315 & 1.57 & 11.64 & \textbf{3.304} & \textbf{6.23} \\
\hspace{0.6em}+\,SIFR & \textbf{24.00} & \textbf{0.385} & \textbf{0.336} & \textbf{1.67} & \textbf{12.86} & \underline{3.290} & \textbf{6.23} \\
\arrayrulecolor{black!22}\specialrule{0.25pt}{2.25pt}{2.75pt}\arrayrulecolor{black}
\textit{Cosmos 3 Super T2I}\,{\scriptsize\textcolor{black!55}{(64B)}} & 22.27 & 0.312 & 0.301 & 1.29 & 9.12 & 3.284 & 5.95 \\
\hspace{0.6em}+\,DiffusionNFT & 23.46 & 0.351 & 0.307 & 1.60 & 10.76 & 3.319 & 6.42 \\
\hspace{0.6em}+\,RAM & \underline{23.70} & \underline{0.365} & \underline{0.311} & \underline{1.62} & \underline{11.40} & \underline{3.329} & \underline{6.46} \\
\hspace{0.6em}+\,SIFR & \textbf{24.07} & \textbf{0.378} & \textbf{0.317} & \textbf{1.67} & \textbf{11.95} & \textbf{3.355} & \textbf{6.62} \\
\specialrule{0.7pt}{0.75pt}{0pt}
\end{tabular}}
\usebox{\preferencebox}
\normalsize
\caption{\textbf{Multi-reward training across model scales.} Rewards on held-out prompts at round $500$; bold is best and underline second best. {\footnotesize\textsuperscript{$\dagger$}\,round $300$, before RAM on FLUX.2 Klein collapses.}}
\label{tab:main}
\endgroup
\end{table}

%% file: figures/preference_training.tex
\begingroup%
\input{style/plots.tex}%
\begin{tikzpicture}[x=1pt,y=1pt]%
\path[use as bounding box] (0,12) rectangle (396,261);%
\begin{pgfinterruptboundingbox}%
\node[anchor=base,inner sep=0pt,text=sifrInk,font=\fontfamily{ptm}\fontsize{8.5}{9.7}\selectfont] at (79.45,238) {SD3.5 Medium (2.5B)};%
\begin{axis}[sifr axis,at={(30.0pt,149.5pt)},width=98.89999999999999pt,height=80.5pt,xmin=0,xmax=500,ymin=22.76952548,ymax=24.759281559999998,axis line style={black,line width=.5pt},y axis line style={draw=black,line width=.5pt},tick style={black,line width=.5pt},ytick style={draw=black,line width=.5pt},tick label style={text=black},xtick={0,100,200,300,400,500},ytick={23.0,23.5,24.0,24.5},xmajorgrids,ymajorgrids,grid style={black!18,line width=.4pt,dashed},major tick length=2.6pt,minor tick length=1.3pt,minor y tick num=1,yticklabel style={/pgf/number format/fixed,/pgf/number format/fixed zerofill,/pgf/number format/precision=1}]%
\addplot[sifrRAM,line width=.9pt,mark=none] coordinates {(0,21.297874) (20,22.805666) (40,23.011872) (60,23.1205) (80,23.33578) (100,23.308116) (120,23.35866) (140,23.42301) (160,23.417966) (180,23.484214) (200,23.531976) (220,23.60501) (240,23.544638) (260,23.386974000000002) (280,22.865803999999997) (300,23.246704) (320,23.261680000000002) (340,23.482784) (360,23.39142) (380,23.610782) (400,23.684674) (420,23.666916) (440,23.545106) (460,23.66117) (480,23.665798) (500,23.555012)};%
\addplot[sifrRAM,only marks,mark=*,mark size=1.4pt,mark options={draw=white,line width=.25pt}] coordinates {(400,23.684674)};%
\addplot[sifrNFT,line width=.9pt,mark=none] coordinates {(0,19.9732) (20,22.497462000000002) (40,22.8371) (60,22.979814) (80,23.041408) (100,23.08982) (120,23.125024) (140,23.155262) (160,23.17614) (180,23.188333999999998) (200,23.19499) (220,23.217064) (240,23.229934) (260,23.234952) (280,23.242231999999998) (300,23.254114) (320,23.256662000000002) (340,23.256376000000003) (360,23.265710000000002) (380,23.275122) (400,23.278918) (420,23.266672) (440,23.267452) (460,23.27078) (480,23.277487999999998) (500,23.286068)};%
\addplot[sifrNFT,only marks,mark=*,mark size=1.4pt,mark options={draw=white,line width=.25pt}] coordinates {(500,23.286068)};%
\addplot[sifrBlue,line width=1.2pt,mark=none] coordinates {(0,21.297874) (20,22.845133999999998) (40,23.17029) (60,23.389262000000002) (80,23.483408) (100,23.633662) (120,23.686598) (140,23.786646) (160,23.861448) (180,23.930166) (200,23.95393) (220,23.990642) (240,24.047244) (260,24.105666) (280,24.149918) (300,24.18156) (320,24.225604) (340,24.28361) (360,24.304228) (380,24.295882) (400,24.352562000000002) (420,24.37279) (440,24.390392000000002) (460,24.392966) (480,24.448658) (500,24.470446)};%
\addplot[sifrBlue,only marks,mark=*,mark size=1.4pt,mark options={draw=white,line width=.25pt}] coordinates {(500,24.470446)};%
\node[anchor=south,font=\fontfamily{ptm}\fontsize{6.2}{7.3}\selectfont,text=sifrRAM,inner sep=.4pt,fill=white,fill opacity=.8,text opacity=1,yshift=2.6pt] at (axis cs:400,23.684674) {23.68};%
\node[anchor=north east,font=\fontfamily{ptm}\fontsize{6.2}{7.3}\selectfont,text=sifrNFT,inner sep=.4pt,fill=white,fill opacity=.8,text opacity=1,yshift=-2.6pt] at (axis cs:500,23.286068) {23.29};%
\node[anchor=south east,font=\fontfamily{ptm}\fontsize{6.2}{7.3}\selectfont,text=sifrBlue,inner sep=.4pt,fill=white,fill opacity=.8,text opacity=1,yshift=2.6pt] at (axis cs:500,24.470446) {24.47};%
\draw[<->,sifrDeepRed,line width=.6pt] (axis cs:120,23.684674) -- (axis cs:400,23.684674);%
\node[anchor=south,font=\fontfamily{ptm}\fontsize{6.3}{7.4}\selectfont\bfseries\boldmath,text=sifrDeepRed,inner sep=.5pt,fill=white,fill opacity=.85,text opacity=1] at (axis cs:265,23.684674) {3.3$\times$ faster};%
\end{axis}%
\begin{axis}[sifr axis,at={(30.0pt,40pt)},width=98.89999999999999pt,height=80.5pt,xmin=0,xmax=500,ymin=1.51583364,ymax=1.64492508,axis line style={black,line width=.5pt},y axis line style={draw=black,line width=.5pt},tick style={black,line width=.5pt},ytick style={draw=black,line width=.5pt},tick label style={text=black},xtick={0,100,200,300,400,500},ytick={1.52,1.54,1.56,1.58,1.6,1.62,1.64},xmajorgrids,ymajorgrids,grid style={black!18,line width=.4pt,dashed},major tick length=2.6pt,minor tick length=1.3pt,minor y tick num=1,yticklabel style={/pgf/number format/fixed,/pgf/number format/fixed zerofill,/pgf/number format/precision=2}]%
\addplot[sifrRAM,line width=.9pt,mark=none] coordinates {(0,1.362631) (20,1.493698) (40,1.515298) (60,1.521399) (80,1.534721) (100,1.529182) (120,1.528996) (140,1.533484) (160,1.52208) (180,1.540672) (200,1.536284) (220,1.547889) (240,1.54897) (260,1.5537) (280,1.556378) (300,1.562312) (320,1.544025) (340,1.562372) (360,1.56377) (380,1.568745) (400,1.563669) (420,1.553777) (440,1.564017) (460,1.579351) (480,1.579623) (500,1.587741)};%
\addplot[sifrRAM,only marks,mark=*,mark size=1.4pt,mark options={draw=white,line width=.25pt}] coordinates {(500,1.587741)};%
\addplot[sifrNFT,line width=.9pt,mark=none] coordinates {(0,1.222433) (20,1.471077) (40,1.506491) (60,1.519424) (80,1.526595) (100,1.533811) (120,1.538898) (140,1.544766) (160,1.548991) (180,1.553471) (200,1.556958) (220,1.560017) (240,1.562419) (260,1.565575) (280,1.566641) (300,1.568463) (320,1.570713) (340,1.572285) (360,1.573214) (380,1.574205) (400,1.575645) (420,1.577228) (440,1.577499) (460,1.578415) (480,1.579068) (500,1.579911)};%
\addplot[sifrNFT,only marks,mark=*,mark size=1.4pt,mark options={draw=white,line width=.25pt}] coordinates {(500,1.579911)};%
\addplot[sifrBlue,line width=1.2pt,mark=none] coordinates {(0,1.362631) (20,1.503453) (40,1.52811) (60,1.540281) (80,1.550504) (100,1.557381) (120,1.56512) (140,1.571929) (160,1.576835) (180,1.580992) (200,1.585268) (220,1.586474) (240,1.590601) (260,1.594799) (280,1.597095) (300,1.601573) (320,1.605008) (340,1.608124) (360,1.610197) (380,1.613582) (400,1.616383) (420,1.618227) (440,1.621196) (460,1.622778) (480,1.624785) (500,1.626186)};%
\addplot[sifrBlue,only marks,mark=*,mark size=1.4pt,mark options={draw=white,line width=.25pt}] coordinates {(500,1.626186)};%
\node[anchor=south east,font=\fontfamily{ptm}\fontsize{6.2}{7.3}\selectfont,text=sifrRAM,inner sep=.4pt,fill=white,fill opacity=.8,text opacity=1,yshift=2.6pt] at (axis cs:500,1.587741) {1.588};%
\node[anchor=north east,font=\fontfamily{ptm}\fontsize{6.2}{7.3}\selectfont,text=sifrNFT,inner sep=.4pt,fill=white,fill opacity=.8,text opacity=1,yshift=-2.6pt] at (axis cs:500,1.579911) {1.580};%
\node[anchor=south east,font=\fontfamily{ptm}\fontsize{6.2}{7.3}\selectfont,text=sifrBlue,inner sep=.4pt,fill=white,fill opacity=.8,text opacity=1,yshift=2.6pt] at (axis cs:500,1.626186) {1.626};%
\draw[<->,sifrDeepRed,line width=.6pt] (axis cs:240,1.587741) -- (axis cs:500,1.587741);%
\node[anchor=south east,font=\fontfamily{ptm}\fontsize{6.3}{7.4}\selectfont\bfseries\boldmath,text=sifrDeepRed,inner sep=.5pt,fill=white,fill opacity=.85,text opacity=1,yshift=1.5pt] at (axis cs:234,1.587741) {2.1$\times$ faster};%
\end{axis}%
\node[anchor=base,inner sep=0pt,text=sifrInk,font=\fontfamily{ptm}\fontsize{8.5}{9.7}\selectfont] at (208,238) {FLUX.2 Klein (9B)};%
\begin{axis}[sifr axis,at={(158.55pt,149.5pt)},width=98.89999999999999pt,height=80.5pt,xmin=0,xmax=500,ymin=23.2216426,ymax=25.1387942,axis line style={black,line width=.5pt},y axis line style={draw=black,line width=.5pt},tick style={black,line width=.5pt},ytick style={draw=black,line width=.5pt},tick label style={text=black},xtick={0,100,200,300,400,500},ytick={23.5,24.0,24.5,25.0},xmajorgrids,ymajorgrids,grid style={black!18,line width=.4pt,dashed},major tick length=2.6pt,minor tick length=1.3pt,minor y tick num=1,yticklabel style={/pgf/number format/fixed,/pgf/number format/fixed zerofill,/pgf/number format/precision=1}]%
\addplot[sifrRAM,line width=.9pt,mark=none] coordinates {(0,21.813402) (20,22.871627999999998) (40,23.084958) (60,23.411596) (80,23.452337999999997) (100,23.524514) (120,23.68301) (140,23.749986) (160,23.756044) (180,23.777468000000002) (200,23.650952) (220,23.755368) (240,23.833289999999998) (260,23.817325999999998) (280,23.84447) (300,23.846056) (320,23.901202) (340,23.996752) (360,24.0227) (380,23.69419) (400,23.942307999999997) (420,23.900447999999997) (440,23.80313) (460,23.901904000000002) (480,23.737246) (500,23.945558)};%
\addplot[sifrRAM,only marks,mark=*,mark size=1.4pt,mark options={draw=white,line width=.25pt}] coordinates {(360,24.0227)};%
\addplot[sifrNFT,line width=.9pt,mark=none] coordinates {(0,19.927622) (20,22.636536) (40,23.01091) (60,23.165298) (80,23.256376000000003) (100,23.314408) (120,23.36243) (140,23.413598) (160,23.444434) (180,23.47267) (200,23.488946000000002) (220,23.511384) (240,23.537436) (260,23.568272) (280,23.59071) (300,23.60644) (320,23.62529) (340,23.63127) (360,23.632804) (380,23.620193999999998) (400,23.620818) (420,23.633792) (440,23.648585999999998) (460,23.65844) (480,23.667904) (500,23.680201999999998)};%
\addplot[sifrNFT,only marks,mark=*,mark size=1.4pt,mark options={draw=white,line width=.25pt}] coordinates {(500,23.680201999999998)};%
\addplot[sifrBlue,line width=1.2pt,mark=none] coordinates {(0,21.306304931640625) (20,23.234146) (40,23.544664) (60,23.78116) (80,23.860278) (100,23.970908) (120,24.120122) (140,24.205272) (160,24.30922) (180,24.410204) (200,24.42427) (220,24.453936000000002) (240,24.523538) (260,24.584014) (280,24.632920000000002) (300,24.64592) (320,24.649534) (340,24.630762) (360,24.635156) (380,24.654656) (400,24.79347) (420,24.85756) (440,24.79633) (460,24.854570000000002) (480,24.860498) (500,24.829766)};%
\addplot[sifrBlue,only marks,mark=*,mark size=1.4pt,mark options={draw=white,line width=.25pt}] coordinates {(480,24.860498)};%
\node[anchor=south,font=\fontfamily{ptm}\fontsize{6.2}{7.3}\selectfont,text=sifrRAM,inner sep=.4pt,fill=white,fill opacity=.8,text opacity=1,yshift=2.6pt] at (axis cs:360,24.0227) {24.02};%
\node[anchor=north east,font=\fontfamily{ptm}\fontsize{6.2}{7.3}\selectfont,text=sifrNFT,inner sep=.4pt,fill=white,fill opacity=.8,text opacity=1,yshift=-2.6pt] at (axis cs:500,23.680201999999998) {23.68};%
\node[anchor=south east,font=\fontfamily{ptm}\fontsize{6.2}{7.3}\selectfont,text=sifrBlue,inner sep=.4pt,fill=white,fill opacity=.8,text opacity=1,yshift=2.6pt] at (axis cs:480,24.860498) {24.86};%
\draw[<->,sifrDeepRed,line width=.6pt] (axis cs:120,24.0227) -- (axis cs:360,24.0227);%
\node[anchor=south,font=\fontfamily{ptm}\fontsize{6.3}{7.4}\selectfont\bfseries\boldmath,text=sifrDeepRed,inner sep=.5pt,fill=white,fill opacity=.85,text opacity=1] at (axis cs:235,24.0227) {3.0$\times$ faster};%
\end{axis}%
\begin{axis}[sifr axis,at={(158.55pt,40pt)},width=98.89999999999999pt,height=80.5pt,xmin=0,xmax=500,ymin=1.53837944,ymax=1.6618536799999999,axis line style={black,line width=.5pt},y axis line style={draw=black,line width=.5pt},tick style={black,line width=.5pt},ytick style={draw=black,line width=.5pt},tick label style={text=black},xtick={0,100,200,300,400,500},ytick={1.54,1.56,1.58,1.6,1.62,1.64,1.66},xmajorgrids,ymajorgrids,grid style={black!18,line width=.4pt,dashed},major tick length=2.6pt,minor tick length=1.3pt,minor y tick num=1,yticklabel style={/pgf/number format/fixed,/pgf/number format/fixed zerofill,/pgf/number format/precision=2}]%
\addplot[sifrRAM,line width=.9pt,mark=none] coordinates {(0,1.42797) (20,1.493987) (40,1.52549) (60,1.543312) (80,1.54713) (100,1.553427) (120,1.551938) (140,1.557289) (160,1.55992) (180,1.562847) (200,1.548131) (220,1.544354) (240,1.552568) (260,1.557451) (280,1.561116) (300,1.559952) (320,1.545058) (340,1.119668) (360,0.799369) (380,0.794926) (400,0.816516) (420,0.824428) (440,0.824939) (460,0.825462) (480,0.82574) (500,0.825736)};%
\node[anchor=south west,inner sep=0pt,text=sifrRAM,font=\fontfamily{ptm}\fontsize{6.5}{7.7}\selectfont] at (rel axis cs:0.70,0.10) {collapse};%
\addplot[sifrRAM,only marks,mark=*,mark size=1.4pt,mark options={draw=white,line width=.25pt}] coordinates {(180,1.562847)};%
\addplot[sifrNFT,line width=.9pt,mark=none] coordinates {(0,1.239847) (20,1.483997) (40,1.5149) (60,1.533212) (80,1.54749) (100,1.556491) (120,1.56315) (140,1.568738) (160,1.574481) (180,1.579397) (200,1.583931) (220,1.587358) (240,1.589148) (260,1.591616) (280,1.593916) (300,1.597349) (320,1.599211) (340,1.601577) (360,1.604457) (380,1.606885) (400,1.608673) (420,1.608531) (440,1.610732) (460,1.613443) (480,1.615525) (500,1.617211)};%
\addplot[sifrNFT,only marks,mark=*,mark size=1.4pt,mark options={draw=white,line width=.25pt}] coordinates {(500,1.617211)};%
\addplot[sifrBlue,line width=1.2pt,mark=none] coordinates {(0,1.3688092231750488) (20,1.525581) (40,1.54973) (60,1.566117) (80,1.576351) (100,1.57821) (120,1.589997) (140,1.599745) (160,1.603905) (180,1.609165) (200,1.609598) (220,1.611767) (240,1.615236) (260,1.616366) (280,1.618141) (300,1.623619) (320,1.624515) (340,1.619996) (360,1.615875) (380,1.627089) (400,1.632906) (420,1.635251) (440,1.633731) (460,1.635099) (480,1.639744) (500,1.64393)};%
\addplot[sifrBlue,only marks,mark=*,mark size=1.4pt,mark options={draw=white,line width=.25pt}] coordinates {(500,1.64393)};%
\node[anchor=north,font=\fontfamily{ptm}\fontsize{6.2}{7.3}\selectfont,text=sifrRAM,inner sep=.4pt,fill=white,fill opacity=.8,text opacity=1,yshift=-2.6pt] at (axis cs:180,1.562847) {1.563};%
\node[anchor=south east,font=\fontfamily{ptm}\fontsize{6.2}{7.3}\selectfont,text=sifrNFT,inner sep=.4pt,fill=white,fill opacity=.8,text opacity=1,yshift=2.6pt] at (axis cs:500,1.617211) {1.617};%
\node[anchor=south east,font=\fontfamily{ptm}\fontsize{6.2}{7.3}\selectfont,text=sifrBlue,inner sep=.4pt,fill=white,fill opacity=.8,text opacity=1,yshift=2.6pt] at (axis cs:500,1.64393) {1.644};%
\draw[<->,sifrDeepRed,line width=.6pt] (axis cs:280,1.617211) -- (axis cs:500,1.617211);%
\node[anchor=south east,font=\fontfamily{ptm}\fontsize{6.3}{7.4}\selectfont\bfseries\boldmath,text=sifrDeepRed,inner sep=.5pt,fill=white,fill opacity=.85,text opacity=1,yshift=1.5pt] at (axis cs:274,1.617211) {1.8$\times$ faster};%
\end{axis}%
\node[anchor=base,inner sep=0pt,text=sifrInk,font=\fontfamily{ptm}\fontsize{8.5}{9.7}\selectfont] at (336.55,238) {Cosmos 3 Super T2I (64B)};%
\begin{axis}[sifr axis,at={(287.1pt,149.5pt)},width=98.89999999999999pt,height=80.5pt,xmin=0,xmax=500,ymin=23.14838032,ymax=24.88863104,axis line style={black,line width=.5pt},y axis line style={draw=black,line width=.5pt},tick style={black,line width=.5pt},ytick style={draw=black,line width=.5pt},tick label style={text=black},xtick={0,100,200,300,400,500},ytick={23.25,23.5,23.75,24.0,24.25,24.5,24.75},xmajorgrids,ymajorgrids,grid style={black!18,line width=.4pt,dashed},major tick length=2.6pt,minor tick length=1.3pt,minor y tick num=1,yticklabel style={/pgf/number format/fixed,/pgf/number format/fixed zerofill,/pgf/number format/precision=2}]%
\addplot[sifrRAM,line width=.9pt,mark=none] coordinates {(0,21.773284) (20,23.154456) (40,23.433306) (60,23.567804) (80,23.662834) (100,23.654851999999998) (120,23.735114) (140,23.779002) (160,23.780354) (180,23.840336) (200,23.878322) (220,23.937289999999997) (240,23.880662) (260,23.97122) (280,23.832952) (300,23.816546) (320,24.028394) (340,23.834668) (360,23.97616) (380,24.075636) (400,23.855078) (420,24.022986) (440,24.103456) (460,24.091054) (480,24.11669) (500,24.136996)};%
\addplot[sifrRAM,only marks,mark=*,mark size=1.4pt,mark options={draw=white,line width=.25pt}] coordinates {(500,24.136996)};%
\addplot[sifrNFT,line width=.9pt,mark=none] coordinates {(0,21.773284) (20,22.33556) (40,22.799166) (60,23.017332) (80,23.142002) (100,23.232586) (120,23.316254) (140,23.35918) (160,23.415001999999998) (180,23.456706) (200,23.496174) (220,23.532678) (240,23.57017) (260,23.599758) (280,23.642008) (300,23.689249999999998) (320,23.713092) (340,23.729784000000002) (360,23.746294) (380,23.773802) (400,23.809317999999998) (420,23.831989999999998) (440,23.85981) (460,23.88542) (480,23.901514) (500,23.921196000000002)};%
\addplot[sifrNFT,only marks,mark=*,mark size=1.4pt,mark options={draw=white,line width=.25pt}] coordinates {(500,23.921196000000002)};%
\addplot[sifrBlue,line width=1.2pt,mark=none] coordinates {(0,21.773284) (20,23.160410000000002) (40,23.44745) (60,23.628254) (80,23.786672) (100,23.830612) (120,23.940202) (140,23.969894) (160,24.088246) (180,24.094616) (200,24.128208) (220,24.196198) (240,24.264032) (260,24.302772) (280,24.33639) (300,24.369436) (320,24.421774) (340,24.460774) (360,24.471096) (380,24.496446) (400,24.518103999999997) (420,24.532326) (440,24.557156) (460,24.583416) (480,24.620804) (500,24.636014)};%
\addplot[sifrBlue,only marks,mark=*,mark size=1.4pt,mark options={draw=white,line width=.25pt}] coordinates {(500,24.636014)};%
\node[anchor=south east,font=\fontfamily{ptm}\fontsize{6.2}{7.3}\selectfont,text=sifrRAM,inner sep=.4pt,fill=white,fill opacity=.8,text opacity=1,yshift=2.6pt] at (axis cs:500,24.136996) {24.14};%
\node[anchor=north east,font=\fontfamily{ptm}\fontsize{6.2}{7.3}\selectfont,text=sifrNFT,inner sep=.4pt,fill=white,fill opacity=.8,text opacity=1,yshift=-2.6pt] at (axis cs:500,23.921196000000002) {23.92};%
\node[anchor=south east,font=\fontfamily{ptm}\fontsize{6.2}{7.3}\selectfont,text=sifrBlue,inner sep=.4pt,fill=white,fill opacity=.8,text opacity=1,yshift=2.6pt] at (axis cs:500,24.636014) {24.64};%
\draw[<->,sifrDeepRed,line width=.6pt] (axis cs:220,24.136996) -- (axis cs:500,24.136996);%
\node[anchor=south east,font=\fontfamily{ptm}\fontsize{6.3}{7.4}\selectfont\bfseries\boldmath,text=sifrDeepRed,inner sep=.5pt,fill=white,fill opacity=.85,text opacity=1,yshift=1.5pt] at (axis cs:214,24.136996) {2.3$\times$ faster};%
\end{axis}%
\begin{axis}[sifr axis,at={(287.1pt,40pt)},width=98.89999999999999pt,height=80.5pt,xmin=0,xmax=500,ymin=1.51837098,ymax=1.63817606,axis line style={black,line width=.5pt},y axis line style={draw=black,line width=.5pt},tick style={black,line width=.5pt},ytick style={draw=black,line width=.5pt},tick label style={text=black},xtick={0,100,200,300,400,500},ytick={1.52,1.54,1.56,1.58,1.6,1.62},xmajorgrids,ymajorgrids,grid style={black!18,line width=.4pt,dashed},major tick length=2.6pt,minor tick length=1.3pt,minor y tick num=1,yticklabel style={/pgf/number format/fixed,/pgf/number format/fixed zerofill,/pgf/number format/precision=2}]%
\addplot[sifrRAM,line width=.9pt,mark=none] coordinates {(0,1.410175) (20,1.524063) (40,1.544399) (60,1.552959) (80,1.553449) (100,1.556253) (120,1.558058) (140,1.561779) (160,1.563087) (180,1.556247) (200,1.566151) (220,1.567496) (240,1.566052) (260,1.570503) (280,1.579478) (300,1.57792) (320,1.574821) (340,1.580242) (360,1.567069) (380,1.580989) (400,1.581304) (420,1.584101) (440,1.584682) (460,1.583575) (480,1.587617) (500,1.586908)};%
\addplot[sifrRAM,only marks,mark=*,mark size=1.4pt,mark options={draw=white,line width=.25pt}] coordinates {(480,1.587617)};%
\addplot[sifrNFT,line width=.9pt,mark=none] coordinates {(0,1.410175) (20,1.458269) (40,1.49382) (60,1.508806) (80,1.518173) (100,1.524168) (120,1.528514) (140,1.533098) (160,1.535344) (180,1.539779) (200,1.542592) (220,1.544519) (240,1.545637) (260,1.547829) (280,1.5492) (300,1.551468) (320,1.553199) (340,1.54736) (360,1.545332) (380,1.548207) (400,1.550715) (420,1.553845) (440,1.555685) (460,1.557652) (480,1.559614) (500,1.561052)};%
\addplot[sifrNFT,only marks,mark=*,mark size=1.4pt,mark options={draw=white,line width=.25pt}] coordinates {(500,1.561052)};%
\addplot[sifrBlue,line width=1.2pt,mark=none] coordinates {(0,1.410175) (20,1.525698) (40,1.547088) (60,1.558115) (80,1.567147) (100,1.573804) (120,1.578677) (140,1.581859) (160,1.584787) (180,1.586424) (200,1.588308) (220,1.590082) (240,1.592512) (260,1.59473) (280,1.597374) (300,1.599521) (320,1.602455) (340,1.606598) (360,1.608614) (380,1.61114) (400,1.613798) (420,1.614298) (440,1.617188) (460,1.619081) (480,1.620084) (500,1.620785)};%
\addplot[sifrBlue,only marks,mark=*,mark size=1.4pt,mark options={draw=white,line width=.25pt}] coordinates {(500,1.620785)};%
\node[anchor=south east,font=\fontfamily{ptm}\fontsize{6.2}{7.3}\selectfont,text=sifrRAM,inner sep=.4pt,fill=white,fill opacity=.8,text opacity=1,yshift=2.6pt] at (axis cs:480,1.587617) {1.588};%
\node[anchor=north east,font=\fontfamily{ptm}\fontsize{6.2}{7.3}\selectfont,text=sifrNFT,inner sep=.4pt,fill=white,fill opacity=.8,text opacity=1,yshift=-2.6pt] at (axis cs:500,1.561052) {1.561};%
\node[anchor=south east,font=\fontfamily{ptm}\fontsize{6.2}{7.3}\selectfont,text=sifrBlue,inner sep=.4pt,fill=white,fill opacity=.8,text opacity=1,yshift=2.6pt] at (axis cs:500,1.620785) {1.621};%
\draw[<->,sifrDeepRed,line width=.6pt] (axis cs:200,1.587617) -- (axis cs:480,1.587617);%
\node[anchor=south east,font=\fontfamily{ptm}\fontsize{6.3}{7.4}\selectfont\bfseries\boldmath,text=sifrDeepRed,inner sep=.5pt,fill=white,fill opacity=.85,text opacity=1,yshift=1.5pt] at (axis cs:194,1.587617) {2.4$\times$ faster};%
\end{axis}%
\node[anchor=center,inner sep=0pt,text=black,font=\fontfamily{ptm}\fontsize{7.8}{9.0}\selectfont,rotate=90] at (4,189.75) {Single reward};%
\node[anchor=center,inner sep=0pt,text=black,font=\fontfamily{ptm}\fontsize{7.8}{9.0}\selectfont,rotate=90] at (4,80.25) {Multiple rewards};%
\node[anchor=base,inner sep=0pt,text=black,font=\fontfamily{ptm}\fontsize{7.6}{8.8}\selectfont] at (208,19) {Training round};%
\node[anchor=base,inner sep=0pt,text=sifrInk,font=\fontfamily{ptm}\fontsize{7.2}{8.4}\selectfont] at (198,252) {\textcolor{sifrBlue}{\rule[1.6pt]{12pt}{1.2pt}}\hspace{3pt}SIFR\hspace{15pt}\textcolor{sifrNFT}{\rule[1.75pt]{12pt}{.9pt}}\hspace{3pt}DiffusionNFT\hspace{15pt}\textcolor{sifrRAM}{\rule[1.75pt]{12pt}{.9pt}}\hspace{3pt}RAM};%
\end{pgfinterruptboundingbox}%
\end{tikzpicture}%
\endgroup%

%% file: figures/rule_based_results.tex
\begin{figure}[H]
\centering
\setlength{\parskip}{0pt}
\setlength{\intextsep}{0pt}
\setlength{\abovecaptionskip}{4pt}
\setlength{\belowcaptionskip}{0pt}
\noindent\begin{minipage}[t]{\dimexpr\textwidth-148pt\relax}
\vspace{0pt}
\input{tables/rule_based}
\end{minipage}\hfill
\begin{minipage}[t]{136pt}
\vspace{0pt}
\centering
\resizebox{\linewidth}{!}{\input{figures/long_caption_training.tex}}
\caption{Long-caption training on Cosmos 3 Super T2I.}
\label{fig:longprompt}
\end{minipage}
\end{figure}

%% file: tables/rule_based.tex
\begin{table}[H]
\centering
\begingroup\footnotesize
\setlength{\tabcolsep}{3pt}
\renewcommand{\arraystretch}{1.08}
\newcommand{\scoreandround}[2]{\begin{tabular}[c]{@{}c@{}}#1\\[-2pt]{\scriptsize(#2)}\end{tabular}}
\newsavebox{\optimizationsummarybox}
\sbox{\optimizationsummarybox}{%
\begin{tabular}{l*{6}{c}}
\toprule[0.7pt]
\multirow{2}{*}{Method} & \multicolumn{2}{c}{\shortstack{SD3.5\\Medium}} & \multicolumn{2}{c}{\shortstack{FLUX.2\\Klein}} & \multicolumn{2}{c}{\shortstack{Cosmos 3\\Super T2I}} \\
\cmidrule(lr){2-3}\cmidrule(lr){4-5}\cmidrule(l){6-7}
& OCR & GenEval & OCR & GenEval & OCR & GenEval \\
\midrule[0.35pt]
Pretrained & 0.631 & 0.652 & 0.783 & 0.809 & 0.728 & 0.846 \\
DiffusionNFT & \scoreandround{\textbf{0.985}}{80} & \scoreandround{0.972}{500} & \scoreandround{0.988}{40} & \scoreandround{0.971}{160} & \scoreandround{0.976}{100} & \scoreandround{0.975}{260} \\
RAM & \scoreandround{0.974}{160} & \scoreandround{0.973}{\textbf{270}} & \scoreandround{0.986}{30} & \scoreandround{0.989}{80} & \scoreandround{0.982}{100} & \scoreandround{0.970}{290} \\
\rowcolor{black!4}\textbf{SIFR} & \scoreandround{0.984}{\textbf{40}} & \scoreandround{\textbf{0.978}}{\textbf{270}} & \scoreandround{\textbf{0.989}}{\textbf{25}} & \scoreandround{\textbf{0.994}}{\textbf{60}} & \scoreandround{\textbf{0.988}}{\textbf{50}} & \scoreandround{\textbf{0.987}}{\textbf{140}} \\
\bottomrule[0.7pt]
\end{tabular}}
\usebox{\optimizationsummarybox}
\endgroup
\caption{\textbf{Rule-based reward optimization.} Peak OCR and GenEval Overall scores within $500$ rounds, with the first evaluation round reaching $0.97$ in parentheses (fewer is better).}
\label{tab:rule-based}
\end{table}

%% file: figures/long_caption_training.tex
\begingroup%
\input{style/plots.tex}%
\definecolor{sifrTeal}{HTML}{008575}%
\definecolor{sifrRed}{HTML}{D45347}%
\definecolor{sifrBase}{HTML}{C5CBD3}%
\begin{tikzpicture}[x=1pt,y=1pt]
\path[use as bounding box] (0,0) rectangle (152,131);
\draw[sifrBlue,line width=1.2pt] (3,124) -- (16,124);
\node[anchor=west,text=black,font=\fontsize{7.2}{8.7}\selectfont,inner sep=0pt] at (21,124) {SIFR};
\draw[sifrNFT,line width=0.9pt] (42,124) -- (55,124);
\node[anchor=west,text=black,font=\fontsize{7.2}{8.7}\selectfont,inner sep=0pt] at (60,124) {DiffusionNFT};
\draw[sifrRAM,line width=0.9pt] (113,124) -- (126,124);
\node[anchor=west,text=black,font=\fontsize{7.2}{8.7}\selectfont,inner sep=0pt] at (131,124) {RAM};
\begin{axis}[sifr axis,scale only axis,anchor=south west,at={(31pt,29pt)},width=112pt,height=82pt,tick label style={font=\fontsize{7.2}{8.5}\selectfont},label style={font=\fontsize{7.8}{9}\selectfont},clip mode=individual,axis line style={black,line width=.5pt},y axis line style={draw=black,line width=.5pt},tick style={black,line width=.5pt},ytick style={draw=black,line width=.5pt},tick label style={text=black},label style={text=black},grid style={black!18,line width=.4pt,dashed},major tick length=2.6pt,minor tick length=1.3pt,axis line style={black,line width=.5pt},y axis line style={draw=black,line width=.5pt},tick style={black,line width=.5pt},ytick style={draw=black,line width=.5pt},tick label style={text=black},label style={text=black},xmin=0,xmax=500,ymin=22.909821999999998,ymax=25.764673999999999,xtick={0,100,200,300,400,500},ytick={23.0,23.5,24.0,24.5,25.0,25.5},xmajorgrids,ymajorgrids,minor y tick num=1,yticklabel style={/pgf/number format/fixed,/pgf/number format/fixed zerofill,/pgf/number format/precision=1},xlabel={Training round},ylabel={PickScore}]
\addplot[sifrBlue,line width=1.2pt,no marks] coordinates {
(0,22.292660000000001)
(20,23.210460000000001)
(40,23.828479999999999)
(60,24.14594)
(80,24.274640000000002)
(100,24.401)
(120,24.481860000000001)
(140,24.478739999999998)
(160,24.593139999999998)
(180,24.637599999999999)
(200,24.70702)
(220,24.76266)
(240,24.818560000000002)
(260,24.870039999999999)
(280,24.914239999999999)
(300,24.9496)
(320,24.996659999999999)
(340,25.03566)
(360,25.086099999999998)
(380,25.143820000000002)
(400,25.182559999999999)
(420,25.204660000000001)
(440,25.252759999999999)
(460,25.284220000000001)
(480,25.311520000000002)
(500,25.350259999999999)
};
\addplot[sifrBlue,only marks,mark=*,mark size=1.4pt,mark options={draw=white,line width=.25pt}] coordinates {
(500,25.350259999999999)
};
\node[anchor=south east,inner sep=.4pt,fill=white,fill opacity=.8,text opacity=1,text=sifrBlue,font=\fontfamily{ptm}\fontsize{6.2}{7.3}\selectfont,yshift=2.6pt] at (axis cs:500,25.350259999999999) {25.35};
\addplot[sifrRAM,line width=.9pt,no marks] coordinates {
(0,22.292660000000001)
(20,23.1816)
(40,23.460059999999999)
(60,23.643879999999999)
(80,23.897379999999998)
(100,24.08146)
(120,24.20834)
(140,24.141259999999999)
(160,24.182079999999999)
(180,24.279579999999999)
(200,24.243960000000001)
(220,24.201840000000001)
(240,24.404640000000001)
(260,24.511240000000001)
(280,24.32066)
(300,24.187280000000001)
(320,24.178180000000001)
(340,24.42258)
(360,23.402080000000002)
(380,24.06456)
(400,24.330279999999998)
(420,24.36824)
(440,24.233560000000001)
(460,24.433759999999999)
(480,24.558820000000001)
(500,24.5505)
};
\addplot[sifrRAM,only marks,mark=*,mark size=1.4pt,mark options={draw=white,line width=.25pt}] coordinates {
(480,24.558820000000001)
};
\node[anchor=south east,inner sep=.4pt,fill=white,fill opacity=.8,text opacity=1,text=sifrRAM,font=\fontfamily{ptm}\fontsize{6.2}{7.3}\selectfont,yshift=2.6pt] at (axis cs:480,24.558820000000001) {24.56};
\addplot[sifrNFT,line width=.9pt,no marks] coordinates {
(0,22.292660000000001)
(20,22.423439999999999)
(40,22.65146)
(60,22.806940000000001)
(80,22.944220000000001)
(100,23.04796)
(120,23.119980000000002)
(140,23.1998)
(160,23.254919999999998)
(180,23.319659999999999)
(200,23.363600000000002)
(220,23.409880000000001)
(240,23.447579999999999)
(260,23.491779999999999)
(280,23.532859999999999)
(300,23.56718)
(320,23.60904)
(340,23.62828)
(360,23.646999999999998)
(380,23.672999999999998)
(400,23.71668)
(420,23.761659999999999)
(440,23.780899999999999)
(460,23.80742)
(480,23.826139999999999)
(500,23.8459)
};
\addplot[sifrNFT,only marks,mark=*,mark size=1.4pt,mark options={draw=white,line width=.25pt}] coordinates {
(500,23.8459)
};
\node[anchor=north east,inner sep=.4pt,fill=white,fill opacity=.8,text opacity=1,text=sifrNFT,font=\fontfamily{ptm}\fontsize{6.2}{7.3}\selectfont,yshift=-2.6pt] at (axis cs:500,23.8459) {23.85};
\end{axis}
\end{tikzpicture}%
\endgroup%

%% file: figures/reward_gain_multi.tex
\begingroup%
\input{style/plots.tex}%
\begin{tikzpicture}[x=1pt,y=1pt]%
\path[use as bounding box] (0,0) rectangle (396,127);%
\begin{pgfinterruptboundingbox}%
\node[anchor=base,inner sep=0pt,text=sifrInk,font=\fontfamily{ptm}\fontsize{7.6}{8.8}\selectfont] at (198,116) {\textcolor{sifrBlue}{$\bullet$}\hspace{2pt}Higher than both\hspace{9pt}\textcolor{sifrMuted}{$\bullet$}\hspace{2pt}Other outcomes\hspace{16pt}\tikz[baseline=-.6ex]{\fill[sifrBlue,fill opacity=.85] (0,-2.2pt) rectangle (7pt,2.2pt);}\hspace{2pt}SIFR\hspace{9pt}\tikz[baseline=-.6ex]{\draw[sifrNFT,line width=.8pt] (0,0) -- (9pt,0);}\hspace{2pt}DiffusionNFT\hspace{9pt}\tikz[baseline=-.6ex]{\draw[sifrRAM,line width=.8pt] (0,0) -- (9pt,0);}\hspace{2pt}RAM\hspace{9pt}\tikz[baseline=-.6ex]{\fill[sifrBase!75] (0,-2.2pt) rectangle (7pt,2.2pt);}\hspace{2pt}Pretrained};%
\node[anchor=base,inner sep=0pt,text=sifrInk,font=\fontfamily{ptm}\fontsize{8.0}{9.2}\selectfont] at (71.0,102) {SD3.5 Medium (2.5B)};%
\begin{axis}[sifr axis,at={(36.0pt,24pt)},width=70pt,height=70pt,xmin=-0.12,xmax=0.24,ymin=-0.12,ymax=0.24,xtick={-0.1,0,0.1,0.2},ytick={-0.1,0,0.1,0.2},ymajorgrids=false,axis line style={black,line width=.5pt},y axis line style={draw=black,line width=.5pt},tick style={black,line width=.5pt},ytick style={draw=black,line width=.5pt},major tick length=2.6pt,minor tick length=1.3pt,minor tick num=1,tick label style={font=\fontfamily{ptm}\fontsize{6.2}{7.4}\selectfont,text=black},xticklabel style={/pgf/number format/fixed,/pgf/number format/precision=1},yticklabel style={/pgf/number format/fixed,/pgf/number format/precision=1},clip=true,enlargelimits=false]%
\path[fill=sifrBlue!4] (axis cs:0,0) rectangle (axis cs:0.24,0.24);%
\addplot[sifrBase,line width=.85pt,mark=none] coordinates {(-0.12,0) (0.24,0)};%
\addplot[sifrBase,line width=.85pt,mark=none] coordinates {(0,-0.12) (0,0.24)};%
\node[anchor=south west,inner sep=0pt] at (rel axis cs:0,0) {\includegraphics[width=70pt,height=70pt]{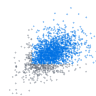}};%
\end{axis}%
\node[anchor=north east,inner sep=0pt,text=sifrBlue,font=\fontfamily{ptm}\fontsize{7.6}{8.8}\selectfont,fill=white,inner sep=.5pt,font=\fontfamily{ptm}\fontsize{7.6}{8.6}\selectfont\bfseries] at (104.95,92.95) {81.4\%};%
\node[anchor=north west,inner sep=0pt,text=sifrInk,font=\fontfamily{ptm}\fontsize{6.2}{7.4}\selectfont,fill=white,inner sep=.5pt] at (37.05,92.95) {2.7\%};%
\node[anchor=south east,inner sep=0pt,text=sifrInk,font=\fontfamily{ptm}\fontsize{6.2}{7.4}\selectfont,fill=white,inner sep=.5pt] at (104.95,25.05) {12.5\%};%
\node[anchor=south west,inner sep=0pt,text=sifrInk,font=\fontfamily{ptm}\fontsize{6.2}{7.4}\selectfont,fill=white,inner sep=.5pt] at (37.05,25.05) {3.4\%};%
\node[anchor=base,inner sep=0pt,text=sifrInk,font=\fontfamily{ptm}\fontsize{8.0}{9.2}\selectfont] at (163.0,102) {FLUX.2 Klein (9B)};%
\begin{axis}[sifr axis,at={(128.0pt,24pt)},width=70pt,height=70pt,xmin=-0.14,xmax=0.26,ymin=-0.14,ymax=0.26,xtick={-0.1,0,0.1,0.2},ytick={-0.1,0,0.1,0.2},ymajorgrids=false,axis line style={black,line width=.5pt},y axis line style={draw=black,line width=.5pt},tick style={black,line width=.5pt},ytick style={draw=black,line width=.5pt},major tick length=2.6pt,minor tick length=1.3pt,minor tick num=1,tick label style={font=\fontfamily{ptm}\fontsize{6.2}{7.4}\selectfont,text=black},xticklabel style={/pgf/number format/fixed,/pgf/number format/precision=1},yticklabel style={/pgf/number format/fixed,/pgf/number format/precision=1},clip=true,enlargelimits=false]%
\path[fill=sifrBlue!4] (axis cs:0,0) rectangle (axis cs:0.26,0.26);%
\addplot[sifrBase,line width=.85pt,mark=none] coordinates {(-0.14,0) (0.26,0)};%
\addplot[sifrBase,line width=.85pt,mark=none] coordinates {(0,-0.14) (0,0.26)};%
\node[anchor=south west,inner sep=0pt] at (rel axis cs:0,0) {\includegraphics[width=70pt,height=70pt]{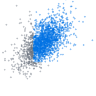}};%
\end{axis}%
\node[anchor=north east,inner sep=0pt,text=sifrBlue,font=\fontfamily{ptm}\fontsize{7.6}{8.8}\selectfont,fill=white,inner sep=.5pt,font=\fontfamily{ptm}\fontsize{7.6}{8.6}\selectfont\bfseries] at (196.95,92.95) {79.9\%};%
\node[anchor=north west,inner sep=0pt,text=sifrInk,font=\fontfamily{ptm}\fontsize{6.2}{7.4}\selectfont,fill=white,inner sep=.5pt] at (129.05,92.95) {15.3\%};%
\node[anchor=south east,inner sep=0pt,text=sifrInk,font=\fontfamily{ptm}\fontsize{6.2}{7.4}\selectfont,fill=white,inner sep=.5pt] at (196.95,25.05) {1.3\%};%
\node[anchor=south west,inner sep=0pt,text=sifrInk,font=\fontfamily{ptm}\fontsize{6.2}{7.4}\selectfont,fill=white,inner sep=.5pt] at (129.05,25.05) {3.5\%};%
\node[anchor=base,inner sep=0pt,text=sifrInk,font=\fontfamily{ptm}\fontsize{8.0}{9.2}\selectfont] at (255.0,102) {Cosmos 3 Super T2I (64B)};%
\begin{axis}[sifr axis,at={(220.0pt,24pt)},width=70pt,height=70pt,xmin=-0.12,xmax=0.22,ymin=-0.12,ymax=0.22,xtick={-0.1,0,0.1,0.2},ytick={-0.1,0,0.1,0.2},ymajorgrids=false,axis line style={black,line width=.5pt},y axis line style={draw=black,line width=.5pt},tick style={black,line width=.5pt},ytick style={draw=black,line width=.5pt},major tick length=2.6pt,minor tick length=1.3pt,minor tick num=1,tick label style={font=\fontfamily{ptm}\fontsize{6.2}{7.4}\selectfont,text=black},xticklabel style={/pgf/number format/fixed,/pgf/number format/precision=1},yticklabel style={/pgf/number format/fixed,/pgf/number format/precision=1},clip=true,enlargelimits=false]%
\path[fill=sifrBlue!4] (axis cs:0,0) rectangle (axis cs:0.22,0.22);%
\addplot[sifrBase,line width=.85pt,mark=none] coordinates {(-0.12,0) (0.22,0)};%
\addplot[sifrBase,line width=.85pt,mark=none] coordinates {(0,-0.12) (0,0.22)};%
\node[anchor=south west,inner sep=0pt] at (rel axis cs:0,0) {\includegraphics[width=70pt,height=70pt]{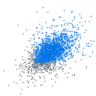}};%
\end{axis}%
\node[anchor=north east,inner sep=0pt,text=sifrBlue,font=\fontfamily{ptm}\fontsize{7.6}{8.8}\selectfont,fill=white,inner sep=.5pt,font=\fontfamily{ptm}\fontsize{7.6}{8.6}\selectfont\bfseries] at (288.95,92.95) {80.9\%};%
\node[anchor=north west,inner sep=0pt,text=sifrInk,font=\fontfamily{ptm}\fontsize{6.2}{7.4}\selectfont,fill=white,inner sep=.5pt] at (221.05,92.95) {2.5\%};%
\node[anchor=south east,inner sep=0pt,text=sifrInk,font=\fontfamily{ptm}\fontsize{6.2}{7.4}\selectfont,fill=white,inner sep=.5pt] at (288.95,25.05) {12.6\%};%
\node[anchor=south west,inner sep=0pt,text=sifrInk,font=\fontfamily{ptm}\fontsize{6.2}{7.4}\selectfont,fill=white,inner sep=.5pt] at (221.05,25.05) {4.0\%};%
\begin{axis}[sifr axis,at={(318pt,24pt)},width=72pt,height=70pt,xmin=1.15,xmax=1.875,ymin=-1.6959375,ymax=10.59960938,xtick={1.2,1.4,1.6,1.8},ytick={0,3,6,9},ymajorgrids=false,axis line style={black,line width=.5pt},y axis line style={draw=black,line width=.5pt},tick style={black,line width=.5pt},ytick style={draw=black,line width=.5pt},major tick length=2.6pt,minor tick length=1.3pt,minor tick num=1,tick label style={font=\fontfamily{ptm}\fontsize{6.2}{7.4}\selectfont,text=black},xticklabel style={/pgf/number format/fixed,/pgf/number format/precision=1},clip=true]%
\addplot[const plot,draw=none,fill=sifrBase!75] coordinates {(1.15,0) (1.15,0.43945) (1.175,0.73242) (1.2,0.68359) (1.225,1.2695) (1.25,2.0508) (1.275,2.3926) (1.3,3.6133) (1.325,4.248) (1.35,5.0293) (1.375,6.6406) (1.4,7.9102) (1.425,7.7637) (1.45,8.6914) (1.475,7.5195) (1.5,7.8613) (1.525,6.6406) (1.55,6.1035) (1.575,5.5664) (1.6,4.1504) (1.625,3.3691) (1.65,2.1973) (1.675,1.6602) (1.7,1.4648) (1.725,0.29297) (1.75,0.48828) (1.775,0.29297) (1.8,0.048828) (1.825,0.14648) (1.85,0) (1.875,0) (1.875,0)} \closedcycle;%
\addplot[const plot,draw=none,fill=sifrBlue,fill opacity=.85] coordinates {(1.15,0) (1.15,0) (1.175,0.048828) (1.2,0.048828) (1.225,0) (1.25,0.097656) (1.275,0.14648) (1.3,0.29297) (1.325,0.14648) (1.35,0.78125) (1.375,0.78125) (1.4,1.5137) (1.425,1.8066) (1.45,3.6621) (1.475,4.248) (1.5,5.5176) (1.525,7.0312) (1.55,7.1289) (1.575,9.082) (1.6,9.2773) (1.625,9.1309) (1.65,7.8613) (1.675,6.6895) (1.7,6.0547) (1.725,6.3965) (1.75,4.1504) (1.775,3.0762) (1.8,2.4902) (1.825,0.97656) (1.85,0.83008) (1.875,0.83008) (1.875,0)} \closedcycle;%
\addplot[const plot,sifrNFT,line width=.6pt,mark=none] coordinates {(1.15,0) (1.15,0.048828) (1.175,0) (1.2,0.14648) (1.225,0.3418) (1.25,0.3418) (1.275,0.3418) (1.3,0.97656) (1.325,0.92773) (1.35,1.9531) (1.375,2.9785) (1.4,3.6133) (1.425,4.8828) (1.45,5.0781) (1.475,7.666) (1.5,8.7402) (1.525,8.6426) (1.55,8.6914) (1.575,7.666) (1.6,7.6172) (1.625,5.7129) (1.65,6.3965) (1.675,5.3711) (1.7,4.3945) (1.725,2.9785) (1.75,2.002) (1.775,0.87891) (1.8,0.73242) (1.825,0.53711) (1.85,0.14648) (1.875,0.14648) (1.875,0)};%
\addplot[const plot,sifrRAM,line width=.6pt,mark=none] coordinates {(1.15,0) (1.15,0.097656) (1.175,0.048828) (1.2,0) (1.225,0.14648) (1.25,0.097656) (1.275,0.19531) (1.3,0.3418) (1.325,0.58594) (1.35,1.0254) (1.375,1.9531) (1.4,2.5879) (1.425,3.1738) (1.45,4.8828) (1.475,6.0547) (1.5,7.959) (1.525,8.3008) (1.55,9.8145) (1.575,9.3262) (1.6,7.4219) (1.625,6.9824) (1.65,6.9336) (1.675,5.7129) (1.7,5.5176) (1.725,3.1738) (1.75,2.6367) (1.775,2.1484) (1.8,1.4648) (1.825,0.92773) (1.85,0.19531) (1.875,0.19531) (1.875,0)};%
\draw[sifrMuted,line width=.9pt] (axis cs:1.469,-1.44155) -- (axis cs:1.469,-0.339188);%
\draw[sifrNFT,line width=.9pt] (axis cs:1.56105,-1.44155) -- (axis cs:1.56105,-0.339188);%
\draw[sifrRAM,line width=.9pt] (axis cs:1.58691,-1.44155) -- (axis cs:1.58691,-0.339188);%
\draw[sifrBlue,line width=.9pt] (axis cs:1.62079,-1.44155) -- (axis cs:1.62079,-0.339188);%
\end{axis}%
\node[anchor=base,inner sep=0pt,text=sifrInk,font=\fontfamily{ptm}\fontsize{8.0}{9.2}\selectfont] at (354.0,102) {Cosmos 3 Super T2I};%
\node[anchor=center,inner sep=0pt,text=black,font=\fontfamily{ptm}\fontsize{6.8}{8.0}\selectfont,rotate=90] at (305,59.0) {Samples (\%)};%
\node[anchor=base,inner sep=0pt,text=black,font=\fontfamily{ptm}\fontsize{7.2}{8.4}\selectfont] at (354.0,4.5) {Multi-reward score};%
\node[anchor=center,inner sep=0pt,text=black,font=\fontfamily{ptm}\fontsize{7.8}{9.0}\selectfont,rotate=90] at (5,59.0) {Multiple rewards};%
\node[anchor=base,inner sep=0pt,text=sifrNFT,font=\fontfamily{ptm}\fontsize{7.6}{8.8}\selectfont] at (163.0,4.5) {$\Delta$ vs. DiffusionNFT};%
\node[anchor=center,inner sep=0pt,text=sifrRAM,font=\fontfamily{ptm}\fontsize{7.0}{8.2}\selectfont,rotate=90] at (16,59.0) {$\Delta$ vs. RAM};%
\end{pgfinterruptboundingbox}%
\end{tikzpicture}%
\endgroup%

%% file: tables/prompt_sets_cosmos.tex
\begin{table}[!t]
\centering
\begingroup\small
\setlength{\tabcolsep}{2.1pt}
\renewcommand{\arraystretch}{1.06}
\newsavebox{\cosmosgeneralizationbox}
\sbox{\cosmosgeneralizationbox}{%
\begin{tabular}{@{}l@{\hspace{7pt}}*{3}{C{38pt}}!{\hspace{4pt}}*{4}{C{38pt}}@{}}
\toprule[0.7pt]
\multicolumn{1}{@{}c@{\hspace{7pt}}}{\multirow{2}{*}[-2.5pt]{\footnotesize Method}} & \multicolumn{3}{c}{\footnotesize Training rewards} & \multicolumn{4}{c}{\footnotesize Held-out rewards} \\
\cmidrule(r){2-4}\cmidrule(l){5-8}
& \footnotesize PickScore & \footnotesize HPSv2.1 & \footnotesize ClipScore & \footnotesize ImgRwd & \footnotesize HPSv3 & \footnotesize UniRwd2 & \footnotesize Aesthetic \\
\specialrule{0.35pt}{2.25pt}{2.75pt}
\multicolumn{8}{@{}l}{\textit{UniGenBench + physical prompts}} \\[1pt]
DiffusionNFT & 23.12 & 0.313 & 0.294 & 1.08 & 12.03 & 3.376 & 6.01 \\
RAM & \underline{23.50} & \underline{0.330} & \underline{0.296} & \underline{1.14} & \underline{12.66} & \underline{3.401} & \underline{6.08} \\
SIFR & \textbf{23.93} & \textbf{0.346} & \textbf{0.297} & \textbf{1.18} & \textbf{13.11} & \textbf{3.416} & \textbf{6.30} \\
\arrayrulecolor{black!22}\specialrule{0.25pt}{2.25pt}{2.75pt}\arrayrulecolor{black}
\multicolumn{8}{@{}l}{\textit{DrawBench}} \\[1pt]
DiffusionNFT & 23.82 & 0.331 & 0.294 & 1.42 & 12.46 & \underline{3.339} & 5.87 \\
RAM & \underline{24.03} & \underline{0.347} & \underline{0.297} & \underline{1.50} & \underline{13.17} & 3.330 & \underline{5.92} \\
SIFR & \textbf{24.36} & \textbf{0.359} & \textbf{0.299} & \textbf{1.55} & \textbf{13.64} & \textbf{3.356} & \textbf{6.24} \\
\arrayrulecolor{black!22}\specialrule{0.25pt}{2.25pt}{2.75pt}\arrayrulecolor{black}
\multicolumn{8}{@{}l}{\textit{Long captions}} \\[1pt]
DiffusionNFT & 22.79 & 0.297 & 0.274 & 0.84 & 11.95 & 3.554 & 5.79 \\
RAM & \underline{23.19} & \underline{0.312} & \underline{0.277} & \underline{0.89} & \underline{12.54} & \underline{3.586} & \underline{5.90} \\
SIFR & \textbf{23.61} & \textbf{0.326} & \textbf{0.278} & \textbf{0.98} & \textbf{13.22} & \textbf{3.600} & \textbf{6.07} \\
\specialrule{0.7pt}{0.75pt}{0pt}
\end{tabular}}
\usebox{\cosmosgeneralizationbox}
\endgroup
\caption{\textbf{Transfer across prompt sets.} Reward scores for Cosmos 3 Super Text2Image on three prompt sets after $500$ rounds of multi-reward training on Pick-a-Pic.}
\label{tab:cosmos-generalization}
\end{table}

%% file: figures/ablation_guidance.tex
\begingroup%
\input{style/plots.tex}%
\definecolor{sifrInkB}{HTML}{424750}%
\begin{tikzpicture}[x=1pt,y=1pt]
\path[use as bounding box] (0,0) rectangle (396,116);
\definecolor{fixed0}{HTML}{8C98A5}
\definecolor{fixed1}{HTML}{788593}
\definecolor{fixed2}{HTML}{657383}
\definecolor{fixed3}{HTML}{4D5D6E}
\definecolor{fixed4}{HTML}{35495C}
\node[anchor=west,text=sifrInk,font=\fontsize{7.5}{9.0}\selectfont,inner sep=0pt] at (28,106) {(a) Weight sign};
\begin{axis}[sifr axis,scale only axis,anchor=south west,at={(28pt,38pt)},width=72pt,height=60pt,tick label style={font=\fontsize{7.2}{8.5}\selectfont},label style={font=\fontsize{7.8}{9}\selectfont},clip mode=individual,axis line style={black,line width=.5pt},y axis line style={draw=black,line width=.5pt},tick style={black,line width=.5pt},ytick style={draw=black,line width=.5pt},tick label style={text=black},label style={text=black},grid style={black!18,line width=.4pt,dashed},major tick length=2.6pt,minor tick length=1.3pt,xmin=0,xmax=502,ymin=15.8,ymax=25,xtick={0,250,500},ytick={16,20,24},ylabel={Reward},tick label style={font=\fontsize{6.5}{7.5}\selectfont}]
\addplot[fill=sifrBlue,fill opacity=.14,draw=none,no marks] coordinates {
(0,21.297871351242065)
(20,22.849108173592658)
(40,23.167533378948409)
(60,23.381196762847225)
(80,23.481710427020388)
(100,23.621053865819835)
(120,23.683363314310565)
(140,23.756652448484786)
(160,23.835096009318171)
(180,23.85018262048807)
(200,23.917054417967673)
(220,23.987676832536852)
(240,24.048442938819878)
(260,24.105296726183379)
(280,24.15153979961952)
(300,24.18337840843239)
(320,24.219113643438845)
(340,24.254498591672391)
(360,24.268967222034103)
(380,24.29798340861128)
(400,24.327336335051513)
(420,24.325651611482382)
(440,24.355156031680078)
(460,24.359235136952577)
(480,24.373919348463353)
(500,24.423885906030026)
(500,24.477503215979251)
(480,24.457074224407535)
(460,24.424995095286192)
(440,24.387899153002135)
(420,24.370711519722857)
(400,24.363653397690797)
(380,24.325297990797694)
(360,24.311297028403313)
(340,24.284287899085871)
(320,24.25746602269885)
(300,24.216252456028876)
(280,24.175143642178369)
(260,24.13111444795501)
(240,24.076941074038192)
(220,24.022764391243459)
(200,23.970780130982522)
(180,23.929108452993169)
(160,23.860034895197415)
(140,23.785630530208856)
(120,23.716848496755109)
(100,23.634553500742054)
(80,23.539126641537351)
(60,23.475460821661038)
(40,23.278662064522866)
(20,22.941370453294343)
(0,21.297871351242065)
} \closedcycle;
\addplot[sifrBlue,line width=1.2pt,no marks] coordinates {
(0,21.297871351242065)
(20,22.895239313443501)
(40,23.223097721735638)
(60,23.428328792254131)
(80,23.51041853427887)
(100,23.627803683280945)
(120,23.700105905532837)
(140,23.771141489346821)
(160,23.847565452257793)
(180,23.88964553674062)
(200,23.943917274475098)
(220,24.005220611890156)
(240,24.062692006429035)
(260,24.118205587069195)
(280,24.163341720898945)
(300,24.199815432230633)
(320,24.238289833068848)
(340,24.269393245379131)
(360,24.290132125218708)
(380,24.311640699704487)
(400,24.345494866371155)
(420,24.348181565602619)
(440,24.371527592341106)
(460,24.392115116119385)
(480,24.415496786435444)
(500,24.450694561004639)
};
\addplot[sifrRed,line width=.95pt,no marks] coordinates {
(0,21.297871351242065)
(20,21.106285572052002)
(40,20.217836856842041)
(60,19.760204315185547)
(80,19.345790386199951)
(100,18.848931312561035)
(120,18.223668575286865)
(140,17.672070264816284)
(160,17.298444986343384)
(180,17.003288269042969)
(200,16.659512162208557)
(220,16.684394478797913)
(240,16.619772672653198)
(260,16.665972948074341)
(280,16.730267763137817)
(300,16.739549040794373)
(320,16.723938703536987)
(340,16.702927589416504)
(360,16.649911642074585)
(380,16.54081130027771)
(400,16.442052245140076)
(420,16.384215116500854)
(440,16.374197721481323)
(460,16.407601952552795)
(480,16.32624626159668)
(500,16.356568098068237)
};
\addplot[sifrBase,line width=.6pt,dashed,no marks] coordinates {
(0,21.297871351242065)
(500,21.297871351242065)
};
\node[anchor=south east,text=sifrMuted,font=\fontsize{6}{7.5}\selectfont,inner sep=0pt] at (axis cs:480,21.447871351242064) {Pretrained};
\end{axis}
\node[anchor=west,text=sifrInk,font=\fontsize{7.5}{9.0}\selectfont,inner sep=0pt] at (126,106) {(b) Step penalty};
\begin{axis}[sifr axis,scale only axis,anchor=south west,at={(126pt,38pt)},width=74pt,height=60pt,tick label style={font=\fontsize{7.2}{8.5}\selectfont},label style={font=\fontsize{7.8}{9}\selectfont},clip mode=individual,axis line style={black,line width=.5pt},y axis line style={draw=black,line width=.5pt},tick style={black,line width=.5pt},ytick style={draw=black,line width=.5pt},tick label style={text=black},label style={text=black},grid style={black!18,line width=.4pt,dashed},major tick length=2.6pt,minor tick length=1.3pt,xmin=0,xmax=502,ymin=1.43,ymax=1.625,xtick={0,250,500},ytick={1.45,1.50,1.55,1.60},tick label style={font=\fontsize{6.5}{7.5}\selectfont}]
\addplot[fixed0,line width=.95pt,no marks] coordinates {
(20,1.506803035736084)
(40,1.5311468839645386)
(60,1.5441828966140747)
(80,1.5532586574554443)
(100,1.5603442192077637)
(120,1.5665420293807983)
(140,1.570404052734375)
(160,1.5728162527084351)
(180,1.5716725587844849)
(200,1.5762795209884644)
(220,1.581839919090271)
(240,1.5870188474655151)
(260,1.5897600650787354)
(280,1.5898599624633789)
(300,1.5910873413085938)
(320,1.5943472385406494)
(340,1.5960767269134521)
(360,1.6031596660614014)
(380,1.5966315269470215)
(400,1.5935006141662598)
(420,1.591361403465271)
(440,1.5936702489852905)
(460,1.5964145660400391)
(480,1.59587562084198)
(500,1.5925623178482056)
};
\addplot[fixed1,line width=.95pt,no marks] coordinates {
(20,1.5038069486618042)
(40,1.5312910079956055)
(60,1.5482218265533447)
(80,1.557033896446228)
(100,1.5630624294281006)
(120,1.5683516263961792)
(140,1.5738612413406372)
(160,1.5786434412002563)
(180,1.5796482563018799)
(200,1.5845896005630493)
(220,1.589777946472168)
(240,1.5866332054138184)
(260,1.5894885063171387)
(280,1.5878536701202393)
(300,1.5863516330718994)
(320,1.5887635946273804)
(340,1.589047908782959)
(360,1.5881694555282593)
(380,1.5905139446258545)
(400,1.5871469974517822)
(420,1.5893614292144775)
(440,1.5908502340316772)
(460,1.5936946868896484)
(480,1.5904765129089355)
(500,1.5903685092926025)
};
\addplot[fixed2,line width=.95pt,no marks] coordinates {
(20,1.5047507286071777)
(40,1.5306720733642578)
(60,1.5496797561645508)
(80,1.5591464042663574)
(100,1.5701594352722168)
(120,1.5739724636077881)
(140,1.5777552127838135)
(160,1.5828410387039185)
(180,1.5839384794235229)
(200,1.5868295431137085)
(220,1.5915040969848633)
(240,1.5918941497802734)
(260,1.5924500226974487)
(280,1.5909247398376465)
(300,1.5905808210372925)
(320,1.5903980731964111)
(340,1.5925328731536865)
(360,1.5941734313964844)
(380,1.5947896242141724)
(400,1.5937912464141846)
(420,1.5941439867019653)
(440,1.5980262756347656)
(460,1.5970203876495361)
(480,1.595646858215332)
(500,1.5961039066314697)
};
\addplot[fixed3,line width=.95pt,no marks] coordinates {
(20,1.4884481430053711)
(40,1.5159780979156494)
(60,1.5341993570327759)
(80,1.5432939529418945)
(100,1.5503028631210327)
(120,1.5568363666534424)
(140,1.5640344619750977)
(160,1.5695328712463379)
(180,1.57361900806427)
(200,1.5788764953613281)
(220,1.579569935798645)
(240,1.5826001167297363)
(260,1.5871902704238892)
(280,1.5906250476837158)
(300,1.5945446491241455)
(320,1.5976208448410034)
(340,1.6003650426864624)
(360,1.6021409034729004)
(380,1.6014341115951538)
(400,1.6001325845718384)
(420,1.6007180213928223)
(440,1.6012799739837646)
(460,1.6013143062591553)
(480,1.6007606983184814)
(500,1.6009447574615479)
};
\addplot[fixed4,line width=.95pt,no marks] coordinates {
(20,1.4371000528335571)
(40,1.4749152660369873)
(60,1.4983690977096558)
(80,1.5091791152954102)
(100,1.5165799856185913)
(120,1.5217995643615723)
(140,1.527719259262085)
(160,1.5326638221740723)
(180,1.5372920036315918)
(200,1.5420242547988892)
(220,1.5460132360458374)
(240,1.5495773553848267)
(260,1.5529292821884155)
(280,1.5567672252655029)
(300,1.5585595369338989)
(320,1.5627328157424927)
(340,1.5659676790237427)
(360,1.5676995515823364)
(380,1.5704351663589478)
(400,1.5725879669189453)
(420,1.5740023851394653)
(440,1.576289176940918)
(460,1.5763862133026123)
(480,1.578270435333252)
(500,1.5795333385467529)
};
\addplot[sifrBlue,line width=1.2pt,no marks] coordinates {
(20,1.5047711133956909)
(40,1.5329959392547607)
(60,1.5478334426879883)
(80,1.5577373504638672)
(100,1.5721087455749512)
(120,1.5764153003692627)
(140,1.5820020437240601)
(160,1.5873806476593018)
(180,1.5909631252288818)
(200,1.5943005084991455)
(220,1.5961167812347412)
(240,1.5995450019836426)
(260,1.6006886959075928)
(280,1.6026369333267212)
(300,1.6042766571044922)
(320,1.6063301563262939)
(340,1.6058846712112427)
(360,1.6096881628036499)
(380,1.610682487487793)
(400,1.6100649833679199)
(420,1.612265944480896)
(440,1.6139177083969116)
(460,1.611817479133606)
(480,1.6136274337768555)
(500,1.6158890724182129)
};
\draw[sifrBlue,line width=1.2pt] (axis cs:265,1.478) -- (axis cs:315,1.478);
\node[anchor=east,text=black,font=\fontsize{6.5}{8.0}\selectfont,inner sep=0pt] at (axis cs:498,1.478) {SIFR};
\draw[fixed2,line width=.95pt] (axis cs:265,1.456) -- (axis cs:315,1.456);
\node[anchor=east,text=black,font=\fontsize{6.5}{8.0}\selectfont,inner sep=0pt] at (axis cs:498,1.456) {Fixed $\lambda$};
\end{axis}
\draw[sifrBlue,line width=1.2pt] (26,9.5) -- (36,9.5);
\node[anchor=west,text=black,font=\fontsize{6.5}{8.0}\selectfont,inner sep=0pt] at (39,9.5) {Signed + budget};
\draw[sifrRed,line width=.95pt] (104,9.5) -- (114,9.5);
\node[anchor=west,text=black,font=\fontsize{6.5}{8.0}\selectfont,inner sep=0pt] at (117,9.5) {Positive + budget};
\node[anchor=west,text=sifrInk,font=\fontsize{7.5}{9.0}\selectfont,inner sep=0pt] at (238,106) {(c) Multiple rewards};
\begin{axis}[sifr axis,at={(238pt,38pt)},width=68pt,height=60pt,scale only axis,axis line style={black,line width=.5pt},y axis line style={draw=black,line width=.5pt},tick style={black,line width=.5pt},ytick style={draw=black,line width=.5pt},tick label style={text=black,font=\fontfamily{ptm}\fontsize{6.5}{7.5}\selectfont},grid style={black!18,line width=.4pt,dashed},major tick length=2.6pt,xmin=-.45,xmax=2.45,ymin=1.5154652790000001,ymax=1.635310721,xtick={0,1,2},xticklabels={},xtick style={draw=none},ytick={1.53,1.56,1.59,1.62},yticklabels={1.53,1.56,1.59,1.62},ymajorgrids,clip=false]
\draw[sifrBlue,line width=.5pt,opacity=.45] (axis cs:0,1.5989979999999999) -- (axis cs:0,1.620784);
\draw[sifrBlue,line width=1.1pt] (axis cs:-0.3,1.620784) -- (axis cs:0.3,1.620784);
\draw[sifrBlue,line width=1pt,dash pattern=on 2pt off 1.5pt] (axis cs:-0.3,1.5989979999999999) -- (axis cs:0.3,1.5989979999999999);
\addplot[sifrBlue,only marks,mark=*,mark size=2.4pt,mark options={fill=white,draw=sifrBlue,line width=.8pt}] coordinates {(0,1.5989979999999999)};
\addplot[sifrBlue,only marks,mark=*,mark size=1.8pt] coordinates {(0,1.620784)};
\draw[sifrNFT,line width=.5pt,opacity=.45] (axis cs:1,1.529992) -- (axis cs:1,1.561051);
\draw[sifrNFT,line width=1.1pt] (axis cs:0.7,1.561051) -- (axis cs:1.3,1.561051);
\draw[sifrNFT,line width=1pt,dash pattern=on 2pt off 1.5pt] (axis cs:0.7,1.529992) -- (axis cs:1.3,1.529992);
\addplot[sifrNFT,only marks,mark=*,mark size=2.4pt,mark options={fill=white,draw=sifrNFT,line width=.8pt}] coordinates {(1,1.529992)};
\addplot[sifrNFT,only marks,mark=*,mark size=1.8pt] coordinates {(1,1.561051)};
\draw[sifrRAM,line width=.5pt,opacity=.45] (axis cs:2,1.557061) -- (axis cs:2,1.5876170000000001);
\draw[sifrRAM,line width=1.1pt] (axis cs:1.7,1.5876170000000001) -- (axis cs:2.3,1.5876170000000001);
\draw[sifrRAM,line width=1pt,dash pattern=on 2pt off 1.5pt] (axis cs:1.7,1.557061) -- (axis cs:2.3,1.557061);
\addplot[sifrRAM,only marks,mark=*,mark size=2.4pt,mark options={fill=white,draw=sifrRAM,line width=.8pt}] coordinates {(2,1.557061)};
\addplot[sifrRAM,only marks,mark=*,mark size=1.8pt] coordinates {(2,1.5876170000000001)};
\end{axis}
\node[anchor=west,text=sifrInk,font=\fontsize{7.5}{9.0}\selectfont,inner sep=0pt] at (322,106) {(d) OCR};
\begin{axis}[sifr axis,at={(322pt,38pt)},width=68pt,height=60pt,scale only axis,axis line style={black,line width=.5pt},y axis line style={draw=black,line width=.5pt},tick style={black,line width=.5pt},ytick style={draw=black,line width=.5pt},tick label style={text=black,font=\fontfamily{ptm}\fontsize{6.5}{7.5}\selectfont},grid style={black!18,line width=.4pt,dashed},major tick length=2.6pt,xmin=-.45,xmax=2.45,ymin=0.70472883900000005,ymax=1.026777161,xtick={0,1,2},xticklabels={},xtick style={draw=none},ytick={0.8,0.9,1.0},yticklabels={0.80,0.90,1.00},ymajorgrids,clip=false]
\draw[sifrBlue,line width=.5pt,opacity=.45] (axis cs:0,0.98571399999999998) -- (axis cs:0,0.98774099999999998);
\draw[sifrBlue,line width=1.1pt] (axis cs:-0.3,0.98774099999999998) -- (axis cs:0.3,0.98774099999999998);
\draw[sifrBlue,line width=1pt,dash pattern=on 2pt off 1.5pt] (axis cs:-0.3,0.98571399999999998) -- (axis cs:0.3,0.98571399999999998);
\addplot[sifrBlue,only marks,mark=*,mark size=2.4pt,mark options={fill=white,draw=sifrBlue,line width=.8pt}] coordinates {(0,0.98571399999999998)};
\addplot[sifrBlue,only marks,mark=*,mark size=1.8pt] coordinates {(0,0.98774099999999998)};
\draw[sifrNFT,line width=.5pt,opacity=.45] (axis cs:1,0.94466700000000003) -- (axis cs:1,0.97553000000000001);
\draw[sifrNFT,line width=1.1pt] (axis cs:0.7,0.97553000000000001) -- (axis cs:1.3,0.97553000000000001);
\draw[sifrNFT,line width=1pt,dash pattern=on 2pt off 1.5pt] (axis cs:0.7,0.94466700000000003) -- (axis cs:1.3,0.94466700000000003);
\addplot[sifrNFT,only marks,mark=*,mark size=2.4pt,mark options={fill=white,draw=sifrNFT,line width=.8pt}] coordinates {(1,0.94466700000000003)};
\addplot[sifrNFT,only marks,mark=*,mark size=1.8pt] coordinates {(1,0.97553000000000001)};
\draw[sifrRAM,line width=.5pt,opacity=.45] (axis cs:2,0.74376500000000001) -- (axis cs:2,0.98167199999999999);
\draw[sifrRAM,line width=1.1pt] (axis cs:1.7,0.98167199999999999) -- (axis cs:2.3,0.98167199999999999);
\draw[sifrRAM,line width=1pt,dash pattern=on 2pt off 1.5pt] (axis cs:1.7,0.74376500000000001) -- (axis cs:2.3,0.74376500000000001);
\addplot[sifrRAM,only marks,mark=*,mark size=2.4pt,mark options={fill=white,draw=sifrRAM,line width=.8pt}] coordinates {(2,0.74376500000000001)};
\addplot[sifrRAM,only marks,mark=*,mark size=1.8pt] coordinates {(2,0.98167199999999999)};
\end{axis}
\fill[sifrBlue] (227,15) circle[radius=2pt];
\node[anchor=west,text=black,font=\fontsize{6.5}{8.0}\selectfont,inner sep=0pt] at (233,15) {SIFR};
\fill[sifrNFT] (255.8,15) circle[radius=2pt];
\node[anchor=west,text=black,font=\fontsize{6.5}{8.0}\selectfont,inner sep=0pt] at (261.8,15) {DiffusionNFT};
\fill[sifrRAM] (318.2,15) circle[radius=2pt];
\node[anchor=west,text=black,font=\fontsize{6.5}{8.0}\selectfont,inner sep=0pt] at (324.2,15) {RAM};
\draw[sifrInkB,line width=1.1pt] (224,4) -- (236,4);\fill[sifrInkB] (230,4) circle[radius=1.8pt];
\node[anchor=west,text=black,font=\fontsize{6.5}{8.0}\selectfont,inner sep=0pt] at (239,4) {With CFG};
\draw[sifrInkB,line width=1pt,dash pattern=on 2pt off 1.5pt] (282,4) -- (294,4);\filldraw[fill=white,draw=sifrInkB,line width=.8pt] (288,4) circle[radius=1.8pt];
\node[anchor=west,text=black,font=\fontsize{6.5}{8.0}\selectfont,inner sep=0pt] at (297,4) {No CFG};
\draw[sifrBase,dashed,line width=.5pt,rounded corners=5pt] (.5,.5) rectangle (203,115.5);
\draw[sifrBase,dashed,line width=.5pt,rounded corners=5pt] (211,.5) rectangle (395.5,115.5);
\end{tikzpicture}%
\endgroup%

%% file: figures/feedback_control.tex
\begingroup%
\input{style/plots.tex}%
\definecolor{sifrTeal}{HTML}{008575}%
\definecolor{sifrRed}{HTML}{D45347}%
\definecolor{sifrBase}{HTML}{C5CBD3}%
\begin{tikzpicture}[x=1pt,y=1pt]
\path[use as bounding box] (0,0) rectangle (168,118);
\node[anchor=west,text=sifrTeal,font=\fontsize{7.5}{9.0}\selectfont,inner sep=0pt] at (25,109) {$D_{\mathrm{ema}}^{(k)}/\delta$};
\node[anchor=east,text=sifrBlue,font=\fontsize{7.5}{9.0}\selectfont,inner sep=0pt] at (137,109) {$\lambda^{(k+1)}$};
\begin{axis}[sifr axis,scale only axis,anchor=south west,at={(25pt,25pt)},width=112pt,height=74pt,tick label style={font=\fontsize{7.2}{8.5}\selectfont},label style={font=\fontsize{7.8}{9}\selectfont},clip mode=individual,axis line style={black,line width=.5pt},y axis line style={draw=black,line width=.5pt},tick style={black,line width=.5pt},ytick style={draw=black,line width=.5pt},tick label style={text=black},label style={text=black},grid style={black!18,line width=.4pt,dashed},major tick length=2.6pt,minor tick length=1.3pt,xmin=0,xmax=300,xtick={0,150,300},ymode=log,minor y tick num=0,tick label style={font=\fontsize{6.5}{7.5}\selectfont},major tick length=1.8pt,ymin=.4,ymax=35,ytick={.5,1,5,25},yticklabels={0.5,1,5,25},y axis line style={sifrTeal,line width=.5pt},ytick style={sifrTeal},yticklabel style={text=sifrTeal}]
\addplot[sifrTeal,line width=1.05pt,no marks] coordinates {
(1,27.355616266994552)
(2,26.352737375554721)
(3,24.953295465596376)
(4,23.571941280117105)
(5,22.23126303139172)
(6,20.953083337111966)
(7,19.886017152309719)
(8,19.063958913344376)
(9,18.131875504036657)
(10,17.187498093394485)
(11,16.199233914406843)
(12,15.276822358671527)
(13,14.391502251765774)
(14,13.549231739387373)
(15,12.786682702717005)
(16,12.082026379928452)
(17,11.47081674326423)
(18,10.813731635088308)
(19,10.208599006918178)
(20,9.6099866928073219)
(21,9.0602337775632726)
(22,8.5327491221510527)
(23,8.0166297479091018)
(24,7.5412520811349513)
(25,7.1445093659376351)
(26,6.7351013038979284)
(27,6.3457548318341495)
(28,5.9805062469400578)
(29,5.7016976161354362)
(30,5.4959803350783858)
(31,5.2143299646812116)
(32,5.0005376508472379)
(33,4.7988252846153738)
(34,4.6187822045418736)
(35,4.444174114685131)
(36,4.2805507921552701)
(37,4.0905921179566098)
(38,3.8700040762778811)
(39,3.677758030831539)
(40,3.5100474384717395)
(41,3.3082992548646013)
(42,3.1131703813228619)
(43,2.941020323242991)
(44,2.8708951736136141)
(45,2.7442727863749163)
(46,2.6073745085929088)
(47,2.4655889212417663)
(48,2.3339577058168732)
(49,2.201184576594982)
(50,2.0857346555392908)
(51,1.9746768242295556)
(52,1.8695877027271999)
(53,1.7878080205765023)
(54,1.6904695770265954)
(55,1.5988504993491748)
(56,1.5134923555531528)
(57,1.4371182979824584)
(58,1.3642121963308567)
(59,1.2995331261909899)
(60,1.2315074001864117)
(61,1.1678754889634488)
(62,1.1082642092475818)
(63,1.0462378548431328)
(64,0.99304873560182627)
(65,0.9551302267941808)
(66,0.91731951540240486)
(67,0.87237976210610602)
(68,0.83898836576416935)
(69,0.80488578732558724)
(70,0.76308997276128043)
(71,0.72961901954142505)
(72,0.69806972976755666)
(73,0.6616940937528577)
(74,0.63451010237159544)
(75,0.60687223418363889)
(76,0.59028323990974174)
(77,0.57360341913963853)
(78,0.55772665279378308)
(79,0.55078818103575777)
(80,0.53590329288244998)
(81,0.51545259399317489)
(82,0.5043080243967405)
(83,0.52588018760960376)
(84,0.51608930476834003)
(85,0.50983349854467386)
(86,0.52600070421822054)
(87,0.52525074857657161)
(88,0.51590810263439713)
(89,0.53113765421573211)
(90,0.5230812156809066)
(91,0.52330784439903433)
(92,0.51986569753442746)
(93,0.52557161154560195)
(94,0.52184778495042183)
(95,0.51070453136606886)
(96,0.51518900429177816)
(97,0.49639466302912644)
(98,0.50913687264013152)
(99,0.51574200883482257)
(100,0.53621618780966929)
(101,0.54678718156811223)
(102,0.54460208196110405)
(103,0.54397689680944405)
(104,0.56831515262928978)
(105,0.57674096445244671)
(106,0.57973379968700378)
(107,0.59172870973312786)
(108,0.59582055687369939)
(109,0.62584617699011613)
(110,0.62692401201110914)
(111,0.62679292351790183)
(112,0.67666351635246003)
(113,0.70150405249173819)
(114,0.71458436274053305)
(115,0.73814416681443151)
(116,0.76122463580888211)
(117,0.77117073635168065)
(118,0.81440182827623586)
(119,0.84364664842412918)
(120,0.87722498631978696)
(121,0.8750773318857995)
(122,0.87775375311458781)
(123,0.86501886778866732)
(124,0.83803441320742067)
(125,0.83184391059101503)
(126,0.81668491028551427)
(127,0.7992846069929983)
(128,0.78322742786839372)
(129,0.7784560969892137)
(130,0.7974081343850179)
(131,0.84894784119956612)
(132,0.9082221280712518)
(133,0.9116150348994253)
(134,0.91881426646826148)
(135,0.90228827964061631)
(136,0.88834575972868191)
(137,0.88878691090822437)
(138,0.88269635098153232)
(139,0.86785658935720988)
(140,0.85352695394401767)
(141,0.8355149275183491)
(142,0.82286334104529701)
(143,0.81294034274172899)
(144,0.80030040024453153)
(145,0.7805925339533244)
(146,0.75989482513407469)
(147,0.76675700392373214)
(148,0.79374533562028082)
(149,0.79160388607854237)
(150,0.78825594836133217)
(151,0.80675395039552866)
(152,0.81677796689686377)
(153,0.79469924352218435)
(154,0.7941858961156596)
(155,0.78834838288841058)
(156,0.77887647310892527)
(157,0.78047277396375292)
(158,0.77219614195133879)
(159,0.76544510296810631)
(160,0.76725602519590341)
(161,0.78706215930712398)
(162,0.80074415759659223)
(163,0.83636043022912143)
(164,0.85067499966432025)
(165,0.84423860315307631)
(166,0.84196861121333033)
(167,0.86423141214096133)
(168,0.87410231092302837)
(169,0.85563692952658432)
(170,0.84943183726513249)
(171,0.86459874197165487)
(172,0.88600789880655784)
(173,0.88210530786529473)
(174,0.88850431308197819)
(175,0.90280489693296784)
(176,0.91875166441167055)
(177,0.92531010102968991)
(178,0.95646693056542464)
(179,1.02071896442653)
(180,1.0771306793425817)
(181,1.1480877229690878)
(182,1.2680076827721558)
(183,1.2984731329975678)
(184,1.3588885554406762)
(185,1.383067150855793)
(186,1.3652039752808987)
(187,1.3443228552143089)
(188,1.334759562980903)
(189,1.326412378383699)
(190,1.333985834960099)
(191,1.4516268802497285)
(192,1.5978976541863759)
(193,1.7085934380598897)
(194,1.7250168073852667)
(195,1.7472015484312897)
(196,1.7306491468417526)
(197,1.6999348777199863)
(198,1.6924901324633355)
(199,1.6414370124349063)
(200,1.6063883538819563)
(201,1.5957418799509888)
(202,1.5931222297337342)
(203,1.6721264647718517)
(204,1.6449983804894632)
(205,1.639352248003985)
(206,1.6795304718879793)
(207,1.6673860057340779)
(208,1.6255326154493448)
(209,1.5500822812957646)
(210,1.5027303791268769)
(211,1.4422942415153657)
(212,1.3685483202504569)
(213,1.3132843245165575)
(214,1.2440035877788354)
(215,1.1828511204918339)
(216,1.1143376493008987)
(217,1.0509657865185504)
(218,0.99254998275277007)
(219,0.9469088002933147)
(220,0.91449042316055984)
(221,0.86783013697009126)
(222,0.81839279348677207)
(223,0.78608830736385416)
(224,0.74982760144037008)
(225,0.71538998728405301)
(226,0.68858238558264007)
(227,0.66944446534472335)
(228,0.65161304341850246)
(229,0.64310729145163259)
(230,0.62553808611146566)
(231,0.60495707760760853)
(232,0.6083422800278806)
(233,0.60225077981333774)
(234,0.58301232535328895)
(235,0.57080429913308162)
(236,0.56558615983526084)
(237,0.57006916012417774)
(238,0.55777601112642039)
(239,0.54776054192890444)
(240,0.53995520294073041)
(241,0.57180700395389228)
(242,0.58448537960383395)
(243,0.58381053433939845)
(244,0.58561205020037677)
(245,0.57814919019988997)
(246,0.56577146283126645)
(247,0.55133826754697068)
(248,0.55005838360811932)
(249,0.54888978132393118)
(250,0.57052505162708045)
(251,0.60955696401210746)
(252,0.614281651943972)
(253,0.61888154529998596)
(254,0.62267281269987196)
(255,0.62590171837852659)
(256,0.63627197371262378)
(257,0.63833961353600155)
(258,0.63793638687613252)
(259,0.64402832710419611)
(260,0.64254440508776556)
(261,0.66259026992189396)
(262,0.69570545695967545)
(263,0.69848031452284354)
(264,0.70236491224093622)
(265,0.71942701280779187)
(266,0.71306342167514203)
(267,0.71351429022988389)
(268,0.71133701122719106)
(269,0.72692386829977917)
(270,0.73457410391039546)
(271,0.76058500575709465)
(272,0.77774369194306825)
(273,0.79383432766324502)
(274,0.82908596446556226)
(275,0.82798682303343141)
(276,0.80990752673404043)
(277,0.84787222536584861)
(278,0.83038075200496997)
(279,0.83507307013885568)
(280,0.84303787920557238)
(281,0.87721397408077917)
(282,0.88611275758402896)
(283,0.911480810991249)
(284,0.90954511099482749)
(285,0.94253533664312372)
(286,0.92434119676723914)
(287,0.92070894844411255)
(288,0.96105922951418687)
(289,0.99007051724342887)
(290,1.0396185813719083)
(291,1.0446908691123018)
(292,1.0849557118827706)
(293,1.1096594952278136)
(294,1.13675909136564)
(295,1.1627851820969108)
(296,1.1585980881828541)
(297,1.2100710291386982)
(298,1.3151528593734847)
(299,1.2957299080465337)
(300,1.270422522383758)
};
\addplot[sifrMuted,line width=.65pt,dashed,no marks] coordinates {
(0,1)
(300,1)
};
\end{axis}
\begin{axis}[sifr axis,scale only axis,anchor=south west,at={(25pt,25pt)},width=112pt,height=74pt,tick label style={font=\fontsize{7.2}{8.5}\selectfont},label style={font=\fontsize{7.8}{9}\selectfont},clip mode=individual,axis line style={black,line width=.5pt},y axis line style={draw=black,line width=.5pt},tick style={black,line width=.5pt},ytick style={draw=black,line width=.5pt},tick label style={text=black},label style={text=black},grid style={black!18,line width=.4pt,dashed},major tick length=2.6pt,minor tick length=1.3pt,xmin=0,xmax=300,xtick={0,150,300},ymode=log,minor y tick num=0,tick label style={font=\fontsize{6.5}{7.5}\selectfont},major tick length=1.8pt,ymin=.008,ymax=.65,ytick={.01,.05,.1,.5},yticklabels={0.01,0.05,0.1,0.5},axis y line*=right,axis x line=none,xtick=\empty,ymajorgrids=false,y axis line style={sifrBlue,line width=.5pt},ytick style={sifrBlue},yticklabel style={text=sifrBlue}]
\addplot[sifrBlue,line width=1.05pt,no marks] coordinates {
(0,0.010000003501772881)
(1,0.010717734625362936)
(2,0.011486983549970355)
(3,0.012311444133449168)
(4,0.013195079107728949)
(5,0.014142135623730958)
(6,0.015157165665103988)
(7,0.016245047927124719)
(8,0.017411011265922493)
(9,0.01866065983073616)
(10,0.020000000000000011)
(11,0.021435469250725876)
(12,0.022973967099940713)
(13,0.024622888266898339)
(14,0.026390158215457902)
(15,0.028284271247461919)
(16,0.030314331330207979)
(17,0.032490095854249439)
(18,0.034822022531844986)
(19,0.03732131966147232)
(20,0.040000000000000029)
(21,0.042870938501451752)
(22,0.045947934199881427)
(23,0.049245776533796685)
(24,0.052780316430915804)
(25,0.056568542494923837)
(26,0.060628662660415965)
(27,0.064980191708498891)
(28,0.069644045063689985)
(29,0.07464263932294464)
(30,0.080000000000000057)
(31,0.085741877002903519)
(32,0.091895868399762867)
(33,0.098491553067593371)
(34,0.10556063286183162)
(35,0.11313708498984769)
(36,0.12125732532083194)
(37,0.12996038341699778)
(38,0.13928809012737997)
(39,0.14928527864588931)
(40,0.16000000000000014)
(41,0.17148375400580704)
(42,0.18379173679952576)
(43,0.19698310613518677)
(44,0.21112126572366327)
(45,0.22627416997969541)
(46,0.24251465064166391)
(47,0.25992076683399562)
(48,0.27857618025475994)
(49,0.29857055729177862)
(50,0.32000000000000028)
(51,0.34236603587628478)
(52,0.36363683789005313)
(53,0.3840460096967851)
(54,0.40287327821713254)
(55,0.41994813682677057)
(56,0.43516435954308863)
(57,0.44855106366242592)
(58,0.4600190009480955)
(59,0.46966977868772131)
(60,0.47726730386922606)
(61,0.48285333900340788)
(62,0.48649044674224984)
(63,0.48649044674224984)
(64,0.48649044674224984)
(65,0.48649044674224984)
(66,0.48371035633059778)
(67,0.47945034320985996)
(68,0.4741291976370251)
(69,0.4677601042645384)
(70,0.4601415755320245)
(71,0.4515981942009813)
(72,0.44224526308768292)
(73,0.43199544471027079)
(74,0.4211888105789624)
(75,0.40986657261298171)
(76,0.39839033758790293)
(77,0.38678799005125392)
(78,0.37511050449940669)
(79,0.36361065667004761)
(80,0.35209989862400332)
(81,0.34047056362454364)
(82,0.328971105565313)
(83,0.31833568548978947)
(84,0.30783511765841565)
(85,0.29755186755299229)
(86,0.28793461610827381)
(87,0.27861372280995378)
(88,0.26942003308485279)
(89,0.26080488527652923)
(90,0.25232427622443371)
(91,0.24412326650701707)
(92,0.23613245900050217)
(93,0.22849356420711078)
(94,0.22104472543320983)
(95,0.21367361369423851)
(96,0.20661251764308639)
(97,0.19952466877479205)
(98,0.19285022327428447)
(99,0.1864844087308723)
(100,0.1805848214114511)
(101,0.17500005240508648)
(102,0.16956231427317514)
(103,0.16428642235247479)
(104,0.15944344280519102)
(105,0.15483363037832118)
(106,0.15038829081825222)
(107,0.14619207591043121)
(108,0.14215325857966749)
(109,0.13851399876176368)
(110,0.13497799129330326)
(111,0.13153105644366017)
(112,0.1286159738068697)
(113,0.12598222867781209)
(114,0.12351435082620109)
(115,0.12129273084716176)
(116,0.11930177888828709)
(117,0.11742443310130803)
(118,0.11592348010892792)
(119,0.11467393240830744)
(120,0.11370218471074821)
(121,0.11272189012246818)
(122,0.11177078051647822)
(123,0.11072990995204401)
(124,0.1094937416241049)
(125,0.10822492513544779)
(126,0.10685847205077934)
(127,0.10538209414931964)
(128,0.10381050887985779)
(129,0.10222854602179571)
(130,0.10080302408238567)
(131,0.09975310850138476)
(132,0.099120537585534033)
(133,0.09851514394084282)
(134,0.097962320076733139)
(135,0.097301076745598236)
(136,0.096550942749373175)
(137,0.095809521489056762)
(138,0.095033665253742891)
(139,0.094167180323496469)
(140,0.093215962527092189)
(141,0.092159220827824434)
(142,0.091034591874977699)
(143,0.08986185776105711)
(144,0.088626548387288465)
(145,0.087288898453300789)
(146,0.085848186818458627)
(147,0.084471423544813587)
(148,0.083272370739627191)
(149,0.082078154127959629)
(150,0.080882292052261384)
(151,0.079806113978434776)
(152,0.078798986430642287)
(153,0.077685588981151674)
(154,0.076585198238588337)
(155,0.075469850956573492)
(156,0.07432193546644883)
(157,0.073199578937053172)
(158,0.072052823412846204)
(159,0.070890852252402783)
(160,0.069756375321462297)
(161,0.068734351259332346)
(162,0.067791561805471912)
(163,0.067026971588875814)
(164,0.066336792246468504)
(165,0.065624435686343935)
(166,0.064909514838928345)
(167,0.064301532159937017)
(168,0.063742842067554345)
(169,0.063108180900507779)
(170,0.062452971728594528)
(171,0.061869573816033419)
(172,0.061382648088187074)
(173,0.060883083034380593)
(174,0.060414374276503459)
(175,0.060008727511660372)
(176,0.059671725870148198)
(177,0.059671725870148198)
(178,0.059671725870148198)
(179,0.059671725870148198)
(180,0.059991602613761845)
(181,0.060610567239034109)
(182,0.061747045716963926)
(183,0.063037810484501092)
(184,0.064625622922708098)
(185,0.066364558959462694)
(186,0.068065955842921744)
(187,0.069710002515621594)
(188,0.07134644945455608)
(189,0.072979075442611133)
(190,0.074688258391675214)
(191,0.077063309562683271)
(192,0.080324155665613947)
(193,0.084367845189676119)
(194,0.088716037888125696)
(195,0.093431892470966449)
(196,0.098285596227797853)
(197,0.10317156440963328)
(198,0.10824455167852307)
(199,0.1131658068339662)
(200,0.11802372908043542)
(201,0.12299938785777079)
(202,0.12816153733976521)
(203,0.13427363031239614)
(204,0.14041293486476067)
(205,0.14677548947681912)
(206,0.15385423064465165)
(207,0.1611386658252475)
(208,0.16827909648474132)
(209,0.17481926832261147)
(210,0.18101851200603461)
(211,0.18665403123561583)
(212,0.19148368819780062)
(213,0.19568727327808108)
(214,0.19902508665771779)
(215,0.2015636382788136)
(216,0.20316743380922356)
(217,0.20316743380922356)
(218,0.20316743380922356)
(219,0.20316743380922356)
(220,0.20196680737301664)
(221,0.200124975155553)
(222,0.19762158098391383)
(223,0.19471301687263948)
(224,0.19136567700021154)
(225,0.18762747369796159)
(226,0.18362077951931366)
(227,0.17946142571407575)
(228,0.17517963665966751)
(229,0.17089921985094761)
(230,0.16652047972078896)
(231,0.1620226296050522)
(232,0.15768326513218522)
(233,0.15339533762744459)
(234,0.1490251541964511)
(235,0.14465701589602575)
(236,0.14036613516374644)
(237,0.13624486230929561)
(238,0.13213195652658244)
(239,0.12805428087638077)
(240,0.12403532052308029)
(241,0.12040803796808118)
(242,0.11698959654408651)
(243,0.11366288939651391)
(244,0.11044457076214831)
(245,0.10726187809008857)
(246,0.10408156540290811)
(247,0.10089456018626652)
(248,0.097796465538864188)
(249,0.094785823703244315)
(250,0.092005736109802255)
(251,0.089549135547742229)
(252,0.08718667561193981)
(253,0.084913610991648361)
(254,0.082721543571909306)
(255,0.080604102995813928)
(256,0.078597339344317965)
(257,0.076651521919517393)
(258,0.074751787370744366)
(259,0.072929924893836331)
(260,0.071145146941612086)
(261,0.069500549086724331)
(262,0.068049988670604966)
(263,0.0666425198903252)
(264,0.065281736978758609)
(265,0.064024414100008198)
(266,0.062763616651690862)
(267,0.061529570273043975)
(268,0.060310684857819923)
(269,0.059179848557153331)
(270,0.058101016929596544)
(271,0.057144787799406176)
(272,0.056271182650114736)
(273,0.055472768084435367)
(274,0.054819467488344022)
(275,0.054169733618244703)
(276,0.053460663723039911)
(277,0.052899899134756789)
(278,0.052281591063509984)
(279,0.051687318337795413)
(280,0.051128019462180339)
(281,0.050694721575285796)
(282,0.05029610969856807)
(283,0.049988453537465144)
(284,0.049676013675567121)
(285,0.049676013675567121)
(286,0.049416181227230217)
(287,0.049145333050154956)
(288,0.049145333050154956)
(289,0.049145333050154956)
(290,0.049145333050154956)
(291,0.049145333050154956)
(292,0.049435588020503761)
(293,0.049812780472230575)
(294,0.050287221794101208)
(295,0.050857846326013909)
(296,0.051420020253468139)
(297,0.052174225567733677)
(298,0.053326497233007204)
(299,0.054430887445934911)
(300,0.055460776299773126)
};
\end{axis}
\node[anchor=center,text=sifrInk,font=\fontsize{7}{8.5}\selectfont,inner sep=0pt] at (81,3) {Training round};
\end{tikzpicture}%
\endgroup%

%% file: sections/appendix_qualitative.tex
\begingroup
\input{style/plots.tex}
\setlength{\intextsep}{2pt}
\setlength{\abovecaptionskip}{3pt}
\setlength{\belowcaptionskip}{0pt}

\section{Additional qualitative examples}
\label{sec:additional-qualitative}
\tikzset{gallerychip/.style={anchor=south west,inner sep=1pt,rounded corners=0.8pt,fill=white,fill opacity=0.92,text opacity=1,font=\fontfamily{ptm}\fontsize{7}{8}\selectfont}}

\begin{figure}[H]
\centering
\begin{tikzpicture}[x={0.96\textwidth/396},y={0.96\textwidth/396}]
\node[anchor=south west,inner sep=0pt,outer sep=0pt] at (0,0) {\includegraphics[page=1,width=0.96\textwidth]{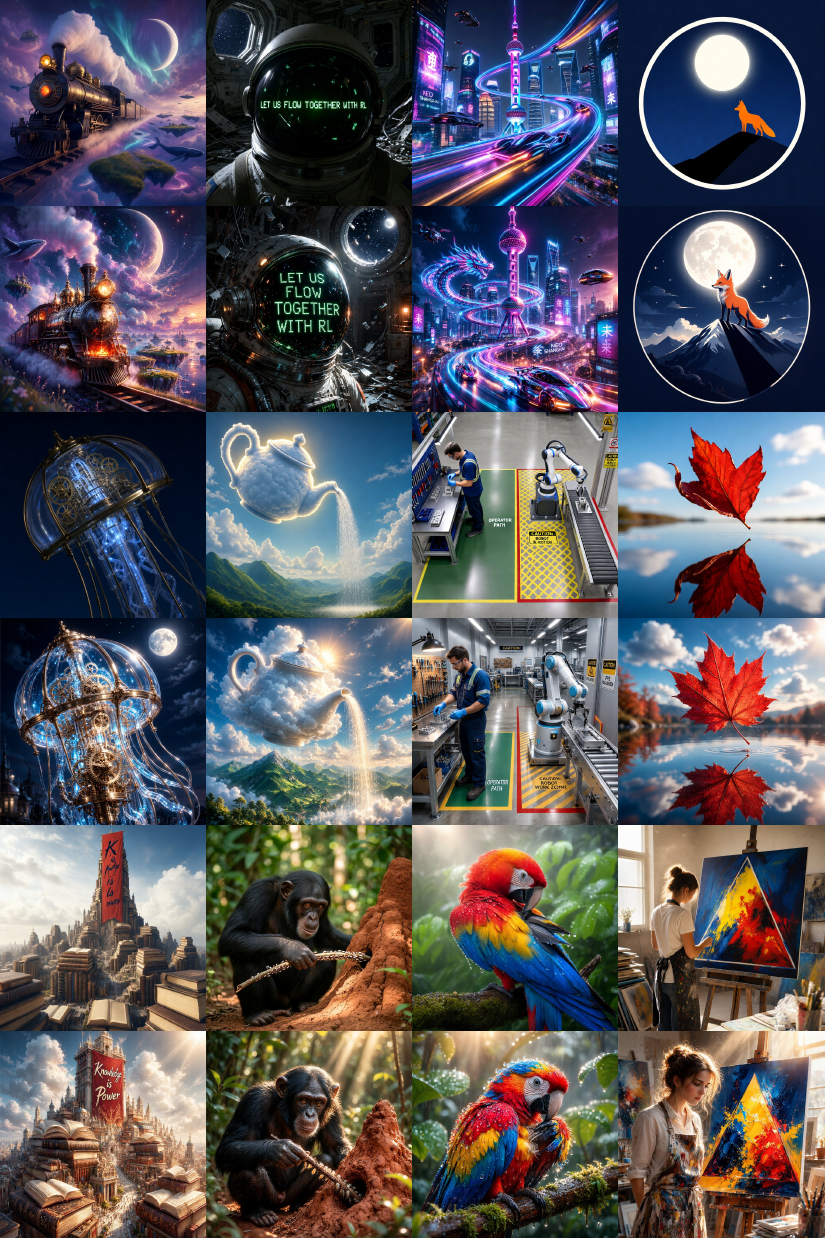}};
\node[gallerychip,text=sifrMuted] at (2.5000,497.5000) {15.58};
\node[gallerychip,text=sifrBlue] at (2.5000,398.5000) {16.06 (+0.48)};
\node[gallerychip,text=sifrMuted] at (101.5000,497.5000) {11.75};
\node[gallerychip,text=sifrBlue] at (101.5000,398.5000) {13.30 (+1.55)};
\node[gallerychip,text=sifrMuted] at (200.5000,497.5000) {11.71};
\node[gallerychip,text=sifrBlue] at (200.5000,398.5000) {12.89 (+1.18)};
\node[gallerychip,text=sifrMuted] at (299.5000,497.5000) {7.64};
\node[gallerychip,text=sifrBlue] at (299.5000,398.5000) {11.36 (+3.72)};
\node[gallerychip,text=sifrMuted] at (2.5000,299.5000) {9.29};
\node[gallerychip,text=sifrBlue] at (2.5000,200.5000) {11.99 (+2.70)};
\node[gallerychip,text=sifrMuted] at (101.5000,299.5000) {9.36};
\node[gallerychip,text=sifrBlue] at (101.5000,200.5000) {11.38 (+2.02)};
\node[gallerychip,text=sifrMuted] at (200.5000,299.5000) {13.16};
\node[gallerychip,text=sifrBlue] at (200.5000,200.5000) {15.11 (+1.95)};
\node[gallerychip,text=sifrMuted] at (299.5000,299.5000) {8.76};
\node[gallerychip,text=sifrBlue] at (299.5000,200.5000) {10.90 (+2.13)};
\node[gallerychip,text=sifrMuted] at (2.5000,101.5000) {11.32};
\node[gallerychip,text=sifrBlue] at (2.5000,2.5000) {14.83 (+3.51)};
\node[gallerychip,text=sifrMuted] at (101.5000,101.5000) {13.71};
\node[gallerychip,text=sifrBlue] at (101.5000,2.5000) {14.37 (+0.66)};
\node[gallerychip,text=sifrMuted] at (200.5000,101.5000) {12.51};
\node[gallerychip,text=sifrBlue] at (200.5000,2.5000) {13.92 (+1.41)};
\node[gallerychip,text=sifrMuted] at (299.5000,101.5000) {12.80};
\node[gallerychip,text=sifrBlue] at (299.5000,2.5000) {14.77 (+1.97)};
\end{tikzpicture}
\small
\caption{Cosmos 3 Super Text2Image after multi-reward training. Pretrained (top) and SIFR (bottom) share prompts and initial noise. Labels: HPSv3 and SIFR's gain.}
\label{fig:additional-qualitative-1}
\end{figure}
\clearpage

\begin{figure}[H]
\centering
\begin{tikzpicture}[x={0.96\textwidth/396},y={0.96\textwidth/396}]
\node[anchor=south west,inner sep=0pt,outer sep=0pt] at (0,0) {\includegraphics[page=2,width=0.96\textwidth]{figures/qualitative_gallery.pdf}};
\node[gallerychip,text=sifrMuted] at (2.5000,497.5000) {14.15};
\node[gallerychip,text=sifrBlue] at (2.5000,398.5000) {15.07 (+0.92)};
\node[gallerychip,text=sifrMuted] at (101.5000,497.5000) {11.63};
\node[gallerychip,text=sifrBlue] at (101.5000,398.5000) {14.40 (+2.77)};
\node[gallerychip,text=sifrMuted] at (200.5000,497.5000) {10.46};
\node[gallerychip,text=sifrBlue] at (200.5000,398.5000) {13.43 (+2.97)};
\node[gallerychip,text=sifrMuted] at (299.5000,497.5000) {10.84};
\node[gallerychip,text=sifrBlue] at (299.5000,398.5000) {13.05 (+2.20)};
\node[gallerychip,text=sifrMuted] at (2.5000,299.5000) {13.74};
\node[gallerychip,text=sifrBlue] at (2.5000,200.5000) {15.44 (+1.70)};
\node[gallerychip,text=sifrMuted] at (101.5000,299.5000) {11.78};
\node[gallerychip,text=sifrBlue] at (101.5000,200.5000) {13.23 (+1.45)};
\node[gallerychip,text=sifrMuted] at (200.5000,299.5000) {11.27};
\node[gallerychip,text=sifrBlue] at (200.5000,200.5000) {11.91 (+0.64)};
\node[gallerychip,text=sifrMuted] at (299.5000,299.5000) {13.37};
\node[gallerychip,text=sifrBlue] at (299.5000,200.5000) {13.73 (+0.36)};
\node[gallerychip,text=sifrMuted] at (2.5000,101.5000) {12.46};
\node[gallerychip,text=sifrBlue] at (2.5000,2.5000) {12.57 (+0.11)};
\node[gallerychip,text=sifrMuted] at (101.5000,101.5000) {10.81};
\node[gallerychip,text=sifrBlue] at (101.5000,2.5000) {14.45 (+3.64)};
\node[gallerychip,text=sifrMuted] at (200.5000,101.5000) {10.67};
\node[gallerychip,text=sifrBlue] at (200.5000,2.5000) {12.55 (+1.88)};
\node[gallerychip,text=sifrMuted] at (299.5000,101.5000) {11.08};
\node[gallerychip,text=sifrBlue] at (299.5000,2.5000) {12.20 (+1.12)};
\end{tikzpicture}
\small
\caption{More multi-reward examples, continuing \cref{fig:additional-qualitative-1}. Pretrained (top) and SIFR (bottom) share prompts and initial noise. Labels: HPSv3 and SIFR's gain.}
\label{fig:additional-qualitative-2}
\end{figure}
\clearpage

\begin{figure}[H]
\centering
\begin{tikzpicture}[x={0.96\textwidth/396},y={0.96\textwidth/396}]
\node[anchor=south west,inner sep=0pt,outer sep=0pt] at (0,0) {\includegraphics[page=3,width=0.96\textwidth]{figures/qualitative_gallery.pdf}};
\node[gallerychip,text=sifrMuted] at (2.5000,497.5000) {13.90};
\node[gallerychip,text=sifrBlue] at (2.5000,398.5000) {15.79 (+1.88)};
\node[gallerychip,text=sifrMuted] at (101.5000,497.5000) {13.73};
\node[gallerychip,text=sifrBlue] at (101.5000,398.5000) {15.32 (+1.60)};
\node[gallerychip,text=sifrMuted] at (200.5000,497.5000) {12.27};
\node[gallerychip,text=sifrBlue] at (200.5000,398.5000) {15.05 (+2.77)};
\node[gallerychip,text=sifrMuted] at (299.5000,497.5000) {10.91};
\node[gallerychip,text=sifrBlue] at (299.5000,398.5000) {13.72 (+2.81)};
\node[gallerychip,text=sifrMuted] at (2.5000,299.5000) {13.26};
\node[gallerychip,text=sifrBlue] at (2.5000,200.5000) {14.30 (+1.04)};
\node[gallerychip,text=sifrMuted] at (101.5000,299.5000) {10.37};
\node[gallerychip,text=sifrBlue] at (101.5000,200.5000) {12.50 (+2.13)};
\node[gallerychip,text=sifrMuted] at (200.5000,299.5000) {11.83};
\node[gallerychip,text=sifrBlue] at (200.5000,200.5000) {14.69 (+2.86)};
\node[gallerychip,text=sifrMuted] at (299.5000,299.5000) {7.78};
\node[gallerychip,text=sifrBlue] at (299.5000,200.5000) {14.43 (+6.65)};
\node[gallerychip,text=sifrMuted] at (2.5000,101.5000) {11.33};
\node[gallerychip,text=sifrBlue] at (2.5000,2.5000) {13.81 (+2.48)};
\node[gallerychip,text=sifrMuted] at (101.5000,101.5000) {12.82};
\node[gallerychip,text=sifrBlue] at (101.5000,2.5000) {15.23 (+2.41)};
\node[gallerychip,text=sifrMuted] at (200.5000,101.5000) {10.57};
\node[gallerychip,text=sifrBlue] at (200.5000,2.5000) {12.46 (+1.89)};
\node[gallerychip,text=sifrMuted] at (299.5000,101.5000) {12.14};
\node[gallerychip,text=sifrBlue] at (299.5000,2.5000) {14.49 (+2.34)};
\end{tikzpicture}
\small
\caption{More multi-reward examples, continuing \cref{fig:additional-qualitative-1}. Pretrained (top) and SIFR (bottom) share prompts and initial noise. Labels: HPSv3 and SIFR's gain.}
\label{fig:additional-qualitative-3}
\end{figure}
\endgroup

%% file: style/plots.tex
\definecolor{sifrBlue}{HTML}{0071E3}%
\definecolor{sifrNFT}{HTML}{D55E00}%
\definecolor{sifrRAM}{HTML}{009E73}%
\definecolor{sifrInk}{HTML}{1D1D1F}%
\definecolor{sifrMuted}{HTML}{69717D}%
\definecolor{sifrGrid}{HTML}{E9ECF0}%
\definecolor{sifrBase}{HTML}{C5CBD3}%
\definecolor{sifrRed}{HTML}{D45347}%
\definecolor{sifrDeepRed}{HTML}{B2182B}%
\definecolor{sifrTeal}{HTML}{008575}%
\def\sifrTickFont{\fontfamily{ptm}\fontsize{7.2}{8.5}\selectfont}%
\def\sifrLabelFont{\fontfamily{ptm}\fontsize{8}{9.3}\selectfont}%
\def\sifrTitleFont{\fontfamily{ptm}\fontsize{8.7}{10}\selectfont}%
\tikzset{%
  sifr text/.style={font=\sifrTickFont,text=sifrInk,inner sep=0pt},%
  sifr title/.style={sifr text,font=\sifrTitleFont},%
  sifr line/.style={line cap=round,line join=round,line width=0.8pt},%
  sifr ours/.style={sifr line,sifrBlue,line width=1.1pt},%
  sifr nft/.style={sifr line,sifrNFT},%
  sifr ram/.style={sifr line,sifrRAM},%
}%
\pgfplotsset{sifr axis/.style={%
  scale only axis,anchor=south west,%
  axis lines*=left,axis line style={black!25,line width=0.35pt},%
  y axis line style={draw=none},ytick style={draw=none},%
  tick style={black!40,line width=0.35pt},tick align=outside,%
  major tick length=1.8pt,%
  tick label style={font=\sifrTickFont,text=sifrMuted},%
  label style={font=\sifrLabelFont,text=sifrInk},%
  title style={font=\sifrTitleFont,text=sifrInk},%
  legend style={font=\sifrTickFont,draw=none,fill=none,cells={anchor=west},%
    inner sep=1pt,column sep=8pt},%
  scaled ticks=false,/pgf/number format/1000 sep={},%
  yticklabel style={/pgf/number format/fixed},%
  ymajorgrids,grid style={sifrGrid,line width=0.35pt},%
  every axis plot/.append style={line cap=round,line join=round},%
  clip=true,clip marker paths=false,%
}}%
\ignorespaces

%% file: figures/collapse_strip.tex
\begingroup\input{style/plots.tex}
\colorlet{cnn}{sifrRed}
\colorlet{csna}{sifrTeal}
\colorlet{csifr}{sifrBlue}
\colorlet{stripink}{sifrInk}
\colorlet{stripmuted}{sifrMuted}
\colorlet{stripgrid}{sifrGrid}
\colorlet{stripaxisgray}{sifrBase}
\pgfplotstableread[col sep=comma]{figures/data/collapse_strip_curves.csv}\stripcurves
\pgfmathsetmacro\stripW{396}
\pgfmathsetmacro\stripCols{6}
\pgfmathsetmacro\stripLabelW{44}
\pgfmathsetmacro\stripLabelGap{4}
\pgfmathsetmacro\stripCell{52}
\pgfmathsetmacro\stripGut{2}
\pgfmathsetmacro\stripRowGap{8}
\pgfmathsetmacro\stripBoxPad{2.5}
\pgfmathsetmacro\stripBoxPadX{4}
\pgfmathsetmacro\stripChartLeft{19}
\pgfmathsetmacro\stripChartW{95}
\pgfmathsetmacro\stripChartSpan{367}
\pgfmathsetmacro\stripChartGap{(\stripChartSpan-3*\stripChartW)/2}
\pgfmathsetmacro\stripChartH{58}
\pgfmathsetmacro\stripChartTop{-35}
\pgfmathsetmacro\stripLegendY{-13}
\pgfmathsetmacro\stripXLabelY{-109}
\pgfmathsetmacro\stripFrameTop{-7.5}
\pgfmathsetmacro\stripFrameBottom{-119}
\pgfmathsetmacro\stripBlockW{\stripCols*\stripCell+(\stripCols-1)*\stripGut}
\pgfmathsetmacro\stripMargin{(\stripW-\stripLabelW-\stripLabelGap-\stripBlockW)/2}
\pgfmathsetmacro\stripXa{\stripMargin+\stripLabelW+\stripLabelGap}
\pgfmathsetmacro\stripH{3*\stripCell+2*\stripRowGap}
\pgfmathsetmacro\stripAxisY{\stripH+\stripBoxPad+4}
\pgfmathsetmacro\stripChartY{\stripChartTop-\stripChartH}
\pgfmathsetmacro\stripTitleY{\stripChartTop+4}
\def\stripTextFont{\fontfamily{ptm}\fontsize{7.2}{8.6}\selectfont}
\def\stripTitleFont{\fontfamily{ptm}\fontsize{7.8}{9.2}\selectfont}
\tikzset{striphead/.style={font=\stripTextFont, text=stripink, inner sep=0pt},
  striptitle/.style={font=\stripTitleFont, text=stripink, anchor=base west, inner sep=0pt},
  striplabel/.style={font=\stripTextFont, align=right, anchor=west, text width=\stripLabelW pt, inner sep=0pt},
  stripchip/.style={anchor=south west, text=stripink, fill=white, inner sep=0.9pt, font=\fontfamily{ptm}\fontsize{7}{8}\selectfont},
  striplegend/.style={font=\stripTextFont, text=stripink, anchor=north, inner sep=0pt},
  striprowbox/.style={draw=stripaxisgray, dashed, line width=0.5pt, rounded corners=3pt}}
\pgfplotsset{stripaxis/.style={sifr axis, width=\stripChartW pt, height=\stripChartH pt, anchor=south west, axis lines*=left,
  axis line style={stripaxisgray, line width=0.4pt}, tick style={stripmuted, line width=0.4pt}, tick align=outside, major tick length=1.8pt,
  y axis line style={draw=stripaxisgray, line width=0.4pt}, ytick style={draw=stripmuted, line width=0.4pt},
  tick label style={font=\stripTextFont, text=stripink}, yticklabel style={inner xsep=1pt},
  xmin=0, xmax=400, xtick={0,100,200,300,400}, xminorticks=false, yminorticks=false,
  ymajorgrids, grid style={stripgrid, line width=0.35pt}},
  stripcurve/.style={line width=0.95pt, line cap=round, line join=round},
  stripours/.style={line width=1.2pt, line cap=round, line join=round}}
\begin{tikzpicture}[x=1pt,y=1pt]
\draw[stripink, line width=0.9pt, ->] (\stripXa,\stripAxisY) -- ({\stripXa+\stripBlockW+6},\stripAxisY);
\node[striphead, anchor=east] at ({\stripXa-3},{\stripAxisY+2.5}) {Round};
\foreach \rd [count=\c from 0] in {0,10,30,100,200,400} {
  \draw[stripink, line width=0.7pt] ({\stripXa+\c*(\stripCell+\stripGut)+\stripCell/2},\stripAxisY) -- ++(0,2.2);
  \node[striphead, anchor=base] at ({\stripXa+\c*(\stripCell+\stripGut)+\stripCell/2}, {\stripAxisY+4}) {\rd}; }
\pgfmathsetmacro\rowY{\stripH-1*\stripCell-0*\stripRowGap}
\node[striplabel, text=cnn] at (\stripMargin, {\rowY+\stripCell/2}) {Positive\\no constraint};
\foreach \rd/\img/\ps [count=\c from 0] in {0/positive_round000.jpg/20.6, 10/positive_round010.jpg/22.4, 30/positive_round030.jpg/22.4, 100/positive_round100.jpg/19.4, 200/positive_round200.jpg/16.5, 400/positive_round400.jpg/16.1} {
  \node[inner sep=0pt, anchor=south west] (f) at ({\stripXa+\c*(\stripCell+\stripGut)}, \rowY) {\includegraphics[width=\stripCell pt, height=\stripCell pt]{figures/images/collapse_strip/\img}};
  \node[stripchip] at ([shift={(1.6,1.6)}]f.south west) {\ps}; }
\pgfmathsetmacro\rowY{\stripH-2*\stripCell-1*\stripRowGap}
\node[striplabel, text=csna] at (\stripMargin, {\rowY+\stripCell/2}) {Signed\\no constraint};
\foreach \rd/\img/\ps [count=\c from 0] in {0/signed_unconstrained_round000.jpg/20.6, 10/signed_unconstrained_round010.jpg/21.6, 30/signed_unconstrained_round030.jpg/24.4, 100/signed_unconstrained_round100.jpg/24.2, 200/signed_unconstrained_round200.jpg/17.6, 400/signed_unconstrained_round400.jpg/16.5} {
  \node[inner sep=0pt, anchor=south west] (f) at ({\stripXa+\c*(\stripCell+\stripGut)}, \rowY) {\includegraphics[width=\stripCell pt, height=\stripCell pt]{figures/images/collapse_strip/\img}};
  \node[stripchip] at ([shift={(1.6,1.6)}]f.south west) {\ps}; }
\pgfmathsetmacro\rowY{\stripH-3*\stripCell-2*\stripRowGap}
\node[striplabel, text=csifr] at (\stripMargin, {\rowY+\stripCell/2}) {SIFR};
\foreach \rd/\img/\ps [count=\c from 0] in {0/sifr_round000.jpg/20.6, 10/sifr_round010.jpg/22.3, 30/sifr_round030.jpg/24.7, 100/sifr_round100.jpg/24.3, 200/sifr_round200.jpg/25.9, 400/sifr_round400.jpg/26.9} {
  \node[inner sep=0pt, anchor=south west] (f) at ({\stripXa+\c*(\stripCell+\stripGut)}, \rowY) {\includegraphics[width=\stripCell pt, height=\stripCell pt]{figures/images/collapse_strip/\img}};
  \node[stripchip] at ([shift={(1.6,1.6)}]f.south west) {\ps}; }
\node[striplegend] at ({\stripW/2},\stripLegendY) {
  \textcolor{cnn}{\rule[1.55pt]{10pt}{0.95pt}}\hspace{3pt}Positive\hspace{13pt}
  \textcolor{csna}{\rule[1.55pt]{10pt}{0.95pt}}\hspace{3pt}Signed\hspace{13pt}
  \textcolor{csifr}{\rule[1.4pt]{10pt}{1.2pt}}\hspace{3pt}SIFR};
\pgfmathsetmacro\chartX{\stripChartLeft}
\node[striptitle] at (\chartX,\stripTitleY) {Training-batch PickScore};
\begin{axis}[stripaxis, at={(\chartX pt,\stripChartY pt)}, ymin=15, ymax=25.5, ytick={15,20,25}, yticklabels={15,20,25}]
  \addplot[cnn, stripcurve] table[x=round, y=positive_pick] {\stripcurves};
  \addplot[csna, stripcurve] table[x=round, y=signed_unconstrained_pick] {\stripcurves};
  \addplot[csifr, stripours] table[x=round, y=sifr_pick] {\stripcurves};
\end{axis}
\pgfmathsetmacro\chartX{\stripChartLeft+\stripChartW+\stripChartGap}
\node[striptitle] at (\chartX,\stripTitleY) {Training loss};
\begin{axis}[stripaxis, at={(\chartX pt,\stripChartY pt)}, y filter/.expression={y>=0 ? log10(1+y/0.001) : -log10(1-y/0.001)}, ymin=-2.6, ymax=3.9, ytick={-2.0043,0.0000,1.0414,3.0004}, yticklabels={$-10^{-1}$,$0$,$10^{-2}$,$10^{0}$}]
  \addplot[cnn, stripcurve] table[x=round, y=positive_loss] {\stripcurves};
  \addplot[csna, stripcurve] table[x=round, y=signed_unconstrained_loss] {\stripcurves};
  \addplot[csifr, stripours] table[x=round, y=sifr_loss] {\stripcurves};
\end{axis}
\pgfmathsetmacro\chartX{\stripChartLeft+2*(\stripChartW+\stripChartGap)}
\node[striptitle] at (\chartX,\stripTitleY) {Gradient norm};
\begin{axis}[stripaxis, at={(\chartX pt,\stripChartY pt)}, ymode=log, ymin=2e-4, ymax=800, ytick={0.01,1,100}, yticklabels={$10^{-2}$,$10^{0}$,$10^{2}$}]
  \addplot[cnn, stripcurve] table[x=round, y=positive_grad] {\stripcurves};
  \addplot[csna, stripcurve] table[x=round, y=signed_unconstrained_grad] {\stripcurves};
  \addplot[csifr, stripours] table[x=round, y=sifr_grad] {\stripcurves};
\end{axis}
\node[striphead, anchor=north] at ({\stripW/2},\stripXLabelY) {Training round};
\end{tikzpicture}\endgroup%

%% file: sections/appendix_method_comparison.tex
\subsection{Derivations for the method comparison}
\label{app:implicit-proof}

We derive the coefficients in \cref{tab:connections} by matching each method's gradient to that of \cref{eq:comparison-form}.
Fix a round $k$ and use $X_t=(1-t)X_0+tX_1$ with each method's rollout law and time sampling.
Rollouts, advantages, and reference predictions are detached; $\sg{\cdot}$ denotes stop-gradient.
The effective advantage $A(X_1)$ includes each method's normalization or transformation, and, as in \cref{tab:connections}, $w(X_1)$ and the coefficients may also depend on $t$.
Within each subsection we keep the source paper's symbols; their $\beta$, $\gamma$, and $\lambda$ are unrelated to SIFR's.
It suffices to compare gradients with respect to $v_\theta(X_t,t)$, since the parameter gradients then agree by the chain rule.

The dynamic and fixed references are $\vmov(x,t)=v_{\theta_{\mathrm{ref}}^{(k)}}(x,t)$ and $\vfix(x,t)=v_{\theta_{\mathrm{ref}}^{\mathrm{fix}}}(x,t)$.
An EMA reference uses $\theta_{\mathrm{ref}}^{(k)}=\theta_{\mathrm{ema}}^{(k)}$.
For the No EMA rows, $\theta_{\mathrm{ref}}^{(k)}=\theta^{(k)}$ is a detached current snapshot, and the coefficients in the No EMA rows match gradients at $\theta=\theta^{(k)}$.

\subsubsection{ORW-CFM-W2}
\label{app:orw-mapping}

Equation~(10) of \citet{fan2025orw}, with their exponential reward weight and straight interpolation, is
\begin{equation}
\begin{aligned}
\mathcal L_{\mathrm{ORW}}^{(k)}(\theta)=\EE\Big[&e^{\tau r(X_1)}\|v_\theta(X_t,t)-(X_1-X_0)\|^2\\
&+\alpha\|v_\theta(X_t,t)-\vfix(X_t,t)\|^2\Big].
\end{aligned}
\label{eq:orw-mapping}
\end{equation}
This is already in the form of \cref{eq:comparison-form}, with
$(w(X_1),c_{\mathrm{ref}}^{(k)},c_{\mathrm{ref}}^{\mathrm{fix}})=(e^{\tau r(X_1)},0,\alpha)$.

\subsubsection{AWM}
\label{app:awm-mapping}

We use AWM's on-policy objective with an EMA reference~\citep[Eqs.~(14)--(17)]{xue2025awm}.
AWM uses data-to-noise time $s=1-t$ and the opposite velocity sign.
Reversing that sign leaves squared errors unchanged, so its time weight $w_{\mathrm{AWM}}(s)=s$ gives
\[
\ell_\theta(X_1)=\EE_{X_0,t}\big[(1-t)\|v_\theta(X_t,t)-(X_1-X_0)\|^2\big].
\]
AWM forms the surrogate $-A(X_1)\rho_\theta(X_1)$, where
\[
\rho_\theta(X_1)=\exp\big[-\ell_\theta(X_1)+\sg{\ell_\theta(X_1)}\big].
\]
At the point where the detached copy is taken, $\rho_\theta(X_1)=1$.
Differentiating only the first term in the exponent therefore gives
\[
\nabla_\theta[-A(X_1)\rho_\theta(X_1)]
=A(X_1)\rho_\theta(X_1)\nabla_\theta\ell_\theta(X_1)
=A(X_1)\nabla_\theta\ell_\theta(X_1).
\]
Adding the velocity regularizer in their Eq.~(16) gives
\begin{equation}
\begin{aligned}
\nabla_\theta\mathcal L_{\mathrm{AWM}}^{(k)}(\theta)
=\nabla_\theta\EE\Big[&(1-t)A(X_1)\|v_\theta(X_t,t)-(X_1-X_0)\|^2\\
&+\beta(1-t)\|v_\theta(X_t,t)-\vmov(X_t,t)\|^2\Big].
\end{aligned}
\label{eq:awm-mapping}
\end{equation}
Hence $w(X_1)=(1-t)A(X_1)$, $c_{\mathrm{ref}}^{(k)}=\beta(1-t)$, and $c_{\mathrm{ref}}^{\mathrm{fix}}=0$.

\subsubsection{DiffusionNFT}
\label{app:nft-mapping}

DiffusionNFT uses two branch predictions in its positive MSE objective~\citep[Eq.~(5)]{zheng2026diffusionnft}:
\begin{equation}
\begin{aligned}
v_\theta^\pm(X_t,t)&=\vmov(X_t,t)\pm\beta\big(v_\theta(X_t,t)-\vmov(X_t,t)\big),\\
\ell_{\mathrm{NFT}}&=\frac{1+A(X_1)}{2}\|v_\theta^+(X_t,t)-(X_1-X_0)\|^2\\
&\quad+\frac{1-A(X_1)}{2}\|v_\theta^-(X_t,t)-(X_1-X_0)\|^2.
\end{aligned}
\label{eq:nft-branches}
\end{equation}
Here $A(X_1)\in[-1,1]$ is the clipped and rescaled advantage, so that $(1+A(X_1))/2$ is its reward weight in $[0,1]$.
The reference prediction is detached; the two branches have derivatives $\beta I$ and $-\beta I$ with respect to $v_\theta(X_t,t)$.
Consequently,
\begin{equation}
\begin{aligned}
\nabla_{v_\theta(X_t,t)}\ell_{\mathrm{NFT}}
={}&2\beta A(X_1)\big(\vmov(X_t,t)-(X_1-X_0)\big)
+2\beta^2\big(v_\theta(X_t,t)-\vmov(X_t,t)\big)\\
={}&2\beta A(X_1)\big(v_\theta(X_t,t)-(X_1-X_0)\big)\\
&+2\beta[\beta-A(X_1)]\big(v_\theta(X_t,t)-\vmov(X_t,t)\big).
\end{aligned}
\label{eq:nft-mapping}
\end{equation}
This gives $w(X_1)=\beta A(X_1)$, $c_{\mathrm{ref}}^{(k)}=\beta[\beta-A(X_1)]$, and $c_{\mathrm{ref}}^{\mathrm{fix}}=0$.
The quadratic coefficients sum to $\beta^2$, so the Hessian with respect to the velocity is $2\beta^2I$ with the reference detached.

For the No EMA row, the reference equals the current prediction at $\theta=\theta^{(k)}$, so the second term vanishes:
\begin{equation}
\left.\nabla_{v_\theta(X_t,t)}\ell_{\mathrm{NFT}}\right|_{\theta=\theta^{(k)}}
=2\beta A(X_1)\big(v_\theta(X_t,t)-(X_1-X_0)\big).
\label{eq:nft-detach-gradient}
\end{equation}
At this point, the update gradient equals that of
\begin{equation}
\widetilde\ell_{\mathrm{NFT}}=\beta A(X_1)\|v_\theta(X_t,t)-(X_1-X_0)\|^2.
\label{eq:nft-signed}
\end{equation}

\subsubsection{RAM}
\label{app:ram-mapping}

RAM forms a detached regression target~\citep[Eq.~(17), Algorithm~1]{bergmeister2026ram}.
Its EMA variant is
\begin{equation}
\ell_{\mathrm{RAM}}
=\left\|v_\theta(X_t,t)-\sg{\vfix(X_t,t)+\eta A(X_1)\big((X_1-X_0)-\vmov(X_t,t)\big)}\right\|^2.
\label{eq:ram-detached}
\end{equation}
Differentiating with the target fixed and regrouping the terms gives
\[
\begin{aligned}
\nabla_{v_\theta(X_t,t)}\ell_{\mathrm{RAM}}
={}&2\big(v_\theta(X_t,t)-\vfix(X_t,t)\big)
+2\eta A(X_1)\big(\vmov(X_t,t)-(X_1-X_0)\big)\\
={}&2\eta A(X_1)\big(v_\theta(X_t,t)-(X_1-X_0)\big)\\
&-2\eta A(X_1)\big(v_\theta(X_t,t)-\vmov(X_t,t)\big)
+2\big(v_\theta(X_t,t)-\vfix(X_t,t)\big).
\end{aligned}
\]
This is the gradient of
\begin{equation}
\begin{aligned}
\widetilde\ell_{\mathrm{RAM}}^{\mathrm{lag}}
={}&\eta A(X_1)\|v_\theta(X_t,t)-(X_1-X_0)\|^2\\
&-\eta A(X_1)\|v_\theta(X_t,t)-\vmov(X_t,t)\|^2
+\|v_\theta(X_t,t)-\vfix(X_t,t)\|^2.
\end{aligned}
\label{eq:ram-lag-signed}
\end{equation}
Thus $w(X_1)=\eta A(X_1)$, $c_{\mathrm{ref}}^{(k)}=-\eta A(X_1)$, and $c_{\mathrm{ref}}^{\mathrm{fix}}=1$.
The quadratic coefficients sum to one, so the velocity Hessian is $2I$ with the target detached.
For the No EMA row, the dynamic-reference term has zero gradient at $\theta=\theta^{(k)}$. The update gradient then equals that of
\begin{equation}
\widetilde\ell_{\mathrm{RAM}}
=\eta A(X_1)\|v_\theta(X_t,t)-(X_1-X_0)\|^2
+\|v_\theta(X_t,t)-\vfix(X_t,t)\|^2.
\label{eq:ram-signed}
\end{equation}

\subsubsection{AdvantageFlow}
\label{app:advantageflow-mapping}

Equation~(5) of \citet{kveton2026advantageflow} is written in clean-sample prediction space.
In our time convention, the prediction is
\[
f_\psi(X_t,t)=X_t+(1-t)v_\psi(X_t,t),
\qquad\psi\in\{\theta,\theta_{\mathrm{ref}}^{(k)},\theta_{\mathrm{ref}}^{\mathrm{fix}}\}.
\]
The objective is
\begin{equation}
\begin{aligned}
\ell_{\mathrm{AF}}
={}&A(X_1)\|f_\theta(X_t,t)-X_1\|^2\\
&+\gamma(X_1)\|f_\theta(X_t,t)-f_{\theta_{\mathrm{ref}}^{(k)}}(X_t,t)\|^2\\
&+\lambda\|f_\theta(X_t,t)-f_{\theta_{\mathrm{ref}}^{\mathrm{fix}}}(X_t,t)\|^2.
\end{aligned}
\label{eq:advantageflow-prediction}
\end{equation}
The prediction errors are proportional to the corresponding velocity errors:
\[
\begin{aligned}
f_\theta(X_t,t)-X_1&=(1-t)\big(v_\theta(X_t,t)-(X_1-X_0)\big),\\
f_\theta(X_t,t)-f_\psi(X_t,t)&=(1-t)\big(v_\theta(X_t,t)-v_\psi(X_t,t)\big).
\end{aligned}
\]
Substituting these identities into \cref{eq:advantageflow-prediction} gives
\begin{equation}
\begin{aligned}
w(X_1)&=(1-t)^2A(X_1),\\
c_{\mathrm{ref}}^{(k)}&=(1-t)^2\gamma(X_1),\qquad
c_{\mathrm{ref}}^{\mathrm{fix}}=(1-t)^2\lambda.
\end{aligned}
\label{eq:advantageflow-mapping}
\end{equation}
AdvantageFlow clips advantages to $[-1,1]$, uses $\gamma(X_1)=1.1$ or $1-A(X_1)$, and keeps $\lambda$ fixed (Secs.~3.2--3.3).
Its dynamic reference is an EMA model.

\subsubsection{FlowAWR}
\label{app:flowawr-mapping}

FlowAWR forms a detached regression target from an EMA model~\citep[Eqs.~(10)--(11)]{fu2026flowawr}:
\begin{equation}
\ell_{\mathrm{AWR}}=
\left\|v_\theta(X_t,t)-\sg{\vmov(X_t,t)
+A(X_1)\big((X_1-X_0)-\vmov(X_t,t)\big)}\right\|^2.
\label{eq:flowawr-target}
\end{equation}
Expanding the square separates the signed fitting and reference terms:
\begin{equation}
\begin{aligned}
\ell_{\mathrm{AWR}}
={}&A(X_1)\|v_\theta(X_t,t)-(X_1-X_0)\|^2\\
&+[1-A(X_1)]\|v_\theta(X_t,t)-\vmov(X_t,t)\|^2\\
&-A(X_1)[1-A(X_1)]\|(X_1-X_0)-\vmov(X_t,t)\|^2.
\end{aligned}
\label{eq:flowawr-mapping}
\end{equation}
The last term is independent of $\theta$, yielding $w(X_1)=A(X_1)$, $c_{\mathrm{ref}}^{(k)}=1-A(X_1)$, and $c_{\mathrm{ref}}^{\mathrm{fix}}=0$.
For a group of $G$ rollouts $X_{1,1},\ldots,X_{1,G}$, FlowAWR uses
\[
A(X_{1,i})=
\frac{\exp(r(X_{1,i})/\gamma_{\mathrm{AWR}})}{G^{-1}\sum_{j=1}^G\exp(r(X_{1,j})/\gamma_{\mathrm{AWR}})}-1,
\]
with $\gamma_{\mathrm{AWR}}$ set from the reward standard deviation in the current experience buffer.
Although $1-A(X_1)$ can be negative, the quadratic coefficients sum to one, giving the same positive velocity Hessian as the original regression loss.

\subsubsection{Current-snapshot gradients and lagged references}
\label{app:reference-lag}

Once the model moves away from the snapshot $\theta^{(k)}$ at which the No EMA rows hold, the dynamic-reference terms in \cref{eq:nft-mapping,eq:ram-lag-signed} become active.

DiffusionNFT's soft-update ablation reports that hard-copying the model after each round gives fast initial improvement but later destabilizes training, while an increasing EMA decay improves stability~\citep[Secs.~3.3 and~4.4]{zheng2026diffusionnft}.
The coefficient $\beta[\beta-A(X_1)]$ in \cref{eq:nft-mapping} is positive for $A(X_1)<\beta$, in particular for every below-average sample, giving a restoring gradient toward the reference.
A lagged reference keeps this gradient active as the model changes, while EMA smooths changes in the reference across rounds.
This may explain why DiffusionNFT relies on a high EMA decay: the strength of its restoring gradient is set by how far the model has moved from the reference.
SIFR uses a fixed decay of $0.5$ throughout (\cref{sec:controller}), since the multiplier sets the regularization strength instead.

%% file: sections/appendix_results.tex
\clearpage
\suppressfloats[t]
\section{Additional experimental results}
\label{app:generalization}

We report complete reward evaluations on held-out Pick-a-Pic prompts, UniGenBench with physical prompts, and DrawBench.
Each table compares single-reward training (top) and multi-reward training (bottom) across all three backbones.

\subsection{Held-out Pick-a-Pic prompts}
When trained on PickScore alone, SIFR achieves the highest PickScore and Aesthetic Score on SD3.5 Medium, FLUX.2 Klein, and Cosmos 3 Super Text2Image (\cref{tab:single-reward}).
\input{tables/single_reward}

\clearpage
\subsection{Transferability across held-out prompt sets}
\Cref{tab:crossset-unigenbench,tab:crossset-drawbench} evaluate the same checkpoints on UniGenBench with physical prompts and on DrawBench.
The transfer gains in \cref{tab:cosmos-generalization} extend to the smaller backbones and to single-reward training: SIFR has the highest PickScore and Aesthetic Score in all twelve combinations of backbone, training reward, and prompt set.

The pretrained evaluations use CFG $2$ with $40$ Euler steps for SD3.5 Medium, CFG $4$ with $28$ steps for FLUX.2 Klein, and CFG $6$ with $35$ steps for Cosmos 3 Super Text2Image.
\begingroup
\input{tables/prompt_set_unigenbench}
\clearpage
\input{tables/prompt_set_drawbench}
\FloatBarrier
\endgroup

\subsection{Single-reward gains across prompts}
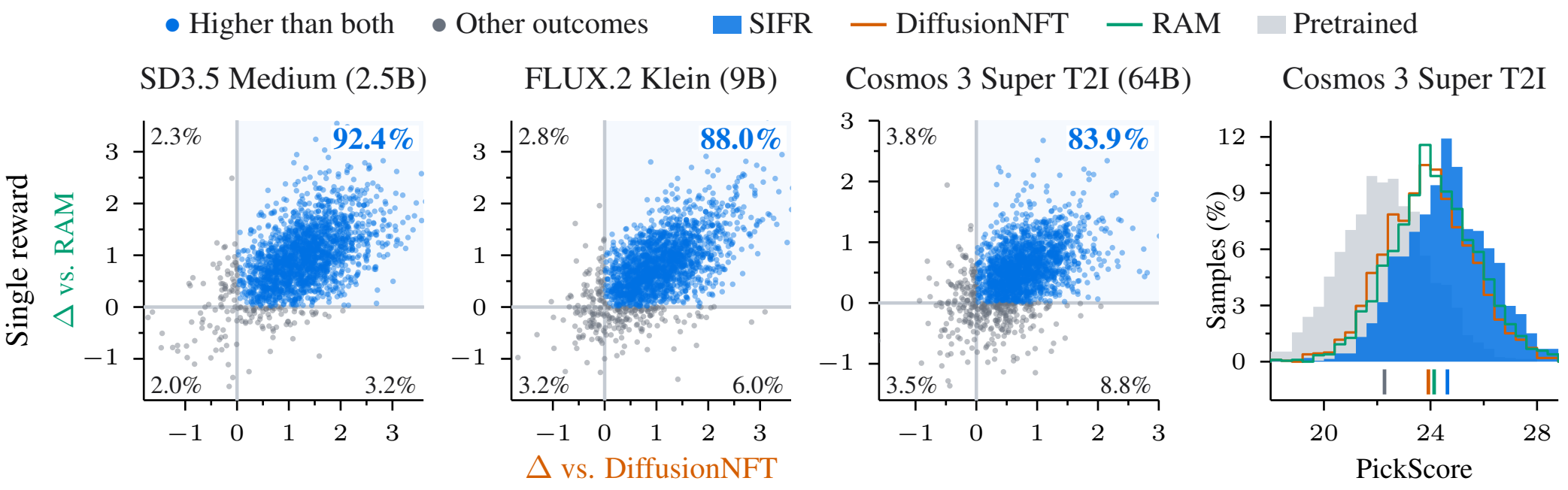
\begin{figure}[!htb]
\centering
\input{figures/reward_gain_single.tex}
\caption{\textbf{Per-prompt PickScore gains after single-reward training}, constructed as in \cref{fig:reward-gain-distributions}. Right: the PickScore distribution on Cosmos 3 Super T2I.}
\label{fig:reward-gain-single}
\end{figure}

After single-reward training, SIFR scores above both baselines in $83.9\%$--$92.4\%$ of the $2048$ held-out Pick-a-Pic prompts and noise seeds (\cref{fig:reward-gain-single}).
\FloatBarrier

\input{sections/appendix_analyses}

%% file: tables/single_reward.tex
\begin{table}[H]
\centering
\begingroup\small
\setlength{\tabcolsep}{2.1pt}
\renewcommand{\arraystretch}{1.06}
\newsavebox{\singlepreferencebox}
\sbox{\singlepreferencebox}{%
\begin{tabular}{@{}l@{\hspace{7pt}}C{38pt}!{\hspace{4pt}}*{2}{C{38pt}}!{\hspace{4pt}}*{4}{C{38pt}}@{}}
\toprule[0.7pt]
\multicolumn{1}{@{}c@{\hspace{7pt}}}{\multirow{2}{*}[-2.5pt]{\footnotesize Method}} & \multicolumn{1}{c}{\makebox[0pt]{\footnotesize Training reward}} & \multicolumn{6}{c}{\footnotesize Held-out rewards} \\
\cmidrule(l{-10pt}r{-10pt}){2-2}\cmidrule(l{20pt}){3-8}
& \footnotesize PickScore & \footnotesize HPSv2.1 & \footnotesize ClipScore & \footnotesize ImgRwd & \footnotesize HPSv3 & \footnotesize UniRwd2 & \footnotesize Aesthetic \\
\specialrule{0.35pt}{2.25pt}{2.75pt}
\textit{SD3.5 Medium}\,{\scriptsize\textcolor{black!55}{(2.5B)}} & 21.78 & 0.295 & 0.289 & 1.02 & 9.18 & 3.194 & 5.96 \\
\hspace{0.6em}+\,DiffusionNFT & 23.26 & 0.325 & \textbf{0.291} & \underline{1.38} & 9.99 & \underline{3.236} & 6.28 \\
\hspace{0.6em}+\,RAM & \underline{23.56} & \underline{0.326} & 0.288 & \textbf{1.39} & \underline{10.57} & 3.194 & \underline{6.42} \\
\hspace{0.6em}+\,SIFR & \textbf{24.47} & \textbf{0.336} & \underline{0.290} & \underline{1.38} & \textbf{11.01} & \textbf{3.243} & \textbf{6.66} \\
\arrayrulecolor{black!22}\specialrule{0.25pt}{2.25pt}{2.75pt}\arrayrulecolor{black}
\textit{FLUX.2 Klein}\,{\scriptsize\textcolor{black!55}{(9B)}} & 21.82 & 0.299 & 0.290 & 1.18 & 9.47 & 3.249 & 5.86 \\
\hspace{0.6em}+\,DiffusionNFT & 23.68 & \underline{0.336} & \textbf{0.298} & \textbf{1.47} & \underline{11.54} & \textbf{3.271} & \underline{6.40} \\
\hspace{0.6em}+\,RAM & \underline{23.95} & 0.327 & \underline{0.297} & 1.32 & 10.85 & 3.243 & 6.36 \\
\hspace{0.6em}+\,SIFR & \textbf{24.69} & \textbf{0.340} & \textbf{0.298} & \underline{1.42} & \textbf{11.68} & \underline{3.256} & \textbf{6.54} \\
\arrayrulecolor{black!22}\specialrule{0.25pt}{2.25pt}{2.75pt}\arrayrulecolor{black}
\textit{Cosmos 3 Super T2I}\,{\scriptsize\textcolor{black!55}{(64B)}} & 22.27 & 0.312 & \textbf{0.301} & 1.29 & 9.12 & 3.284 & 5.95 \\
\hspace{0.6em}+\,DiffusionNFT & 23.92 & \underline{0.335} & 0.298 & 1.44 & 10.76 & 3.303 & 6.44 \\
\hspace{0.6em}+\,RAM & \underline{24.14} & \textbf{0.345} & 0.294 & \textbf{1.52} & \textbf{11.71} & \textbf{3.312} & \underline{6.59} \\
\hspace{0.6em}+\,SIFR & \textbf{24.64} & \textbf{0.345} & \underline{0.299} & \underline{1.47} & \underline{11.51} & \underline{3.308} & \textbf{6.66} \\
\specialrule{0.55pt}{2.25pt}{4pt}
\multicolumn{1}{@{}c@{\hspace{7pt}}}{\multirow{2}{*}[-2.5pt]{\footnotesize Method}} & \multicolumn{3}{c}{\makebox[0pt]{\footnotesize Training rewards}} & \multicolumn{4}{c}{\footnotesize Held-out rewards} \\
\cmidrule(r){2-4}\cmidrule(l){5-8}
& \footnotesize PickScore & \footnotesize HPSv2.1 & \footnotesize ClipScore & \footnotesize ImgRwd & \footnotesize HPSv3 & \footnotesize UniRwd2 & \footnotesize Aesthetic \\
\specialrule{0.35pt}{2.25pt}{2.75pt}
\textit{SD3.5 Medium}\,{\scriptsize\textcolor{black!55}{(2.5B)}} & 21.78 & 0.295 & 0.289 & 1.02 & 9.18 & 3.194 & 5.96 \\
\hspace{0.6em}+\,DiffusionNFT & 23.06 & 0.348 & 0.318 & 1.56 & 10.52 & 3.249 & \underline{6.12} \\
\hspace{0.6em}+\,RAM & \underline{23.37} & \underline{0.361} & \underline{0.328} & \underline{1.61} & \underline{11.92} & \textbf{3.294} & 6.08 \\
\hspace{0.6em}+\,SIFR & \textbf{23.76} & \textbf{0.374} & \textbf{0.339} & \textbf{1.67} & \textbf{12.12} & \underline{3.290} & \textbf{6.15} \\
\arrayrulecolor{black!22}\specialrule{0.25pt}{2.25pt}{2.75pt}\arrayrulecolor{black}
\textit{FLUX.2 Klein}\,{\scriptsize\textcolor{black!55}{(9B)}} & 21.82 & 0.299 & 0.290 & 1.18 & 9.47 & 3.249 & 5.86 \\
\hspace{0.6em}+\,DiffusionNFT & \underline{23.69} & \underline{0.364} & \underline{0.328} & \underline{1.63} & \underline{12.20} & \underline{3.290} & \underline{6.21} \\
\hspace{0.6em}+\,RAM\smash{\textsuperscript{\ensuremath{\dagger}}} & 23.26 & 0.351 & 0.315 & 1.57 & 11.64 & \textbf{3.304} & \textbf{6.23} \\
\hspace{0.6em}+\,SIFR & \textbf{24.00} & \textbf{0.385} & \textbf{0.336} & \textbf{1.67} & \textbf{12.86} & \underline{3.290} & \textbf{6.23} \\
\arrayrulecolor{black!22}\specialrule{0.25pt}{2.25pt}{2.75pt}\arrayrulecolor{black}
\textit{Cosmos 3 Super T2I}\,{\scriptsize\textcolor{black!55}{(64B)}} & 22.27 & 0.312 & 0.301 & 1.29 & 9.12 & 3.284 & 5.95 \\
\hspace{0.6em}+\,DiffusionNFT & 23.46 & 0.351 & 0.307 & 1.60 & 10.76 & 3.319 & 6.42 \\
\hspace{0.6em}+\,RAM & \underline{23.70} & \underline{0.365} & \underline{0.311} & \underline{1.62} & \underline{11.40} & \underline{3.329} & \underline{6.46} \\
\hspace{0.6em}+\,SIFR & \textbf{24.07} & \textbf{0.378} & \textbf{0.317} & \textbf{1.67} & \textbf{11.95} & \textbf{3.355} & \textbf{6.62} \\
\specialrule{0.7pt}{0.75pt}{0pt}
\end{tabular}}
\usebox{\singlepreferencebox}
\normalsize
\caption{\textbf{Single- and multi-reward training.} Reward scores on $2048$ held-out Pick-a-Pic prompts after $500$ rounds; the upper and lower blocks report single- and multi-reward training, respectively. {\footnotesize\textsuperscript{$\dagger$}\,round $300$, before RAM on FLUX.2 Klein collapses.}}
\label{tab:single-reward}
\endgroup
\end{table}

%% file: tables/prompt_set_unigenbench.tex
\begin{table}[H]
\centering
\begingroup\small
\setlength{\tabcolsep}{2.1pt}
\renewcommand{\arraystretch}{1.06}
\newsavebox{\crosssetunigenbox}
\sbox{\crosssetunigenbox}{%
\begin{tabular}{@{}l@{\hspace{7pt}}C{38pt}!{\hspace{4pt}}*{2}{C{38pt}}!{\hspace{4pt}}*{4}{C{38pt}}@{}}
\toprule[0.7pt]
\multicolumn{1}{@{}c@{\hspace{7pt}}}{\multirow{2}{*}[-2.5pt]{\footnotesize Method}} & \multicolumn{1}{c}{\makebox[0pt]{\footnotesize Training reward}} & \multicolumn{6}{c}{\footnotesize Held-out rewards} \\
\cmidrule(l{-10pt}r{-10pt}){2-2}\cmidrule(l{20pt}){3-8}
& \footnotesize PickScore & \footnotesize HPSv2.1 & \footnotesize ClipScore & \footnotesize ImgRwd & \footnotesize HPSv3 & \footnotesize UniRwd2 & \footnotesize Aesthetic \\
\specialrule{0.35pt}{2.25pt}{2.75pt}
\textit{SD3.5 Medium}\,{\scriptsize\textcolor{black!55}{(2.5B)}} & 21.87 & 0.258 & \textbf{0.287} & 0.43 & 10.86 & 3.188 & 5.75 \\
\hspace{0.6em}+\,DiffusionNFT & 23.07 & 0.304 & \underline{0.278} & \underline{1.01} & 11.65 & \textbf{3.280} & 6.03 \\
\hspace{0.6em}+\,RAM & \underline{23.64} & \underline{0.309} & 0.274 & \underline{1.01} & \textbf{12.94} & 3.250 & \underline{6.21} \\
\hspace{0.6em}+\,SIFR & \textbf{24.39} & \textbf{0.323} & 0.269 & \textbf{1.04} & \underline{12.90} & \underline{3.277} & \textbf{6.38} \\
\arrayrulecolor{black!22}\specialrule{0.25pt}{2.25pt}{2.75pt}\arrayrulecolor{black}
\textit{FLUX.2 Klein}\,{\scriptsize\textcolor{black!55}{(9B)}} & 22.12 & 0.279 & \textbf{0.295} & 0.73 & 11.90 & 3.320 & 5.75 \\
\hspace{0.6em}+\,DiffusionNFT & 23.28 & \underline{0.307} & \underline{0.287} & \textbf{1.03} & \underline{13.72} & \underline{3.347} & \underline{6.05} \\
\hspace{0.6em}+\,RAM & \underline{23.55} & 0.304 & 0.286 & 0.98 & 13.31 & \textbf{3.352} & 6.03 \\
\hspace{0.6em}+\,SIFR & \textbf{24.76} & \textbf{0.316} & 0.283 & \underline{0.99} & \textbf{13.92} & 3.312 & \textbf{6.23} \\
\arrayrulecolor{black!22}\specialrule{0.25pt}{2.25pt}{2.75pt}\arrayrulecolor{black}
\textit{Cosmos 3 Super T2I}\,{\scriptsize\textcolor{black!55}{(64B)}} & 22.44 & 0.288 & \textbf{0.297} & 0.86 & 11.17 & 3.364 & 5.80 \\
\hspace{0.6em}+\,DiffusionNFT & 23.37 & 0.300 & \underline{0.290} & 0.99 & 11.99 & 3.378 & 5.95 \\
\hspace{0.6em}+\,RAM & \underline{23.85} & \underline{0.316} & 0.286 & \underline{1.09} & \underline{12.76} & \underline{3.392} & \underline{6.09} \\
\hspace{0.6em}+\,SIFR & \textbf{24.32} & \textbf{0.320} & 0.287 & \textbf{1.11} & \textbf{12.96} & \textbf{3.399} & \textbf{6.18} \\
\specialrule{0.55pt}{2.25pt}{4pt}
\multicolumn{1}{@{}c@{\hspace{7pt}}}{\multirow{2}{*}[-2.5pt]{\footnotesize Method}} & \multicolumn{3}{c}{\makebox[0pt]{\footnotesize Training rewards}} & \multicolumn{4}{c}{\footnotesize Held-out rewards} \\
\cmidrule(r){2-4}\cmidrule(l){5-8}
& \footnotesize PickScore & \footnotesize HPSv2.1 & \footnotesize ClipScore & \footnotesize ImgRwd & \footnotesize HPSv3 & \footnotesize UniRwd2 & \footnotesize Aesthetic \\
\specialrule{0.35pt}{2.25pt}{2.75pt}
\textit{SD3.5 Medium}\,{\scriptsize\textcolor{black!55}{(2.5B)}} & 21.87 & 0.258 & 0.287 & 0.43 & 10.86 & 3.188 & 5.75 \\
\hspace{0.6em}+\,DiffusionNFT & 22.76 & 0.324 & \underline{0.296} & 1.09 & 12.07 & 3.281 & 5.99 \\
\hspace{0.6em}+\,RAM & \underline{23.45} & \underline{0.342} & 0.294 & \underline{1.18} & \underline{13.70} & \underline{3.312} & \underline{6.13} \\
\hspace{0.6em}+\,SIFR & \textbf{23.95} & \textbf{0.357} & \textbf{0.303} & \textbf{1.27} & \textbf{13.77} & \textbf{3.314} & \textbf{6.23} \\
\arrayrulecolor{black!22}\specialrule{0.25pt}{2.25pt}{2.75pt}\arrayrulecolor{black}
\textit{FLUX.2 Klein}\,{\scriptsize\textcolor{black!55}{(9B)}} & 22.12 & 0.279 & 0.295 & 0.73 & 11.90 & 3.320 & 5.75 \\
\hspace{0.6em}+\,DiffusionNFT & \underline{23.41} & \underline{0.334} & \underline{0.301} & \underline{1.10} & \underline{13.84} & \underline{3.356} & \underline{6.16} \\
\hspace{0.6em}+\,RAM\smash{\textsuperscript{\ensuremath{\dagger}}} & 23.14 & 0.315 & 0.296 & 1.01 & 13.05 & \textbf{3.362} & 6.00 \\
\hspace{0.6em}+\,SIFR & \textbf{24.12} & \textbf{0.359} & \textbf{0.311} & \textbf{1.13} & \textbf{15.19} & 3.343 & \textbf{6.32} \\
\arrayrulecolor{black!22}\specialrule{0.25pt}{2.25pt}{2.75pt}\arrayrulecolor{black}
\textit{Cosmos 3 Super T2I}\,{\scriptsize\textcolor{black!55}{(64B)}} & 22.44 & 0.288 & \textbf{0.297} & 0.86 & 11.17 & 3.364 & 5.80 \\
\hspace{0.6em}+\,DiffusionNFT & 23.12 & 0.313 & 0.294 & 1.08 & 12.03 & 3.376 & 6.01 \\
\hspace{0.6em}+\,RAM & \underline{23.50} & \underline{0.330} & \underline{0.296} & \underline{1.14} & \underline{12.66} & \underline{3.401} & \underline{6.08} \\
\hspace{0.6em}+\,SIFR & \textbf{23.93} & \textbf{0.346} & \textbf{0.297} & \textbf{1.18} & \textbf{13.11} & \textbf{3.416} & \textbf{6.30} \\
\specialrule{0.7pt}{0.75pt}{0pt}
\end{tabular}}
\usebox{\crosssetunigenbox}
\normalsize
\caption{\textbf{Transfer to UniGenBench.} Reward scores on UniGenBench with physical prompts ($1170$ prompts) after $500$ rounds of single-reward (top) and multi-reward (bottom) training on Pick-a-Pic. {\footnotesize\textsuperscript{$\dagger$}\,round $300$, before RAM on FLUX.2 Klein collapses.}}
\label{tab:crossset-unigenbench}
\endgroup
\end{table}

%% file: tables/prompt_set_drawbench.tex
\begin{table}[H]
\centering
\begingroup\small
\setlength{\tabcolsep}{2.1pt}
\renewcommand{\arraystretch}{1.06}
\newsavebox{\crosssetdrawbox}
\sbox{\crosssetdrawbox}{%
\begin{tabular}{@{}l@{\hspace{7pt}}C{38pt}!{\hspace{4pt}}*{2}{C{38pt}}!{\hspace{4pt}}*{4}{C{38pt}}@{}}
\toprule[0.7pt]
\multicolumn{1}{@{}c@{\hspace{7pt}}}{\multirow{2}{*}[-2.5pt]{\footnotesize Method}} & \multicolumn{1}{c}{\makebox[0pt]{\footnotesize Training reward}} & \multicolumn{6}{c}{\footnotesize Held-out rewards} \\
\cmidrule(l{-10pt}r{-10pt}){2-2}\cmidrule(l{20pt}){3-8}
& \footnotesize PickScore & \footnotesize HPSv2.1 & \footnotesize ClipScore & \footnotesize ImgRwd & \footnotesize HPSv3 & \footnotesize UniRwd2 & \footnotesize Aesthetic \\
\specialrule{0.35pt}{2.25pt}{2.75pt}
\textit{SD3.5 Medium}\,{\scriptsize\textcolor{black!55}{(2.5B)}} & 21.97 & 0.255 & 0.274 & 0.51 & 6.95 & 3.040 & 5.33 \\
\hspace{0.6em}+\,DiffusionNFT & 23.70 & \underline{0.315} & \textbf{0.282} & 1.38 & 12.03 & \textbf{3.228} & 5.89 \\
\hspace{0.6em}+\,RAM & \underline{23.82} & 0.314 & \underline{0.278} & \underline{1.39} & \underline{12.65} & 3.167 & \underline{6.20} \\
\hspace{0.6em}+\,SIFR & \textbf{24.60} & \textbf{0.321} & \underline{0.278} & \textbf{1.44} & \textbf{12.91} & \underline{3.224} & \textbf{6.33} \\
\arrayrulecolor{black!22}\specialrule{0.25pt}{2.25pt}{2.75pt}\arrayrulecolor{black}
\textit{FLUX.2 Klein}\,{\scriptsize\textcolor{black!55}{(9B)}} & 22.60 & 0.278 & \textbf{0.296} & 0.99 & 10.42 & 3.219 & 5.20 \\
\hspace{0.6em}+\,DiffusionNFT & 24.18 & \textbf{0.324} & \underline{0.289} & \textbf{1.46} & \textbf{13.99} & \textbf{3.302} & 5.91 \\
\hspace{0.6em}+\,RAM & \underline{24.20} & \underline{0.317} & 0.284 & 1.41 & \underline{13.52} & \underline{3.301} & \underline{5.95} \\
\hspace{0.6em}+\,SIFR & \textbf{24.88} & 0.314 & 0.284 & \underline{1.45} & 13.39 & 3.246 & \textbf{6.27} \\
\arrayrulecolor{black!22}\specialrule{0.25pt}{2.25pt}{2.75pt}\arrayrulecolor{black}
\textit{Cosmos 3 Super T2I}\,{\scriptsize\textcolor{black!55}{(64B)}} & 22.96 & 0.297 & \textbf{0.292} & 1.15 & 10.30 & 3.292 & 5.36 \\
\hspace{0.6em}+\,DiffusionNFT & 24.17 & 0.322 & \underline{0.283} & 1.39 & 12.43 & \underline{3.353} & 5.99 \\
\hspace{0.6em}+\,RAM & \underline{24.43} & \underline{0.329} & 0.278 & \underline{1.45} & \underline{13.25} & 3.343 & \underline{6.32} \\
\hspace{0.6em}+\,SIFR & \textbf{24.83} & \textbf{0.333} & 0.281 & \textbf{1.48} & \textbf{13.29} & \textbf{3.357} & \textbf{6.40} \\
\specialrule{0.55pt}{2.25pt}{4pt}
\multicolumn{1}{@{}c@{\hspace{7pt}}}{\multirow{2}{*}[-2.5pt]{\footnotesize Method}} & \multicolumn{3}{c}{\makebox[0pt]{\footnotesize Training rewards}} & \multicolumn{4}{c}{\footnotesize Held-out rewards} \\
\cmidrule(r){2-4}\cmidrule(l){5-8}
& \footnotesize PickScore & \footnotesize HPSv2.1 & \footnotesize ClipScore & \footnotesize ImgRwd & \footnotesize HPSv3 & \footnotesize UniRwd2 & \footnotesize Aesthetic \\
\specialrule{0.35pt}{2.25pt}{2.75pt}
\textit{SD3.5 Medium}\,{\scriptsize\textcolor{black!55}{(2.5B)}} & 21.97 & 0.255 & 0.274 & 0.51 & 6.95 & 3.040 & 5.33 \\
\hspace{0.6em}+\,DiffusionNFT & 23.49 & 0.334 & \underline{0.305} & 1.50 & 12.65 & 3.233 & 5.68 \\
\hspace{0.6em}+\,RAM & \underline{23.74} & \underline{0.343} & \underline{0.305} & \underline{1.51} & \textbf{14.29} & \textbf{3.270} & \underline{5.69} \\
\hspace{0.6em}+\,SIFR & \textbf{24.14} & \textbf{0.355} & \textbf{0.313} & \textbf{1.56} & \underline{14.24} & \underline{3.268} & \textbf{5.90} \\
\arrayrulecolor{black!22}\specialrule{0.25pt}{2.25pt}{2.75pt}\arrayrulecolor{black}
\textit{FLUX.2 Klein}\,{\scriptsize\textcolor{black!55}{(9B)}} & 22.60 & 0.278 & 0.296 & 0.99 & 10.42 & 3.219 & 5.20 \\
\hspace{0.6em}+\,DiffusionNFT & \underline{24.16} & \underline{0.352} & \underline{0.307} & \textbf{1.57} & \underline{14.74} & \textbf{3.321} & \underline{5.89} \\
\hspace{0.6em}+\,RAM\smash{\textsuperscript{\ensuremath{\dagger}}} & 23.80 & 0.338 & 0.304 & 1.47 & 14.43 & \underline{3.318} & 5.76 \\
\hspace{0.6em}+\,SIFR & \textbf{24.39} & \textbf{0.366} & \textbf{0.318} & \underline{1.54} & \textbf{15.28} & 3.305 & \textbf{5.92} \\
\arrayrulecolor{black!22}\specialrule{0.25pt}{2.25pt}{2.75pt}\arrayrulecolor{black}
\textit{Cosmos 3 Super T2I}\,{\scriptsize\textcolor{black!55}{(64B)}} & 22.96 & 0.297 & 0.292 & 1.15 & 10.30 & 3.292 & 5.36 \\
\hspace{0.6em}+\,DiffusionNFT & 23.82 & 0.331 & 0.294 & 1.42 & 12.46 & \underline{3.339} & 5.87 \\
\hspace{0.6em}+\,RAM & \underline{24.03} & \underline{0.347} & \underline{0.297} & \underline{1.50} & \underline{13.17} & 3.330 & \underline{5.92} \\
\hspace{0.6em}+\,SIFR & \textbf{24.36} & \textbf{0.359} & \textbf{0.299} & \textbf{1.55} & \textbf{13.64} & \textbf{3.356} & \textbf{6.24} \\
\specialrule{0.7pt}{0.75pt}{0pt}
\end{tabular}}
\usebox{\crosssetdrawbox}
\normalsize
\caption{\textbf{Transfer to DrawBench.} Reward scores on DrawBench ($200$ prompts, $5$ images each) after $500$ rounds of single-reward (top) and multi-reward (bottom) training on Pick-a-Pic. {\footnotesize\textsuperscript{$\dagger$}\,round $300$, before RAM on FLUX.2 Klein collapses.}}
\label{tab:crossset-drawbench}
\endgroup
\end{table}

%% file: figures/reward_gain_single.tex
\begingroup%
\input{style/plots.tex}%
\begin{tikzpicture}[x=1pt,y=1pt]%
\path[use as bounding box] (0,0) rectangle (396,127);%
\begin{pgfinterruptboundingbox}%
\node[anchor=base,inner sep=0pt,text=sifrInk,font=\fontfamily{ptm}\fontsize{7.6}{8.8}\selectfont] at (198,116) {\textcolor{sifrBlue}{$\bullet$}\hspace{2pt}Higher than both\hspace{9pt}\textcolor{sifrMuted}{$\bullet$}\hspace{2pt}Other outcomes\hspace{16pt}\tikz[baseline=-.6ex]{\fill[sifrBlue,fill opacity=.85] (0,-2.2pt) rectangle (7pt,2.2pt);}\hspace{2pt}SIFR\hspace{9pt}\tikz[baseline=-.6ex]{\draw[sifrNFT,line width=.8pt] (0,0) -- (9pt,0);}\hspace{2pt}DiffusionNFT\hspace{9pt}\tikz[baseline=-.6ex]{\draw[sifrRAM,line width=.8pt] (0,0) -- (9pt,0);}\hspace{2pt}RAM\hspace{9pt}\tikz[baseline=-.6ex]{\fill[sifrBase!75] (0,-2.2pt) rectangle (7pt,2.2pt);}\hspace{2pt}Pretrained};%
\node[anchor=base,inner sep=0pt,text=sifrInk,font=\fontfamily{ptm}\fontsize{8.0}{9.2}\selectfont] at (71.0,102) {SD3.5 Medium (2.5B)};%
\begin{axis}[sifr axis,at={(36.0pt,24pt)},width=70pt,height=70pt,xmin=-1.8,xmax=3.6,ymin=-1.8,ymax=3.6,xtick={-1,0,1,2,3},ytick={-1,0,1,2,3},ymajorgrids=false,axis line style={black,line width=.5pt},y axis line style={draw=black,line width=.5pt},tick style={black,line width=.5pt},ytick style={draw=black,line width=.5pt},major tick length=2.6pt,minor tick length=1.3pt,minor tick num=1,tick label style={font=\fontfamily{ptm}\fontsize{6.2}{7.4}\selectfont,text=black},xticklabel style={/pgf/number format/fixed,/pgf/number format/precision=1},yticklabel style={/pgf/number format/fixed,/pgf/number format/precision=1},clip=true,enlargelimits=false]%
\path[fill=sifrBlue!4] (axis cs:0,0) rectangle (axis cs:3.6,3.6);%
\addplot[sifrBase,line width=.85pt,mark=none] coordinates {(-1.8,0) (3.6,0)};%
\addplot[sifrBase,line width=.85pt,mark=none] coordinates {(0,-1.8) (0,3.6)};%
\node[anchor=south west,inner sep=0pt] at (rel axis cs:0,0) {\includegraphics[width=70pt,height=70pt]{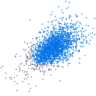}};%
\end{axis}%
\node[anchor=north east,inner sep=0pt,text=sifrBlue,font=\fontfamily{ptm}\fontsize{7.6}{8.8}\selectfont,fill=white,inner sep=.5pt,font=\fontfamily{ptm}\fontsize{7.6}{8.6}\selectfont\bfseries] at (104.95,92.95) {92.4\%};%
\node[anchor=north west,inner sep=0pt,text=sifrInk,font=\fontfamily{ptm}\fontsize{6.2}{7.4}\selectfont,fill=white,inner sep=.5pt] at (37.05,92.95) {2.3\%};%
\node[anchor=south east,inner sep=0pt,text=sifrInk,font=\fontfamily{ptm}\fontsize{6.2}{7.4}\selectfont,fill=white,inner sep=.5pt] at (104.95,25.05) {3.2\%};%
\node[anchor=south west,inner sep=0pt,text=sifrInk,font=\fontfamily{ptm}\fontsize{6.2}{7.4}\selectfont,fill=white,inner sep=.5pt] at (37.05,25.05) {2.0\%};%
\node[anchor=base,inner sep=0pt,text=sifrInk,font=\fontfamily{ptm}\fontsize{8.0}{9.2}\selectfont] at (163.0,102) {FLUX.2 Klein (9B)};%
\begin{axis}[sifr axis,at={(128.0pt,24pt)},width=70pt,height=70pt,xmin=-1.8,xmax=3.6,ymin=-1.8,ymax=3.6,xtick={-1,0,1,2,3},ytick={-1,0,1,2,3},ymajorgrids=false,axis line style={black,line width=.5pt},y axis line style={draw=black,line width=.5pt},tick style={black,line width=.5pt},ytick style={draw=black,line width=.5pt},major tick length=2.6pt,minor tick length=1.3pt,minor tick num=1,tick label style={font=\fontfamily{ptm}\fontsize{6.2}{7.4}\selectfont,text=black},xticklabel style={/pgf/number format/fixed,/pgf/number format/precision=1},yticklabel style={/pgf/number format/fixed,/pgf/number format/precision=1},clip=true,enlargelimits=false]%
\path[fill=sifrBlue!4] (axis cs:0,0) rectangle (axis cs:3.6,3.6);%
\addplot[sifrBase,line width=.85pt,mark=none] coordinates {(-1.8,0) (3.6,0)};%
\addplot[sifrBase,line width=.85pt,mark=none] coordinates {(0,-1.8) (0,3.6)};%
\node[anchor=south west,inner sep=0pt] at (rel axis cs:0,0) {\includegraphics[width=70pt,height=70pt]{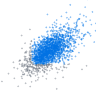}};%
\end{axis}%
\node[anchor=north east,inner sep=0pt,text=sifrBlue,font=\fontfamily{ptm}\fontsize{7.6}{8.8}\selectfont,fill=white,inner sep=.5pt,font=\fontfamily{ptm}\fontsize{7.6}{8.6}\selectfont\bfseries] at (196.95,92.95) {88.0\%};%
\node[anchor=north west,inner sep=0pt,text=sifrInk,font=\fontfamily{ptm}\fontsize{6.2}{7.4}\selectfont,fill=white,inner sep=.5pt] at (129.05,92.95) {2.8\%};%
\node[anchor=south east,inner sep=0pt,text=sifrInk,font=\fontfamily{ptm}\fontsize{6.2}{7.4}\selectfont,fill=white,inner sep=.5pt] at (196.95,25.05) {6.0\%};%
\node[anchor=south west,inner sep=0pt,text=sifrInk,font=\fontfamily{ptm}\fontsize{6.2}{7.4}\selectfont,fill=white,inner sep=.5pt] at (129.05,25.05) {3.2\%};%
\node[anchor=base,inner sep=0pt,text=sifrInk,font=\fontfamily{ptm}\fontsize{8.0}{9.2}\selectfont] at (255.0,102) {Cosmos 3 Super T2I (64B)};%
\begin{axis}[sifr axis,at={(220.0pt,24pt)},width=70pt,height=70pt,xmin=-1.6,xmax=3,ymin=-1.6,ymax=3,xtick={-1,0,1,2,3},ytick={-1,0,1,2,3},ymajorgrids=false,axis line style={black,line width=.5pt},y axis line style={draw=black,line width=.5pt},tick style={black,line width=.5pt},ytick style={draw=black,line width=.5pt},major tick length=2.6pt,minor tick length=1.3pt,minor tick num=1,tick label style={font=\fontfamily{ptm}\fontsize{6.2}{7.4}\selectfont,text=black},xticklabel style={/pgf/number format/fixed,/pgf/number format/precision=1},yticklabel style={/pgf/number format/fixed,/pgf/number format/precision=1},clip=true,enlargelimits=false]%
\path[fill=sifrBlue!4] (axis cs:0,0) rectangle (axis cs:3,3);%
\addplot[sifrBase,line width=.85pt,mark=none] coordinates {(-1.6,0) (3,0)};%
\addplot[sifrBase,line width=.85pt,mark=none] coordinates {(0,-1.6) (0,3)};%
\node[anchor=south west,inner sep=0pt] at (rel axis cs:0,0) {\includegraphics[width=70pt,height=70pt]{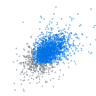}};%
\end{axis}%
\node[anchor=north east,inner sep=0pt,text=sifrBlue,font=\fontfamily{ptm}\fontsize{7.6}{8.8}\selectfont,fill=white,inner sep=.5pt,font=\fontfamily{ptm}\fontsize{7.6}{8.6}\selectfont\bfseries] at (288.95,92.95) {83.9\%};%
\node[anchor=north west,inner sep=0pt,text=sifrInk,font=\fontfamily{ptm}\fontsize{6.2}{7.4}\selectfont,fill=white,inner sep=.5pt] at (221.05,92.95) {3.8\%};%
\node[anchor=south east,inner sep=0pt,text=sifrInk,font=\fontfamily{ptm}\fontsize{6.2}{7.4}\selectfont,fill=white,inner sep=.5pt] at (288.95,25.05) {8.8\%};%
\node[anchor=south west,inner sep=0pt,text=sifrInk,font=\fontfamily{ptm}\fontsize{6.2}{7.4}\selectfont,fill=white,inner sep=.5pt] at (221.05,25.05) {3.5\%};%
\begin{axis}[sifr axis,at={(318pt,24pt)},width=72pt,height=70pt,xmin=18,xmax=28.8,ymin=-2.05875,ymax=12.8671875,xtick={20,24,28},ytick={0,3,6,9,12},ymajorgrids=false,axis line style={black,line width=.5pt},y axis line style={draw=black,line width=.5pt},tick style={black,line width=.5pt},ytick style={draw=black,line width=.5pt},major tick length=2.6pt,minor tick length=1.3pt,minor tick num=1,tick label style={font=\fontfamily{ptm}\fontsize{6.2}{7.4}\selectfont,text=black},xticklabel style={/pgf/number format/fixed,/pgf/number format/precision=1},clip=true]%
\addplot[const plot,draw=none,fill=sifrBase!75] coordinates {(18,0) (18,0.53711) (18.4,0.53711) (18.8,1.5625) (19.2,2.3926) (19.6,2.9297) (20,4.541) (20.4,5.8594) (20.8,7.0312) (21.2,7.8613) (21.6,9.9121) (22,9.5703) (22.4,9.7656) (22.8,8.3008) (23.2,8.0078) (23.6,5.6641) (24,4.1992) (24.4,4.1016) (24.8,2.8809) (25.2,1.416) (25.6,1.0254) (26,0.68359) (26.4,0.24414) (26.8,0.19531) (27.2,0.14648) (27.6,0.14648) (28,0) (28.4,0) (28.8,0) (28.8,0)} \closedcycle;%
\addplot[const plot,draw=none,fill=sifrBlue,fill opacity=.85] coordinates {(18,0) (18,0.048828) (18.4,0) (18.8,0) (19.2,0.19531) (19.6,0.048828) (20,0.14648) (20.4,0.43945) (20.8,0.43945) (21.2,1.3184) (21.6,2.0508) (22,3.1738) (22.4,5.6152) (22.8,5.6152) (23.2,6.9336) (23.6,7.8613) (24,9.4238) (24.4,11.914) (24.8,10.4) (25.2,7.0801) (25.6,6.9824) (26,6.0547) (26.4,5.2734) (26.8,2.9785) (27.2,2.5391) (27.6,1.5137) (28,0.68359) (28.4,0.68359) (28.8,0.68359) (28.8,0)} \closedcycle;%
\addplot[const plot,sifrNFT,line width=.6pt,mark=none] coordinates {(18,0) (18,0.097656) (18.4,0.048828) (18.8,0) (19.2,0.39062) (19.6,0.43945) (20,0.48828) (20.4,1.1719) (20.8,1.5625) (21.2,2.9785) (21.6,4.7363) (22,5.7129) (22.4,7.8613) (22.8,7.5195) (23.2,8.9844) (23.6,10.498) (24,10.254) (24.4,8.6914) (24.8,7.1777) (25.2,6.2012) (25.6,5.2734) (26,3.5645) (26.4,2.2461) (26.8,1.5137) (27.2,1.1719) (27.6,0.73242) (28,0.19531) (28.4,0.19531) (28.8,0.19531) (28.8,0)};%
\addplot[const plot,sifrRAM,line width=.6pt,mark=none] coordinates {(18,0) (18,0.097656) (18.4,0.048828) (18.8,0.097656) (19.2,0) (19.6,0.3418) (20,0.39062) (20.4,0.92773) (20.8,1.2695) (21.2,2.6855) (21.6,3.2227) (22,5.127) (22.4,5.9082) (22.8,7.3242) (23.2,8.8867) (23.6,11.572) (24,9.9121) (24.4,9.082) (24.8,8.1543) (25.2,6.4941) (25.6,5.8594) (26,4.3945) (26.4,2.9785) (26.8,2.002) (27.2,1.3672) (27.6,0.68359) (28,0.58594) (28.4,0.39062) (28.8,0.39062) (28.8,0)};%
\draw[sifrMuted,line width=.9pt] (axis cs:22.2741,-1.74994) -- (axis cs:22.2741,-0.41175);%
\draw[sifrNFT,line width=.9pt] (axis cs:23.9212,-1.74994) -- (axis cs:23.9212,-0.41175);%
\draw[sifrRAM,line width=.9pt] (axis cs:24.137,-1.74994) -- (axis cs:24.137,-0.41175);%
\draw[sifrBlue,line width=.9pt] (axis cs:24.636,-1.74994) -- (axis cs:24.636,-0.41175);%
\end{axis}%
\node[anchor=base,inner sep=0pt,text=sifrInk,font=\fontfamily{ptm}\fontsize{8.0}{9.2}\selectfont] at (354.0,102) {Cosmos 3 Super T2I};%
\node[anchor=center,inner sep=0pt,text=black,font=\fontfamily{ptm}\fontsize{6.8}{8.0}\selectfont,rotate=90] at (305,59.0) {Samples (\%)};%
\node[anchor=base,inner sep=0pt,text=black,font=\fontfamily{ptm}\fontsize{7.2}{8.4}\selectfont] at (354.0,4.5) {PickScore};%
\node[anchor=center,inner sep=0pt,text=black,font=\fontfamily{ptm}\fontsize{7.8}{9.0}\selectfont,rotate=90] at (5,59.0) {Single reward};%
\node[anchor=base,inner sep=0pt,text=sifrNFT,font=\fontfamily{ptm}\fontsize{7.6}{8.8}\selectfont] at (163.0,4.5) {$\Delta$ vs. DiffusionNFT};%
\node[anchor=center,inner sep=0pt,text=sifrRAM,font=\fontfamily{ptm}\fontsize{7.0}{8.2}\selectfont,rotate=90] at (16,59.0) {$\Delta$ vs. RAM};%
\end{pgfinterruptboundingbox}%
\end{tikzpicture}%
\endgroup%

%% file: sections/appendix_analyses.tex
\clearpage
\section{Additional analyses}
\label{app:analysis}

\input{sections/appendix_analyses_rewards}
\input{sections/appendix_analyses_guidance}
\FloatBarrier

%% file: sections/appendix_analyses_rewards.tex
\subsection{Reward distributions and paired comparisons}
\label{app:distribution-protocols}

\paragraph{Evaluation protocol}
For reward distributions, we use each run's highest-reward checkpoint within $500$ rounds and its evaluation guidance.
DiffusionNFT is evaluated guidance-free on SD3.5 Medium and FLUX.2 Klein, following its released protocol; all other trained-model evaluations use CFG.

\begin{figure}[H]
\centering
\includegraphics[width=\textwidth]{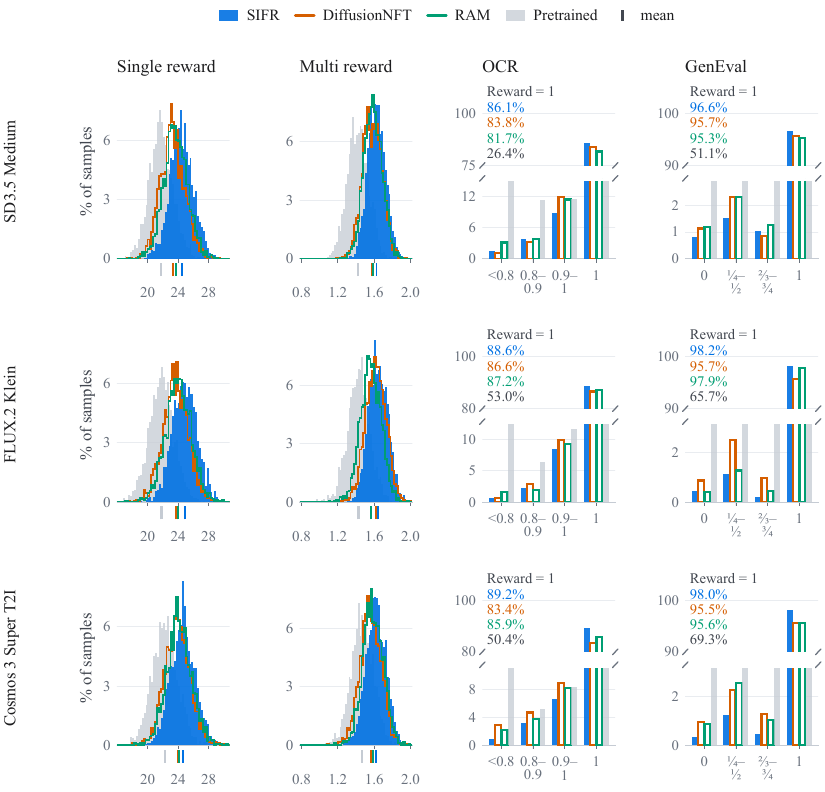}
\caption{Optimized-reward distributions at selected-peak checkpoints. Ticks mark means. OCR and GenEval use reward categories and broken vertical axes; percentages give the perfect-score fractions. GenEval reports per-image partial-credit rewards.}
\label{fig:reward-dist}
\end{figure}

\Cref{fig:reward-dist} shows the full reward distributions.
For OCR and GenEval, most samples receive perfect rewards, and SIFR has the largest perfect-score fraction in each backbone--task comparison.

\begin{figure}[!htbp]
\centering
\includegraphics[width=\textwidth]{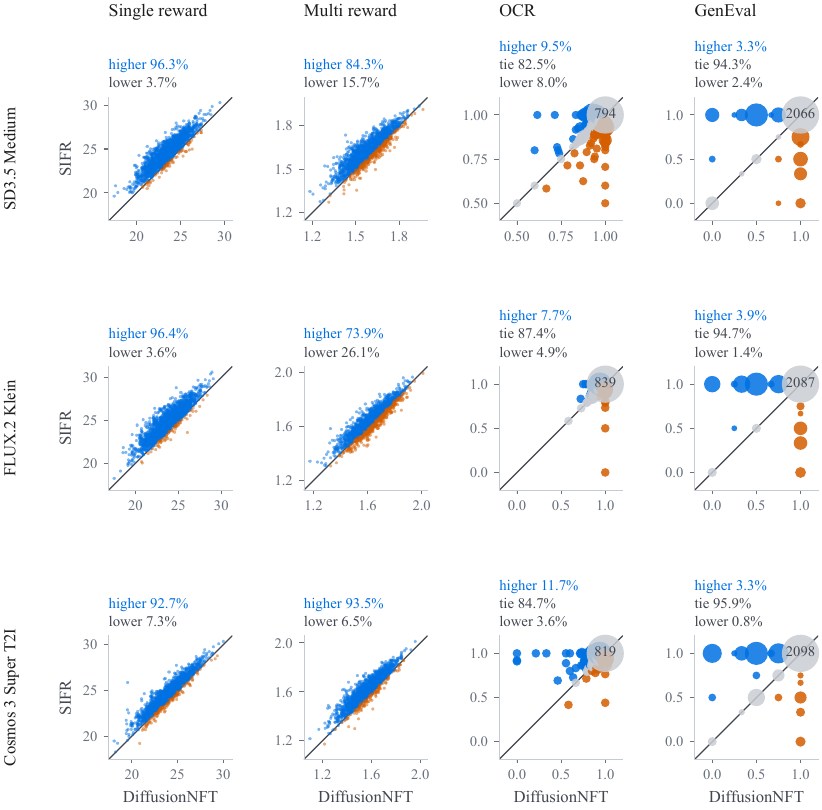}
\caption{Paired rewards against DiffusionNFT at selected-peak checkpoints. Blue points favor SIFR. For OCR and GenEval, circle size indicates repeated pairs; the largest circle is capped and labeled with its count.}
\label{fig:pairwise-nft}
\end{figure}

\begin{figure}[!htbp]
\centering
\includegraphics[width=\textwidth]{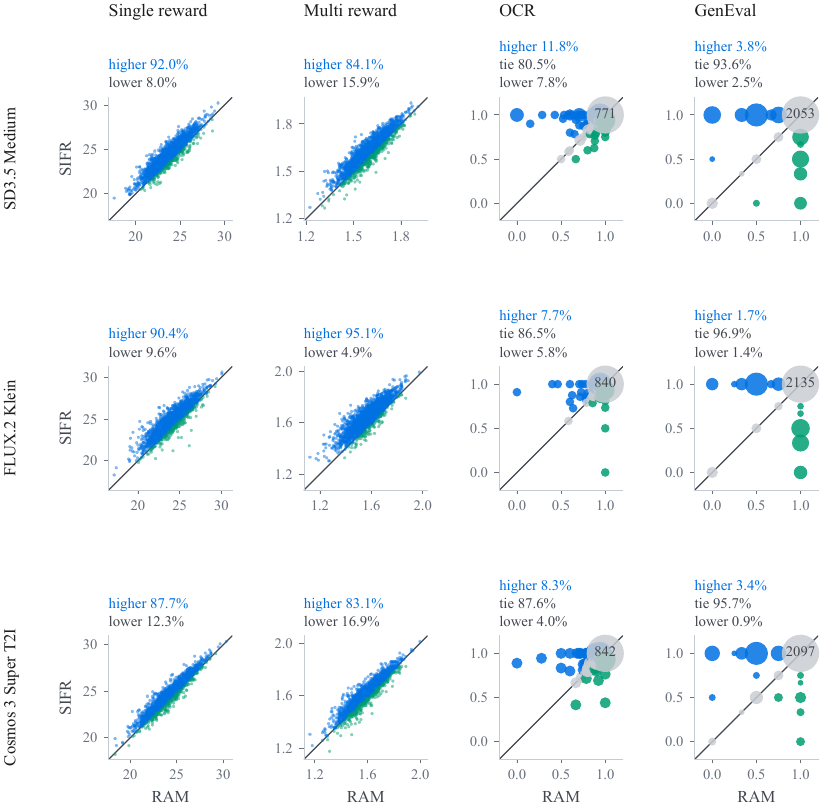}
\caption{Paired rewards against RAM, using the same construction as \cref{fig:pairwise-nft}.}
\label{fig:pairwise-ram}
\end{figure}

\clearpage
\paragraph{Paired rewards}
\Cref{fig:pairwise-nft,fig:pairwise-ram} compare individual samples with DiffusionNFT and RAM.
Every pair matches the prompt and initial noise seed: $2048$ pairs for each preference objective, $1018$ for OCR, and $2212$ for GenEval ($553$ prompts with four seeds).
Blue points indicate higher SIFR reward; the diagonal indicates equal reward.
The preference improvements extend across most samples; on the rule-based tasks most pairs tie at the perfect score.

\clearpage

\clearpage
\paragraph{Higher, equal, and lower reward}
\Cref{tab:pairwise} quantifies the comparisons above.
Intervals use the Wilson method, except for GenEval, where a prompt-cluster bootstrap preserves the dependence among each prompt's four seeds.
\input{tables/paired_rewards}
\FloatBarrier

\paragraph{Held-out reward distribution}
We also examine HPSv3 after multi-reward training on Cosmos 3 Super T2I using the round-$500$ checkpoints in \cref{tab:main}.
All three methods use CFG $2$, $35$ Euler steps, and the same $2048$ prompt--noise pairs.
SIFR's median HPSv3 is $12.05$, compared with $10.78$ for DiffusionNFT and $11.43$ for RAM; its tenth percentile is also higher ($9.26$ versus $8.12$ and $8.68$).

\begin{figure}[!htbp]
\centering
\includegraphics[width=0.8\textwidth]{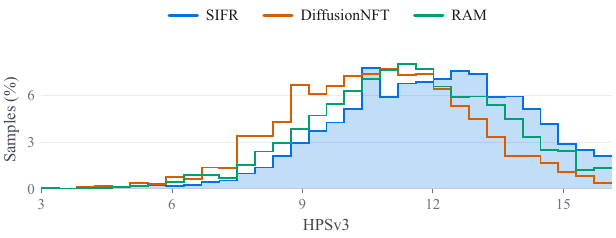}
\caption{Held-out HPSv3 distributions on Cosmos 3 Super T2I after multi-reward training, using round-$500$ checkpoints and matched prompt--noise pairs.}
\label{fig:cosmos-hps-distribution}
\end{figure}
\FloatBarrier

%% file: tables/paired_rewards.tex
\begin{table}[htbp]
\centering
{\small\setlength{\tabcolsep}{2.4pt}\renewcommand{\arraystretch}{1.16}
\begin{tabular}{@{}lcccc@{\hspace{9pt}}cccc@{}}
\toprule
& \multicolumn{4}{c}{vs.\ DiffusionNFT (\%)} & \multicolumn{4}{c}{vs.\ RAM (\%)} \\
\cmidrule(lr){2-5}\cmidrule(lr){6-9}
\tablehead{Objective} & \tablehead{higher} & \tablehead{tie} & \tablehead{lower} & \tablehead{\shortstack{non-tied\\higher}} & \tablehead{higher} & \tablehead{tie} & \tablehead{lower} & \tablehead{\shortstack{non-tied\\higher}} \\
\midrule
\tablegroup{9}{SD3.5 Medium} \\[2pt]
Single reward & 96.3 & 0.0 & 3.7 & 96.3\;[95.4, 97.0] & 92.0 & 0.0 & 8.0 & 92.0\;[90.7, 93.1] \\
Multi reward & 84.3 & 0.0 & 15.7 & 84.3\;[82.7, 85.8] & 84.1 & 0.0 & 15.9 & 84.1\;[82.5, 85.6] \\
OCR & 9.5 & 82.5 & 8.0 & 54.5\;[47.2, 61.6] & 11.8 & 80.5 & 7.8 & 60.3\;[53.4, 66.8] \\
GenEval & 3.3 & 94.3 & 2.4 & 57.5\;[46.8, 67.6] & 3.8 & 93.6 & 2.5 & 60.3\;[50.5, 69.2] \\
\tablesep
\tablegroup{9}{FLUX.2 Klein} \\[2pt]
Single reward & 96.4 & 0.0 & 3.6 & 96.4\;[95.5, 97.2] & 90.4 & 0.0 & 9.6 & 90.4\;[89.0, 91.6] \\
Multi reward & 73.9 & 0.0 & 26.1 & 73.9\;[72.0, 75.8] & 95.1 & 0.0 & 4.9 & 95.1\;[94.0, 95.9] \\
OCR & 7.7 & 87.4 & 4.9 & 60.9\;[52.3, 69.0] & 7.7 & 86.5 & 5.8 & 56.9\;[48.6, 64.9] \\
GenEval & 3.9 & 94.7 & 1.4 & 74.4\;[63.0, 86.0] & 1.7 & 96.9 & 1.4 & 55.9\;[38.4, 73.8] \\
\tablesep
\tablegroup{9}{Cosmos 3 Super T2I} \\[2pt]
Single reward & 92.7 & 0.0 & 7.3 & 92.7\;[91.5, 93.8] & 87.7 & 0.0 & 12.3 & 87.7\;[86.2, 89.0] \\
Multi reward & 93.5 & 0.0 & 6.5 & 93.5\;[92.4, 94.5] & 83.1 & 0.0 & 16.9 & 83.1\;[81.4, 84.6] \\
OCR & 11.7 & 84.7 & 3.6 & 76.3\;[69.0, 82.3] & 8.3 & 87.6 & 4.0 & 67.5\;[58.9, 75.0] \\
GenEval & 3.3 & 95.9 & 0.8 & 81.3\;[70.9, 90.3] & 3.4 & 95.7 & 0.9 & 78.1\;[67.1, 88.3] \\
\bottomrule
\end{tabular}}
\caption{\textbf{Paired reward comparisons.} Percentages of matched prompt--noise pairs on which SIFR scores higher, equal, or lower than each baseline at selected-peak checkpoints (\cref{app:distribution-protocols}). The last column of each block gives the higher-reward fraction among non-tied pairs with a $95\%$ confidence interval. GenEval uses per-image partial-credit rewards.}
\label{tab:pairwise}
\end{table}

%% file: sections/appendix_analyses_guidance.tex
\clearpage
\subsection{Evaluation guidance across backbones}
\label{app:guidance-full}
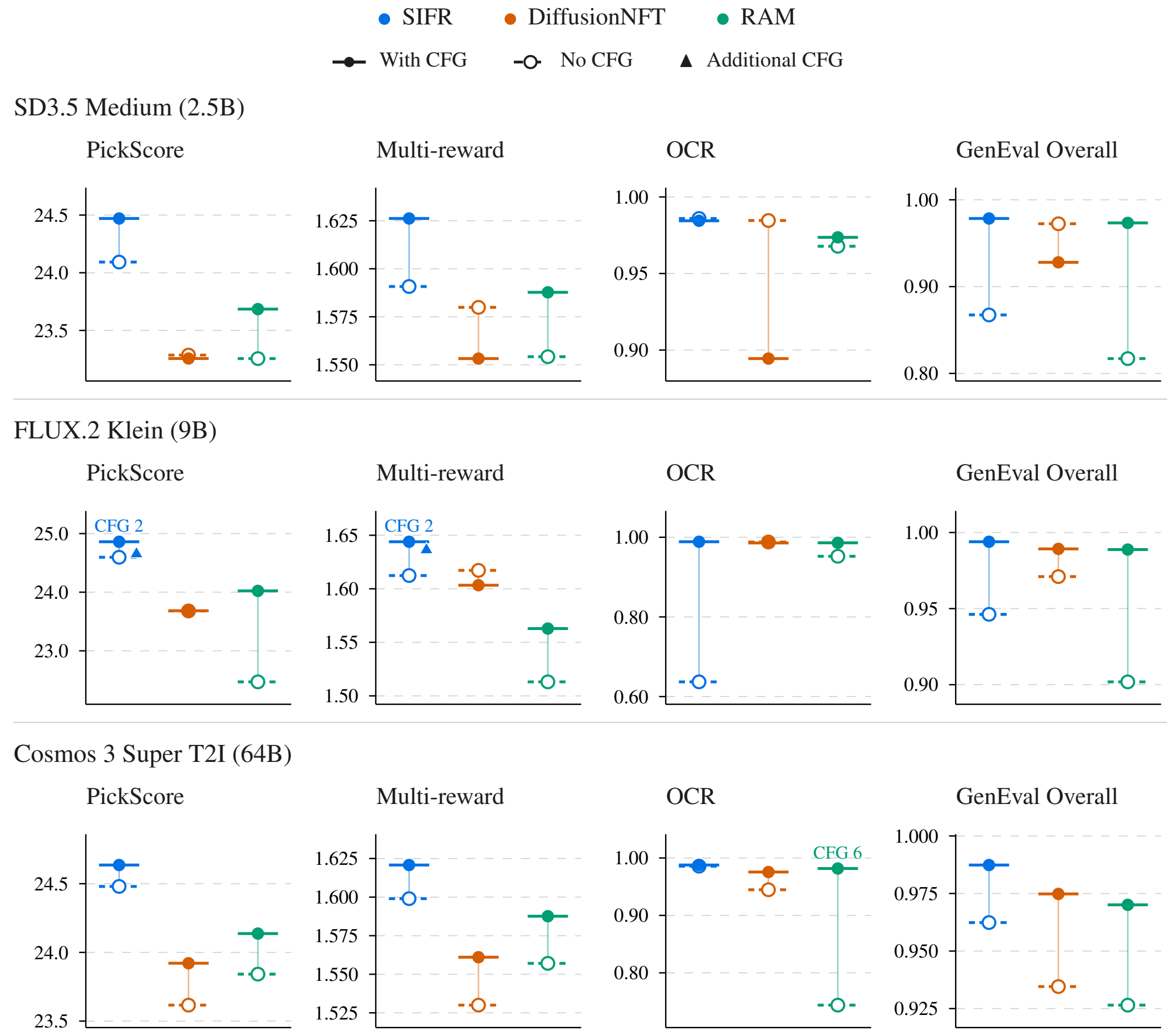
\begin{figure}[!htbp]
\centering
\input{figures/guidance_all_backbones.tex}
\caption{Effect of evaluation CFG across backbones and objectives. Filled and hollow circles show guided and unguided evaluation. Guided settings are CFG $2$ for SD3.5 Medium preference tasks, $4.5$ for its rule-based tasks, $4$ for FLUX.2 Klein, and $2$ for Cosmos 3 Super T2I; labels mark exceptions.}
\label{fig:guidance-full}
\end{figure}

\Cref{fig:guidance-full} extends the guidance comparison in \cref{fig:guidance} to all three backbones and four training objectives.
Each comparison holds the checkpoint, prompts, and noise seeds fixed while changing evaluation guidance.

\FloatBarrier

%% file: figures/guidance_all_backbones.tex
\begingroup%
\input{style/plots.tex}%
\begin{tikzpicture}[x=1pt,y=1pt]
\path[use as bounding box] (0,20) rectangle (396,378);
\node[anchor=base,inner sep=0pt,text=sifrInk,font=\fontfamily{ptm}\fontsize{8}{9.3}\selectfont] at (198,366) {\tikz[baseline=-.5ex]{\fill[sifrBlue] (0,0) circle[radius=1.9pt];}\hspace{4pt}SIFR\hspace{17pt}\tikz[baseline=-.5ex]{\fill[sifrNFT] (0,0) circle[radius=1.9pt];}\hspace{4pt}DiffusionNFT\hspace{17pt}\tikz[baseline=-.5ex]{\fill[sifrRAM] (0,0) circle[radius=1.9pt];}\hspace{4pt}RAM};
\node[anchor=base,inner sep=0pt,text=sifrInk,font=\fontfamily{ptm}\fontsize{7}{8.3}\selectfont] at (198,352) {\tikz[baseline=-.5ex]{\draw[sifrInk,line width=1pt] (0,0)--(11pt,0);\fill[sifrInk] (5.5pt,0) circle[radius=1.7pt];}\hspace{4pt}With CFG\hspace{15pt}\tikz[baseline=-.5ex]{\draw[sifrInk,dashed,line width=.9pt] (0,0)--(11pt,0);\filldraw[white,draw=sifrInk,line width=.7pt] (5.5pt,0) circle[radius=2pt];}\hspace{4pt}No CFG\hspace{15pt}$\blacktriangle$\hspace{4pt}Additional CFG};
\node[anchor=base west,inner sep=0pt,text=sifrInk,font=\fontfamily{ptm}\fontsize{8.5}{9.8}\selectfont] at (8,336) {SD3.5 Medium (2.5B)};
\node[anchor=base west,inner sep=0pt,text=sifrInk,font=\fontfamily{ptm}\fontsize{8}{9.3}\selectfont] at (32,322) {PickScore};
\begin{axis}[sifr axis,at={(32pt,248pt)},width=68pt,height=64pt,axis line style={black,line width=.45pt},y axis line style={draw=black,line width=.45pt},tick style={black,line width=.45pt},ytick style={draw=black,line width=.45pt},tick label style={font=\fontfamily{ptm}\fontsize{6.5}{7.5}\selectfont,text=black},grid style={black!16,line width=.35pt,dashed},major tick length=2.2pt,xmin=-.48,xmax=2.48,ymin=23.061280320000002,ymax=24.737701559999998,xtick=\empty,ytick={23.5,24.0,24.5},yticklabels={23.5,24.0,24.5},clip=false]
\draw[sifrBlue,line width=.5pt,opacity=.45] (axis cs:0,24.092587999999999)--(axis cs:0,24.470445999999999);
\draw[sifrBlue,line width=.95pt,dash pattern=on 2pt off 1.5pt] (axis cs:-0.29,24.092587999999999)--(axis cs:0.29,24.092587999999999);
\addplot[only marks,mark=*,mark size=2.1pt,mark options={fill=white,draw=sifrBlue,line width=.7pt}] coordinates {(0,24.092587999999999)};
\draw[sifrBlue,line width=.95pt] (axis cs:-0.29,24.470445999999999)--(axis cs:0.29,24.470445999999999);
\addplot[only marks,mark=*,mark size=1.7pt,mark options={fill=sifrBlue,draw=sifrBlue,line width=.7pt}] coordinates {(0,24.470445999999999)};
\draw[sifrNFT,line width=.5pt,opacity=.45] (axis cs:1,23.286068)--(axis cs:1,23.256298000000001);
\draw[sifrNFT,line width=.95pt,dash pattern=on 2pt off 1.5pt] (axis cs:0.71,23.286068)--(axis cs:1.29,23.286068);
\addplot[only marks,mark=*,mark size=2.1pt,mark options={fill=white,draw=sifrNFT,line width=.7pt}] coordinates {(1,23.286068)};
\draw[sifrNFT,line width=.95pt] (axis cs:0.71,23.256298000000001)--(axis cs:1.29,23.256298000000001);
\addplot[only marks,mark=*,mark size=1.7pt,mark options={fill=sifrNFT,draw=sifrNFT,line width=.7pt}] coordinates {(1,23.256298000000001)};
\draw[sifrRAM,line width=.5pt,opacity=.45] (axis cs:2,23.255648000000001)--(axis cs:2,23.684674000000001);
\draw[sifrRAM,line width=.95pt,dash pattern=on 2pt off 1.5pt] (axis cs:1.71,23.255648000000001)--(axis cs:2.29,23.255648000000001);
\addplot[only marks,mark=*,mark size=2.1pt,mark options={fill=white,draw=sifrRAM,line width=.7pt}] coordinates {(2,23.255648000000001)};
\draw[sifrRAM,line width=.95pt] (axis cs:1.71,23.684674000000001)--(axis cs:2.29,23.684674000000001);
\addplot[only marks,mark=*,mark size=1.7pt,mark options={fill=sifrRAM,draw=sifrRAM,line width=.7pt}] coordinates {(2,23.684674000000001)};
\end{axis}
\node[anchor=base west,inner sep=0pt,text=sifrInk,font=\fontfamily{ptm}\fontsize{8}{9.3}\selectfont] at (128,322) {Multi-reward};
\begin{axis}[sifr axis,at={(128pt,248pt)},width=68pt,height=64pt,axis line style={black,line width=.45pt},y axis line style={draw=black,line width=.45pt},tick style={black,line width=.45pt},ytick style={draw=black,line width=.45pt},tick label style={font=\fontfamily{ptm}\fontsize{6.5}{7.5}\selectfont,text=black},grid style={black!16,line width=.35pt,dashed},major tick length=2.2pt,xmin=-.48,xmax=2.48,ymin=1.5415534,ymax=1.6422382,xtick=\empty,ytick={1.55,1.575,1.6,1.625},yticklabels={1.550,1.575,1.600,1.625},clip=false]
\draw[sifrBlue,line width=.5pt,opacity=.45] (axis cs:0,1.59076)--(axis cs:0,1.626187);
\draw[sifrBlue,line width=.95pt,dash pattern=on 2pt off 1.5pt] (axis cs:-0.29,1.59076)--(axis cs:0.29,1.59076);
\addplot[only marks,mark=*,mark size=2.1pt,mark options={fill=white,draw=sifrBlue,line width=.7pt}] coordinates {(0,1.59076)};
\draw[sifrBlue,line width=.95pt] (axis cs:-0.29,1.626187)--(axis cs:0.29,1.626187);
\addplot[only marks,mark=*,mark size=1.7pt,mark options={fill=sifrBlue,draw=sifrBlue,line width=.7pt}] coordinates {(0,1.626187)};
\draw[sifrNFT,line width=.5pt,opacity=.45] (axis cs:1,1.5799110000000001)--(axis cs:1,1.5532269999999999);
\draw[sifrNFT,line width=.95pt,dash pattern=on 2pt off 1.5pt] (axis cs:0.71,1.5799110000000001)--(axis cs:1.29,1.5799110000000001);
\addplot[only marks,mark=*,mark size=2.1pt,mark options={fill=white,draw=sifrNFT,line width=.7pt}] coordinates {(1,1.5799110000000001)};
\draw[sifrNFT,line width=.95pt] (axis cs:0.71,1.5532269999999999)--(axis cs:1.29,1.5532269999999999);
\addplot[only marks,mark=*,mark size=1.7pt,mark options={fill=sifrNFT,draw=sifrNFT,line width=.7pt}] coordinates {(1,1.5532269999999999)};
\draw[sifrRAM,line width=.5pt,opacity=.45] (axis cs:2,1.554211)--(axis cs:2,1.5877410000000001);
\draw[sifrRAM,line width=.95pt,dash pattern=on 2pt off 1.5pt] (axis cs:1.71,1.554211)--(axis cs:2.29,1.554211);
\addplot[only marks,mark=*,mark size=2.1pt,mark options={fill=white,draw=sifrRAM,line width=.7pt}] coordinates {(2,1.554211)};
\draw[sifrRAM,line width=.95pt] (axis cs:1.71,1.5877410000000001)--(axis cs:2.29,1.5877410000000001);
\addplot[only marks,mark=*,mark size=1.7pt,mark options={fill=sifrRAM,draw=sifrRAM,line width=.7pt}] coordinates {(2,1.5877410000000001)};
\end{axis}
\node[anchor=base west,inner sep=0pt,text=sifrInk,font=\fontfamily{ptm}\fontsize{8}{9.3}\selectfont] at (224,322) {OCR};
\begin{axis}[sifr axis,at={(224pt,248pt)},width=68pt,height=64pt,axis line style={black,line width=.45pt},y axis line style={draw=black,line width=.45pt},tick style={black,line width=.45pt},ytick style={draw=black,line width=.45pt},tick label style={font=\fontfamily{ptm}\fontsize{6.5}{7.5}\selectfont,text=black},grid style={black!16,line width=.35pt,dashed},major tick length=2.2pt,xmin=-.48,xmax=2.48,ymin=0.87976920000000003,ymax=1.0060806,xtick=\empty,ytick={0.9,0.95,1.0},yticklabels={0.90,0.95,1.00},clip=false]
\draw[sifrBlue,line width=.5pt,opacity=.45] (axis cs:0,0.98594400000000004)--(axis cs:0,0.98444100000000001);
\draw[sifrBlue,line width=.95pt,dash pattern=on 2pt off 1.5pt] (axis cs:-0.29,0.98594400000000004)--(axis cs:0.29,0.98594400000000004);
\addplot[only marks,mark=*,mark size=2.1pt,mark options={fill=white,draw=sifrBlue,line width=.7pt}] coordinates {(0,0.98594400000000004)};
\draw[sifrBlue,line width=.95pt] (axis cs:-0.29,0.98444100000000001)--(axis cs:0.29,0.98444100000000001);
\addplot[only marks,mark=*,mark size=1.7pt,mark options={fill=sifrBlue,draw=sifrBlue,line width=.7pt}] coordinates {(0,0.98444100000000001)};
\draw[sifrNFT,line width=.5pt,opacity=.45] (axis cs:1,0.98461699999999996)--(axis cs:1,0.89441400000000004);
\draw[sifrNFT,line width=.95pt,dash pattern=on 2pt off 1.5pt] (axis cs:0.71,0.98461699999999996)--(axis cs:1.29,0.98461699999999996);
\addplot[only marks,mark=*,mark size=2.1pt,mark options={fill=white,draw=sifrNFT,line width=.7pt}] coordinates {(1,0.98461699999999996)};
\draw[sifrNFT,line width=.95pt] (axis cs:0.71,0.89441400000000004)--(axis cs:1.29,0.89441400000000004);
\addplot[only marks,mark=*,mark size=1.7pt,mark options={fill=sifrNFT,draw=sifrNFT,line width=.7pt}] coordinates {(1,0.89441400000000004)};
\draw[sifrRAM,line width=.5pt,opacity=.45] (axis cs:2,0.96779300000000001)--(axis cs:2,0.97363100000000002);
\draw[sifrRAM,line width=.95pt,dash pattern=on 2pt off 1.5pt] (axis cs:1.71,0.96779300000000001)--(axis cs:2.29,0.96779300000000001);
\addplot[only marks,mark=*,mark size=2.1pt,mark options={fill=white,draw=sifrRAM,line width=.7pt}] coordinates {(2,0.96779300000000001)};
\draw[sifrRAM,line width=.95pt] (axis cs:1.71,0.97363100000000002)--(axis cs:2.29,0.97363100000000002);
\addplot[only marks,mark=*,mark size=1.7pt,mark options={fill=sifrRAM,draw=sifrRAM,line width=.7pt}] coordinates {(2,0.97363100000000002)};
\end{axis}
\node[anchor=base west,inner sep=0pt,text=sifrInk,font=\fontfamily{ptm}\fontsize{8}{9.3}\selectfont] at (320,322) {GenEval Overall};
\begin{axis}[sifr axis,at={(320pt,248pt)},width=68pt,height=64pt,axis line style={black,line width=.45pt},y axis line style={draw=black,line width=.45pt},tick style={black,line width=.45pt},ytick style={draw=black,line width=.45pt},tick label style={font=\fontfamily{ptm}\fontsize{6.5}{7.5}\selectfont,text=black},grid style={black!16,line width=.35pt,dashed},major tick length=2.2pt,xmin=-.48,xmax=2.48,ymin=0.79125604000000005,ymax=1.01396182,xtick=\empty,ytick={0.8,0.9,1.0},yticklabels={0.80,0.90,1.00},clip=false]
\draw[sifrBlue,line width=.5pt,opacity=.45] (axis cs:0,0.867336)--(axis cs:0,0.97845800000000005);
\draw[sifrBlue,line width=.95pt,dash pattern=on 2pt off 1.5pt] (axis cs:-0.29,0.867336)--(axis cs:0.29,0.867336);
\addplot[only marks,mark=*,mark size=2.1pt,mark options={fill=white,draw=sifrBlue,line width=.7pt}] coordinates {(0,0.867336)};
\draw[sifrBlue,line width=.95pt] (axis cs:-0.29,0.97845800000000005)--(axis cs:0.29,0.97845800000000005);
\addplot[only marks,mark=*,mark size=1.7pt,mark options={fill=sifrBlue,draw=sifrBlue,line width=.7pt}] coordinates {(0,0.97845800000000005)};
\draw[sifrNFT,line width=.5pt,opacity=.45] (axis cs:1,0.972383)--(axis cs:1,0.92796000000000001);
\draw[sifrNFT,line width=.95pt,dash pattern=on 2pt off 1.5pt] (axis cs:0.71,0.972383)--(axis cs:1.29,0.972383);
\addplot[only marks,mark=*,mark size=2.1pt,mark options={fill=white,draw=sifrNFT,line width=.7pt}] coordinates {(1,0.972383)};
\draw[sifrNFT,line width=.95pt] (axis cs:0.71,0.92796000000000001)--(axis cs:1.29,0.92796000000000001);
\addplot[only marks,mark=*,mark size=1.7pt,mark options={fill=sifrNFT,draw=sifrNFT,line width=.7pt}] coordinates {(1,0.92796000000000001)};
\draw[sifrRAM,line width=.5pt,opacity=.45] (axis cs:2,0.81707700000000005)--(axis cs:2,0.97338000000000002);
\draw[sifrRAM,line width=.95pt,dash pattern=on 2pt off 1.5pt] (axis cs:1.71,0.81707700000000005)--(axis cs:2.29,0.81707700000000005);
\addplot[only marks,mark=*,mark size=2.1pt,mark options={fill=white,draw=sifrRAM,line width=.7pt}] coordinates {(2,0.81707700000000005)};
\draw[sifrRAM,line width=.95pt] (axis cs:1.71,0.97338000000000002)--(axis cs:2.29,0.97338000000000002);
\addplot[only marks,mark=*,mark size=1.7pt,mark options={fill=sifrRAM,draw=sifrRAM,line width=.7pt}] coordinates {(2,0.97338000000000002)};
\end{axis}
\node[anchor=base west,inner sep=0pt,text=sifrInk,font=\fontfamily{ptm}\fontsize{8.5}{9.8}\selectfont] at (8,229) {FLUX.2 Klein (9B)};
\draw[sifrBase,line width=.35pt] (8,242)--(390,242);
\node[anchor=base west,inner sep=0pt,text=sifrInk,font=\fontfamily{ptm}\fontsize{8}{9.3}\selectfont] at (32,215) {PickScore};
\begin{axis}[sifr axis,at={(32pt,141pt)},width=68pt,height=64pt,axis line style={black,line width=.45pt},y axis line style={draw=black,line width=.45pt},tick style={black,line width=.45pt},ytick style={draw=black,line width=.45pt},tick label style={font=\fontfamily{ptm}\fontsize{6.5}{7.5}\selectfont,text=black},grid style={black!16,line width=.35pt,dashed},major tick length=2.2pt,xmin=-.48,xmax=2.48,ymin=22.087919360000001,ymax=25.38633188,xtick=\empty,ytick={23,24,25},yticklabels={23.0,24.0,25.0},clip=false]
\draw[sifrBlue,line width=.5pt,opacity=.45] (axis cs:0,24.595844)--(axis cs:0,24.860498);
\draw[sifrBlue,line width=.95pt,dash pattern=on 2pt off 1.5pt] (axis cs:-0.29,24.595844)--(axis cs:0.29,24.595844);
\addplot[only marks,mark=*,mark size=2.1pt,mark options={fill=white,draw=sifrBlue,line width=.7pt}] coordinates {(0,24.595844)};
\draw[sifrBlue,line width=.95pt] (axis cs:-0.29,24.860498)--(axis cs:0.29,24.860498);
\addplot[only marks,mark=*,mark size=1.7pt,mark options={fill=sifrBlue,draw=sifrBlue,line width=.7pt}] coordinates {(0,24.860498)};
\node[anchor=south,yshift=3pt,inner sep=0pt,text=sifrBlue,font=\fontfamily{ptm}\fontsize{6}{7}\selectfont] at (axis cs:0,24.860498) {CFG 2};
\addplot[sifrBlue,only marks,mark=triangle*,mark size=2.5pt,mark options={draw=white,line width=.3pt}] coordinates {(0.25,24.652498000000001)};
\draw[sifrNFT,line width=.5pt,opacity=.45] (axis cs:1,23.680202000000001)--(axis cs:1,23.683738000000002);
\draw[sifrNFT,line width=.95pt,dash pattern=on 2pt off 1.5pt] (axis cs:0.71,23.680202000000001)--(axis cs:1.29,23.680202000000001);
\addplot[only marks,mark=*,mark size=2.1pt,mark options={fill=white,draw=sifrNFT,line width=.7pt}] coordinates {(1,23.680202000000001)};
\draw[sifrNFT,line width=.95pt] (axis cs:0.71,23.683738000000002)--(axis cs:1.29,23.683738000000002);
\addplot[only marks,mark=*,mark size=1.7pt,mark options={fill=sifrNFT,draw=sifrNFT,line width=.7pt}] coordinates {(1,23.683738000000002)};
\draw[sifrRAM,line width=.5pt,opacity=.45] (axis cs:2,22.470344000000001)--(axis cs:2,24.0227);
\draw[sifrRAM,line width=.95pt,dash pattern=on 2pt off 1.5pt] (axis cs:1.71,22.470344000000001)--(axis cs:2.29,22.470344000000001);
\addplot[only marks,mark=*,mark size=2.1pt,mark options={fill=white,draw=sifrRAM,line width=.7pt}] coordinates {(2,22.470344000000001)};
\draw[sifrRAM,line width=.95pt] (axis cs:1.71,24.0227)--(axis cs:2.29,24.0227);
\addplot[only marks,mark=*,mark size=1.7pt,mark options={fill=sifrRAM,draw=sifrRAM,line width=.7pt}] coordinates {(2,24.0227)};
\end{axis}
\node[anchor=base west,inner sep=0pt,text=sifrInk,font=\fontfamily{ptm}\fontsize{8}{9.3}\selectfont] at (128,215) {Multi-reward};
\begin{axis}[sifr axis,at={(128pt,141pt)},width=68pt,height=64pt,axis line style={black,line width=.45pt},y axis line style={draw=black,line width=.45pt},tick style={black,line width=.45pt},ytick style={draw=black,line width=.45pt},tick label style={font=\fontfamily{ptm}\fontsize{6.5}{7.5}\selectfont,text=black},grid style={black!16,line width=.35pt,dashed},major tick length=2.2pt,xmin=-.48,xmax=2.48,ymin=1.4921335600000001,ymax=1.67271898,xtick=\empty,ytick={1.5,1.55,1.6,1.65},yticklabels={1.50,1.55,1.60,1.65},clip=false]
\draw[sifrBlue,line width=.5pt,opacity=.45] (axis cs:0,1.6123769999999999)--(axis cs:0,1.6439299999999999);
\draw[sifrBlue,line width=.95pt,dash pattern=on 2pt off 1.5pt] (axis cs:-0.29,1.6123769999999999)--(axis cs:0.29,1.6123769999999999);
\addplot[only marks,mark=*,mark size=2.1pt,mark options={fill=white,draw=sifrBlue,line width=.7pt}] coordinates {(0,1.6123769999999999)};
\draw[sifrBlue,line width=.95pt] (axis cs:-0.29,1.6439299999999999)--(axis cs:0.29,1.6439299999999999);
\addplot[only marks,mark=*,mark size=1.7pt,mark options={fill=sifrBlue,draw=sifrBlue,line width=.7pt}] coordinates {(0,1.6439299999999999)};
\node[anchor=south,yshift=3pt,inner sep=0pt,text=sifrBlue,font=\fontfamily{ptm}\fontsize{6}{7}\selectfont] at (axis cs:0,1.6439299999999999) {CFG 2};
\addplot[sifrBlue,only marks,mark=triangle*,mark size=2.5pt,mark options={draw=white,line width=.3pt}] coordinates {(0.25,1.636307)};
\draw[sifrNFT,line width=.5pt,opacity=.45] (axis cs:1,1.617211)--(axis cs:1,1.6033310000000001);
\draw[sifrNFT,line width=.95pt,dash pattern=on 2pt off 1.5pt] (axis cs:0.71,1.617211)--(axis cs:1.29,1.617211);
\addplot[only marks,mark=*,mark size=2.1pt,mark options={fill=white,draw=sifrNFT,line width=.7pt}] coordinates {(1,1.617211)};
\draw[sifrNFT,line width=.95pt] (axis cs:0.71,1.6033310000000001)--(axis cs:1.29,1.6033310000000001);
\addplot[only marks,mark=*,mark size=1.7pt,mark options={fill=sifrNFT,draw=sifrNFT,line width=.7pt}] coordinates {(1,1.6033310000000001)};
\draw[sifrRAM,line width=.5pt,opacity=.45] (axis cs:2,1.5130710000000001)--(axis cs:2,1.5628470000000001);
\draw[sifrRAM,line width=.95pt,dash pattern=on 2pt off 1.5pt] (axis cs:1.71,1.5130710000000001)--(axis cs:2.29,1.5130710000000001);
\addplot[only marks,mark=*,mark size=2.1pt,mark options={fill=white,draw=sifrRAM,line width=.7pt}] coordinates {(2,1.5130710000000001)};
\draw[sifrRAM,line width=.95pt] (axis cs:1.71,1.5628470000000001)--(axis cs:2.29,1.5628470000000001);
\addplot[only marks,mark=*,mark size=1.7pt,mark options={fill=sifrRAM,draw=sifrRAM,line width=.7pt}] coordinates {(2,1.5628470000000001)};
\end{axis}
\node[anchor=base west,inner sep=0pt,text=sifrInk,font=\fontfamily{ptm}\fontsize{8}{9.3}\selectfont] at (224,215) {OCR};
\begin{axis}[sifr axis,at={(224pt,141pt)},width=68pt,height=64pt,axis line style={black,line width=.45pt},y axis line style={draw=black,line width=.45pt},tick style={black,line width=.45pt},ytick style={draw=black,line width=.45pt},tick label style={font=\fontfamily{ptm}\fontsize{6.5}{7.5}\selectfont,text=black},grid style={black!16,line width=.35pt,dashed},major tick length=2.2pt,xmin=-.48,xmax=2.48,ymin=0.58064492000000001,ymax=1.06611236,xtick=\empty,ytick={0.6,0.8,1.0},yticklabels={0.60,0.80,1.00},clip=false]
\draw[sifrBlue,line width=.5pt,opacity=.45] (axis cs:0,0.63693100000000002)--(axis cs:0,0.98871900000000001);
\draw[sifrBlue,line width=.95pt,dash pattern=on 2pt off 1.5pt] (axis cs:-0.29,0.63693100000000002)--(axis cs:0.29,0.63693100000000002);
\addplot[only marks,mark=*,mark size=2.1pt,mark options={fill=white,draw=sifrBlue,line width=.7pt}] coordinates {(0,0.63693100000000002)};
\draw[sifrBlue,line width=.95pt] (axis cs:-0.29,0.98871900000000001)--(axis cs:0.29,0.98871900000000001);
\addplot[only marks,mark=*,mark size=1.7pt,mark options={fill=sifrBlue,draw=sifrBlue,line width=.7pt}] coordinates {(0,0.98871900000000001)};
\draw[sifrNFT,line width=.5pt,opacity=.45] (axis cs:1,0.98808499999999999)--(axis cs:1,0.985734);
\draw[sifrNFT,line width=.95pt,dash pattern=on 2pt off 1.5pt] (axis cs:0.71,0.98808499999999999)--(axis cs:1.29,0.98808499999999999);
\addplot[only marks,mark=*,mark size=2.1pt,mark options={fill=white,draw=sifrNFT,line width=.7pt}] coordinates {(1,0.98808499999999999)};
\draw[sifrNFT,line width=.95pt] (axis cs:0.71,0.985734)--(axis cs:1.29,0.985734);
\addplot[only marks,mark=*,mark size=1.7pt,mark options={fill=sifrNFT,draw=sifrNFT,line width=.7pt}] coordinates {(1,0.985734)};
\draw[sifrRAM,line width=.5pt,opacity=.45] (axis cs:2,0.952241)--(axis cs:2,0.98602100000000004);
\draw[sifrRAM,line width=.95pt,dash pattern=on 2pt off 1.5pt] (axis cs:1.71,0.952241)--(axis cs:2.29,0.952241);
\addplot[only marks,mark=*,mark size=2.1pt,mark options={fill=white,draw=sifrRAM,line width=.7pt}] coordinates {(2,0.952241)};
\draw[sifrRAM,line width=.95pt] (axis cs:1.71,0.98602100000000004)--(axis cs:2.29,0.98602100000000004);
\addplot[only marks,mark=*,mark size=1.7pt,mark options={fill=sifrRAM,draw=sifrRAM,line width=.7pt}] coordinates {(2,0.98602100000000004)};
\end{axis}
\node[anchor=base west,inner sep=0pt,text=sifrInk,font=\fontfamily{ptm}\fontsize{8}{9.3}\selectfont] at (320,215) {GenEval Overall};
\begin{axis}[sifr axis,at={(320pt,141pt)},width=68pt,height=64pt,axis line style={black,line width=.45pt},y axis line style={draw=black,line width=.45pt},tick style={black,line width=.45pt},ytick style={draw=black,line width=.45pt},tick label style={font=\fontfamily{ptm}\fontsize{6.5}{7.5}\selectfont,text=black},grid style={black!16,line width=.35pt,dashed},major tick length=2.2pt,xmin=-.48,xmax=2.48,ymin=0.88710644000000005,ymax=1.0141920200000001,xtick=\empty,ytick={0.9,0.95,1.0},yticklabels={0.90,0.95,1.00},clip=false]
\draw[sifrBlue,line width=.5pt,opacity=.45] (axis cs:0,0.94622099999999998)--(axis cs:0,0.99393200000000004);
\draw[sifrBlue,line width=.95pt,dash pattern=on 2pt off 1.5pt] (axis cs:-0.29,0.94622099999999998)--(axis cs:0.29,0.94622099999999998);
\addplot[only marks,mark=*,mark size=2.1pt,mark options={fill=white,draw=sifrBlue,line width=.7pt}] coordinates {(0,0.94622099999999998)};
\draw[sifrBlue,line width=.95pt] (axis cs:-0.29,0.99393200000000004)--(axis cs:0.29,0.99393200000000004);
\addplot[only marks,mark=*,mark size=1.7pt,mark options={fill=sifrBlue,draw=sifrBlue,line width=.7pt}] coordinates {(0,0.99393200000000004)};
\draw[sifrNFT,line width=.5pt,opacity=.45] (axis cs:1,0.97102500000000003)--(axis cs:1,0.98923799999999995);
\draw[sifrNFT,line width=.95pt,dash pattern=on 2pt off 1.5pt] (axis cs:0.71,0.97102500000000003)--(axis cs:1.29,0.97102500000000003);
\addplot[only marks,mark=*,mark size=2.1pt,mark options={fill=white,draw=sifrNFT,line width=.7pt}] coordinates {(1,0.97102500000000003)};
\draw[sifrNFT,line width=.95pt] (axis cs:0.71,0.98923799999999995)--(axis cs:1.29,0.98923799999999995);
\addplot[only marks,mark=*,mark size=1.7pt,mark options={fill=sifrNFT,draw=sifrNFT,line width=.7pt}] coordinates {(1,0.98923799999999995)};
\draw[sifrRAM,line width=.5pt,opacity=.45] (axis cs:2,0.901841)--(axis cs:2,0.98880100000000004);
\draw[sifrRAM,line width=.95pt,dash pattern=on 2pt off 1.5pt] (axis cs:1.71,0.901841)--(axis cs:2.29,0.901841);
\addplot[only marks,mark=*,mark size=2.1pt,mark options={fill=white,draw=sifrRAM,line width=.7pt}] coordinates {(2,0.901841)};
\draw[sifrRAM,line width=.95pt] (axis cs:1.71,0.98880100000000004)--(axis cs:2.29,0.98880100000000004);
\addplot[only marks,mark=*,mark size=1.7pt,mark options={fill=sifrRAM,draw=sifrRAM,line width=.7pt}] coordinates {(2,0.98880100000000004)};
\end{axis}
\node[anchor=base west,inner sep=0pt,text=sifrInk,font=\fontfamily{ptm}\fontsize{8.5}{9.8}\selectfont] at (8,122) {Cosmos 3 Super T2I (64B)};
\draw[sifrBase,line width=.35pt] (8,135)--(390,135);
\node[anchor=base west,inner sep=0pt,text=sifrInk,font=\fontfamily{ptm}\fontsize{8}{9.3}\selectfont] at (32,108) {PickScore};
\begin{axis}[sifr axis,at={(32pt,34pt)},width=68pt,height=64pt,axis line style={black,line width=.45pt},y axis line style={draw=black,line width=.45pt},tick style={black,line width=.45pt},ytick style={draw=black,line width=.45pt},tick label style={font=\fontfamily{ptm}\fontsize{6.5}{7.5}\selectfont,text=black},grid style={black!16,line width=.35pt,dashed},major tick length=2.2pt,xmin=-.48,xmax=2.48,ymin=23.452867360000003,ymax=24.86040388,xtick=\empty,ytick={23.5,24.0,24.5},yticklabels={23.5,24.0,24.5},clip=false]
\draw[sifrBlue,line width=.5pt,opacity=.45] (axis cs:0,24.480143999999999)--(axis cs:0,24.636013999999999);
\draw[sifrBlue,line width=.95pt,dash pattern=on 2pt off 1.5pt] (axis cs:-0.29,24.480143999999999)--(axis cs:0.29,24.480143999999999);
\addplot[only marks,mark=*,mark size=2.1pt,mark options={fill=white,draw=sifrBlue,line width=.7pt}] coordinates {(0,24.480143999999999)};
\draw[sifrBlue,line width=.95pt] (axis cs:-0.29,24.636013999999999)--(axis cs:0.29,24.636013999999999);
\addplot[only marks,mark=*,mark size=1.7pt,mark options={fill=sifrBlue,draw=sifrBlue,line width=.7pt}] coordinates {(0,24.636013999999999)};
\draw[sifrNFT,line width=.5pt,opacity=.45] (axis cs:1,23.616060000000001)--(axis cs:1,23.921195999999998);
\draw[sifrNFT,line width=.95pt,dash pattern=on 2pt off 1.5pt] (axis cs:0.71,23.616060000000001)--(axis cs:1.29,23.616060000000001);
\addplot[only marks,mark=*,mark size=2.1pt,mark options={fill=white,draw=sifrNFT,line width=.7pt}] coordinates {(1,23.616060000000001)};
\draw[sifrNFT,line width=.95pt] (axis cs:0.71,23.921195999999998)--(axis cs:1.29,23.921195999999998);
\addplot[only marks,mark=*,mark size=1.7pt,mark options={fill=sifrNFT,draw=sifrNFT,line width=.7pt}] coordinates {(1,23.921195999999998)};
\draw[sifrRAM,line width=.5pt,opacity=.45] (axis cs:2,23.841740000000001)--(axis cs:2,24.136996);
\draw[sifrRAM,line width=.95pt,dash pattern=on 2pt off 1.5pt] (axis cs:1.71,23.841740000000001)--(axis cs:2.29,23.841740000000001);
\addplot[only marks,mark=*,mark size=2.1pt,mark options={fill=white,draw=sifrRAM,line width=.7pt}] coordinates {(2,23.841740000000001)};
\draw[sifrRAM,line width=.95pt] (axis cs:1.71,24.136996)--(axis cs:2.29,24.136996);
\addplot[only marks,mark=*,mark size=1.7pt,mark options={fill=sifrRAM,draw=sifrRAM,line width=.7pt}] coordinates {(2,24.136996)};
\end{axis}
\node[anchor=base west,inner sep=0pt,text=sifrInk,font=\fontfamily{ptm}\fontsize{8}{9.3}\selectfont] at (128,108) {Multi-reward};
\begin{axis}[sifr axis,at={(128pt,34pt)},width=68pt,height=64pt,axis line style={black,line width=.45pt},y axis line style={draw=black,line width=.45pt},tick style={black,line width=.45pt},ytick style={draw=black,line width=.45pt},tick label style={font=\fontfamily{ptm}\fontsize{6.5}{7.5}\selectfont,text=black},grid style={black!16,line width=.35pt,dashed},major tick length=2.2pt,xmin=-.48,xmax=2.48,ymin=1.5154652799999999,ymax=1.64075824,xtick=\empty,ytick={1.525,1.55,1.575,1.6,1.625},yticklabels={1.525,1.550,1.575,1.600,1.625},clip=false]
\draw[sifrBlue,line width=.5pt,opacity=.45] (axis cs:0,1.5989979999999999)--(axis cs:0,1.620784);
\draw[sifrBlue,line width=.95pt,dash pattern=on 2pt off 1.5pt] (axis cs:-0.29,1.5989979999999999)--(axis cs:0.29,1.5989979999999999);
\addplot[only marks,mark=*,mark size=2.1pt,mark options={fill=white,draw=sifrBlue,line width=.7pt}] coordinates {(0,1.5989979999999999)};
\draw[sifrBlue,line width=.95pt] (axis cs:-0.29,1.620784)--(axis cs:0.29,1.620784);
\addplot[only marks,mark=*,mark size=1.7pt,mark options={fill=sifrBlue,draw=sifrBlue,line width=.7pt}] coordinates {(0,1.620784)};
\draw[sifrNFT,line width=.5pt,opacity=.45] (axis cs:1,1.529992)--(axis cs:1,1.561051);
\draw[sifrNFT,line width=.95pt,dash pattern=on 2pt off 1.5pt] (axis cs:0.71,1.529992)--(axis cs:1.29,1.529992);
\addplot[only marks,mark=*,mark size=2.1pt,mark options={fill=white,draw=sifrNFT,line width=.7pt}] coordinates {(1,1.529992)};
\draw[sifrNFT,line width=.95pt] (axis cs:0.71,1.561051)--(axis cs:1.29,1.561051);
\addplot[only marks,mark=*,mark size=1.7pt,mark options={fill=sifrNFT,draw=sifrNFT,line width=.7pt}] coordinates {(1,1.561051)};
\draw[sifrRAM,line width=.5pt,opacity=.45] (axis cs:2,1.557061)--(axis cs:2,1.5876170000000001);
\draw[sifrRAM,line width=.95pt,dash pattern=on 2pt off 1.5pt] (axis cs:1.71,1.557061)--(axis cs:2.29,1.557061);
\addplot[only marks,mark=*,mark size=2.1pt,mark options={fill=white,draw=sifrRAM,line width=.7pt}] coordinates {(2,1.557061)};
\draw[sifrRAM,line width=.95pt] (axis cs:1.71,1.5876170000000001)--(axis cs:2.29,1.5876170000000001);
\addplot[only marks,mark=*,mark size=1.7pt,mark options={fill=sifrRAM,draw=sifrRAM,line width=.7pt}] coordinates {(2,1.5876170000000001)};
\end{axis}
\node[anchor=base west,inner sep=0pt,text=sifrInk,font=\fontfamily{ptm}\fontsize{8}{9.3}\selectfont] at (224,108) {OCR};
\begin{axis}[sifr axis,at={(224pt,34pt)},width=68pt,height=64pt,axis line style={black,line width=.45pt},y axis line style={draw=black,line width=.45pt},tick style={black,line width=.45pt},ytick style={draw=black,line width=.45pt},tick label style={font=\fontfamily{ptm}\fontsize{6.5}{7.5}\selectfont,text=black},grid style={black!16,line width=.35pt,dashed},major tick length=2.2pt,xmin=-.48,xmax=2.48,ymin=0.70472884000000002,ymax=1.04141572,xtick=\empty,ytick={0.8,0.9,1.0},yticklabels={0.80,0.90,1.00},clip=false]
\draw[sifrBlue,line width=.5pt,opacity=.45] (axis cs:0,0.98571399999999998)--(axis cs:0,0.98774099999999998);
\draw[sifrBlue,line width=.95pt,dash pattern=on 2pt off 1.5pt] (axis cs:-0.29,0.98571399999999998)--(axis cs:0.29,0.98571399999999998);
\addplot[only marks,mark=*,mark size=2.1pt,mark options={fill=white,draw=sifrBlue,line width=.7pt}] coordinates {(0,0.98571399999999998)};
\draw[sifrBlue,line width=.95pt] (axis cs:-0.29,0.98774099999999998)--(axis cs:0.29,0.98774099999999998);
\addplot[only marks,mark=*,mark size=1.7pt,mark options={fill=sifrBlue,draw=sifrBlue,line width=.7pt}] coordinates {(0,0.98774099999999998)};
\draw[sifrNFT,line width=.5pt,opacity=.45] (axis cs:1,0.94466700000000003)--(axis cs:1,0.97553000000000001);
\draw[sifrNFT,line width=.95pt,dash pattern=on 2pt off 1.5pt] (axis cs:0.71,0.94466700000000003)--(axis cs:1.29,0.94466700000000003);
\addplot[only marks,mark=*,mark size=2.1pt,mark options={fill=white,draw=sifrNFT,line width=.7pt}] coordinates {(1,0.94466700000000003)};
\draw[sifrNFT,line width=.95pt] (axis cs:0.71,0.97553000000000001)--(axis cs:1.29,0.97553000000000001);
\addplot[only marks,mark=*,mark size=1.7pt,mark options={fill=sifrNFT,draw=sifrNFT,line width=.7pt}] coordinates {(1,0.97553000000000001)};
\draw[sifrRAM,line width=.5pt,opacity=.45] (axis cs:2,0.74376500000000001)--(axis cs:2,0.98167199999999999);
\draw[sifrRAM,line width=.95pt,dash pattern=on 2pt off 1.5pt] (axis cs:1.71,0.74376500000000001)--(axis cs:2.29,0.74376500000000001);
\addplot[only marks,mark=*,mark size=2.1pt,mark options={fill=white,draw=sifrRAM,line width=.7pt}] coordinates {(2,0.74376500000000001)};
\draw[sifrRAM,line width=.95pt] (axis cs:1.71,0.98167199999999999)--(axis cs:2.29,0.98167199999999999);
\addplot[only marks,mark=*,mark size=1.7pt,mark options={fill=sifrRAM,draw=sifrRAM,line width=.7pt}] coordinates {(2,0.98167199999999999)};
\node[anchor=south,yshift=3pt,inner sep=0pt,text=sifrRAM,font=\fontfamily{ptm}\fontsize{6}{7}\selectfont] at (axis cs:2,0.98167199999999999) {CFG 6};
\end{axis}
\node[anchor=base west,inner sep=0pt,text=sifrInk,font=\fontfamily{ptm}\fontsize{8}{9.3}\selectfont] at (320,108) {GenEval Overall};
\begin{axis}[sifr axis,at={(320pt,34pt)},width=68pt,height=64pt,axis line style={black,line width=.45pt},y axis line style={draw=black,line width=.45pt},tick style={black,line width=.45pt},ytick style={draw=black,line width=.45pt},tick label style={font=\fontfamily{ptm}\fontsize{6.5}{7.5}\selectfont,text=black},grid style={black!16,line width=.35pt,dashed},major tick length=2.2pt,xmin=-.48,xmax=2.48,ymin=0.91675695999999995,ymax=1.00075618,xtick=\empty,ytick={0.925,0.95,0.975,1.0},yticklabels={0.925,0.950,0.975,1.000},clip=false]
\draw[sifrBlue,line width=.5pt,opacity=.45] (axis cs:0,0.96236600000000005)--(axis cs:0,0.98736500000000005);
\draw[sifrBlue,line width=.95pt,dash pattern=on 2pt off 1.5pt] (axis cs:-0.29,0.96236600000000005)--(axis cs:0.29,0.96236600000000005);
\addplot[only marks,mark=*,mark size=2.1pt,mark options={fill=white,draw=sifrBlue,line width=.7pt}] coordinates {(0,0.96236600000000005)};
\draw[sifrBlue,line width=.95pt] (axis cs:-0.29,0.98736500000000005)--(axis cs:0.29,0.98736500000000005);
\addplot[only marks,mark=*,mark size=1.7pt,mark options={fill=sifrBlue,draw=sifrBlue,line width=.7pt}] coordinates {(0,0.98736500000000005)};
\draw[sifrNFT,line width=.5pt,opacity=.45] (axis cs:1,0.934562)--(axis cs:1,0.97481200000000001);
\draw[sifrNFT,line width=.95pt,dash pattern=on 2pt off 1.5pt] (axis cs:0.71,0.934562)--(axis cs:1.29,0.934562);
\addplot[only marks,mark=*,mark size=2.1pt,mark options={fill=white,draw=sifrNFT,line width=.7pt}] coordinates {(1,0.934562)};
\draw[sifrNFT,line width=.95pt] (axis cs:0.71,0.97481200000000001)--(axis cs:1.29,0.97481200000000001);
\addplot[only marks,mark=*,mark size=1.7pt,mark options={fill=sifrNFT,draw=sifrNFT,line width=.7pt}] coordinates {(1,0.97481200000000001)};
\draw[sifrRAM,line width=.5pt,opacity=.45] (axis cs:2,0.92649599999999999)--(axis cs:2,0.97008099999999997);
\draw[sifrRAM,line width=.95pt,dash pattern=on 2pt off 1.5pt] (axis cs:1.71,0.92649599999999999)--(axis cs:2.29,0.92649599999999999);
\addplot[only marks,mark=*,mark size=2.1pt,mark options={fill=white,draw=sifrRAM,line width=.7pt}] coordinates {(2,0.92649599999999999)};
\draw[sifrRAM,line width=.95pt] (axis cs:1.71,0.97008099999999997)--(axis cs:2.29,0.97008099999999997);
\addplot[only marks,mark=*,mark size=1.7pt,mark options={fill=sifrRAM,draw=sifrRAM,line width=.7pt}] coordinates {(2,0.97008099999999997)};
\end{axis}
\end{tikzpicture}%
\endgroup%